\documentclass{article}

\usepackage[final,nonatbib]{neurips_2021}

\usepackage[utf8]{inputenc} 
\usepackage[T1]{fontenc}    
\usepackage[pagebackref=true,colorlinks,urlcolor=magenta]{hyperref}       
\usepackage{url}            
\usepackage{booktabs}       
\usepackage{amsfonts}       
\usepackage{nicefrac}       
\usepackage[pdftex]{graphicx}
\usepackage{microtype}      
\usepackage[table,dvipsnames,math]{xcolor}         
\usepackage{subcaption}
\usepackage{amsmath}
\usepackage{multirow}

\usepackage{caption}
\usepackage{marvosym}
\usepackage{framed}
\RequirePackage{pdflscape} 

\definecolor{robo_blue}{RGB}{99, 113, 250}
\definecolor{robo_red}{RGB}{239, 99, 75}
\definecolor{robo_green}{RGB}{0, 180, 139}

\title{
The \textcolor{robo_blue}{Robo}\textcolor{robo_red}{Depth} Challenge: Methods and Advancements Towards Robust Depth Estimation
}

\author{%
  Lingdong Kong$^{1,2,\color{robo_green}\spadesuit}$\quad Yaru Niu$^{3,\color{robo_green}\spadesuit}$\quad Shaoyuan Xie$^{4,5,\color{robo_green}\spadesuit}$\quad Hanjiang Hu$^{3,\color{robo_green}\spadesuit}$\quad Lai Xing Ng$^{6,\color{robo_green}\spadesuit}$\\
  \textbf{Benoit R. Cottereau}$^{7,8,\color{robo_green}\spadesuit}$\quad \textbf{Liangjun Zhang}$^{9,\color{robo_green}\spadesuit}$\quad \textbf{Hesheng Wang}$^{10,\color{robo_green}\spadesuit}$\quad
  \textbf{Wei Tsang Ooi}$^{1,\color{robo_green}\spadesuit}$\\ \textbf{Ruijie Zhu}$^{11}$\quad \textbf{Ziyang Song}$^{11}$\quad \textbf{Li Liu}$^{11}$\quad \textbf{Tianzhu Zhang}$^{11,12}$\quad \textbf{Jun Yu}$^{11}$\\ \textbf{Mohan Jing}$^{11}$\quad \textbf{Pengwei Li}$^{11}$\quad \textbf{Xiaohua Qi}$^{11}$\quad \textbf{Cheng Jin}$^{13}$\quad \textbf{Yingfeng Chen}$^{13}$\\
  \textbf{Jie Hou}$^{13}$\quad \textbf{Jie Zhang}$^{14}$\quad \textbf{Zhen Kan}$^{11}$\quad \textbf{Qiang Ling}$^{11}$\quad \textbf{Liang Peng}$^{15}$\quad \textbf{Minglei Li}$^{15}$\\
  \textbf{Di Xu}$^{15}$\quad \textbf{Changpeng Yang}$^{15}$\quad \textbf{Yuanqi Yao}$^{16}$\quad \textbf{Gang Wu}$^{16}$\quad \textbf{Jian Kuai}$^{16}$\\
  \textbf{Xianming Liu}$^{16}$\quad \textbf{Junjun Jiang}$^{16}$\quad \textbf{Jiamian Huang}$^{17}$\quad \textbf{Baojun Li}$^{17}$\quad \textbf{Jiale Chen}$^{18}$\\
  \textbf{Shuang Zhang}$^{18}$\quad \textbf{Sun Ao}$^{16}$\quad \textbf{Zhenyu Li}$^{16}$\quad \textbf{Runze Chen}$^{19,20}$\quad \textbf{Haiyong Luo}$^{19}$\\
  \textbf{Fang Zhao}$^{20}$\quad \textbf{Jingze Yu}$^{19,20}$
  \\\\
  $^{\color{robo_green}\spadesuit}$The Organizing Team\quad $^1$National University of Singapore\quad $^2$CNRS@CREATE\\
  $^3$Carnegie Mellon University\quad $^4$Huazhong University of Science and Technology\\ $^5$University of California, Irvine\quad $^6$Institute for Infocomm Research, A*STAR\\
  $^7$IPAL, CNRS IRL 2955, Singapore\quad $^8$CerCo, CNRS UMR 5549, Université Toulouse III\\ $^9$Baidu Research\quad $^{10}$Shanghai Jiao Tong University\\
  $^{11}$University of Science and Technology of China\\
  $^{12}$Deep Space Exploration Lab\quad $^{13}$NetEase Fuxi\quad $^{14}$Central South University\\
  $^{15}$Huawei Cloud Computing Technology\quad 
  $^{16}$Harbin Institute of Technology\\
  $^{17}$Individual Researcher\quad $^{18}$Tsinghua University\\
  $^{19}$Beijing University of Posts and Telecommunications\\
  $^{20}$Institute of Computing Technology, Chinese Academy of Sciences
  \\
}

\begin{document}
\maketitle

\begin{abstract}
  Accurate depth estimation under out-of-distribution (OoD) scenarios, such as adverse weather conditions, sensor failure, and noise contamination, is desirable for safety-critical applications. Existing depth estimation systems, however, suffer inevitably from real-world corruptions and perturbations and are struggled to provide reliable depth predictions under such cases. In this paper, we summarize the winning solutions from the RoboDepth Challenge -- an academic competition designed to facilitate and advance robust OoD depth estimation. This challenge was developed based on the newly established KITTI-C and NYUDepth2-C benchmarks. We hosted two stand-alone tracks, with an emphasis on robust self-supervised and robust fully-supervised depth estimation, respectively. Out of more than two hundred participants, nine unique and top-performing solutions have appeared, with novel designs ranging from the following aspects: spatial- and frequency-domain augmentations, masked image modeling, image restoration and super-resolution, adversarial training, diffusion-based noise suppression, vision-language pre-training, learned model ensembling, and hierarchical feature enhancement. Extensive experimental analyses along with insightful observations are drawn to better understand the rationale behind each design. We hope this challenge could lay a solid foundation for future research on robust and reliable depth estimation and beyond. The datasets, competition toolkit, workshop recordings, and source code from the winning teams are publicly available on the challenge website\footnote{The RoboDepth Challenge: \url{https://robodepth.github.io}.}.
\end{abstract}

\section{Introduction}
\label{sec:introduction}

The robustness of a learning-based visual perception system is among the most important factors that practitioners pursue \cite{wang2023survey_robustness}. In the context of depth estimation, the robustness of a depth prediction algorithm is often coped with its ability to maintain satisfactory performance under perturbation and degradation. Indeed, since most depth estimation systems target estimating structural information from real-world scenes \cite{silberman2012nyu2,geiger2012kitti,cordts2016cityscapes}, it is inevitable for them to deal with unseen data that are distribution-shifted from those seen during training. 

\begin{figure}[t]
    \centering
    \includegraphics[width=1.0\linewidth]{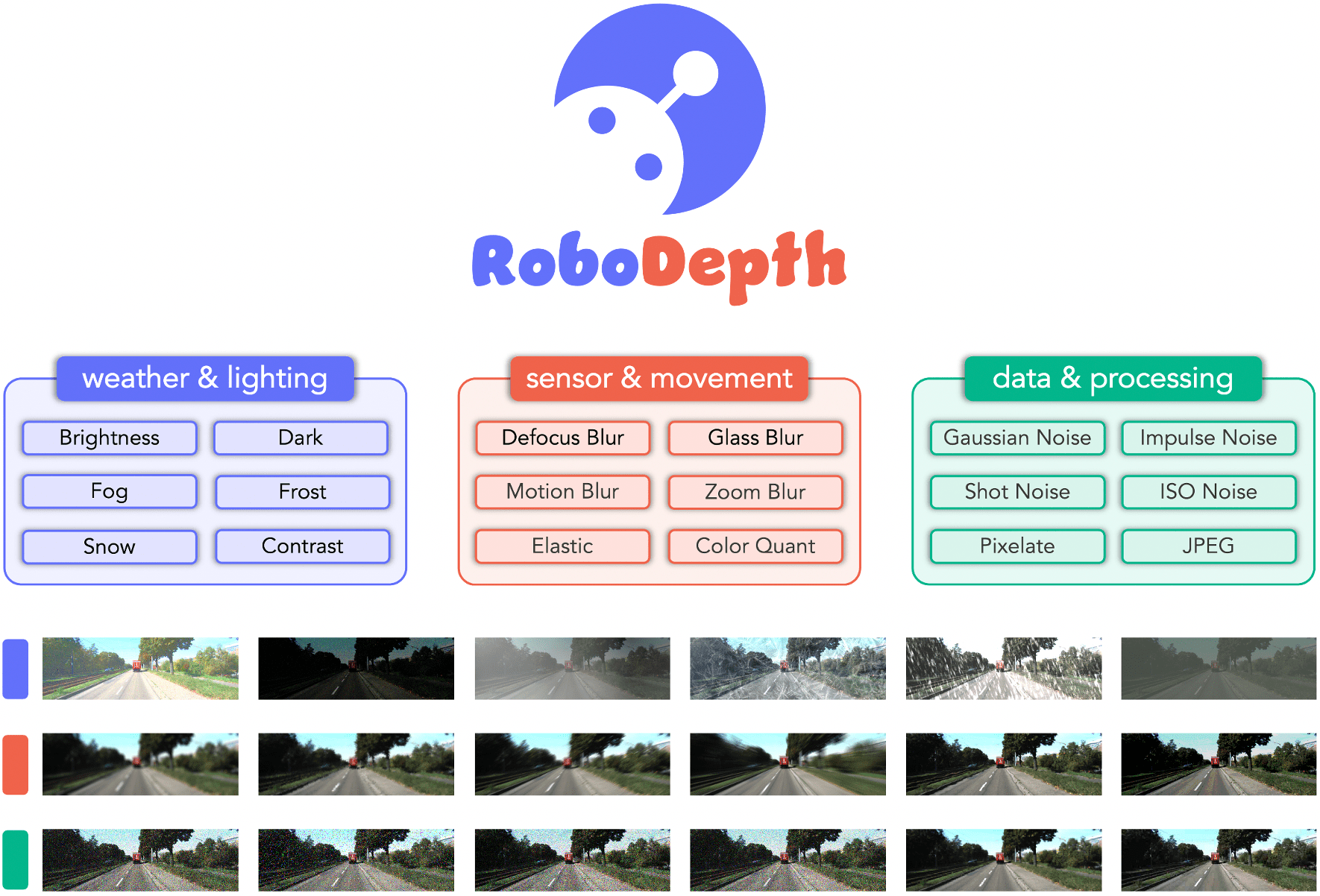}
    \caption{The RoboDepth Challenge adopts the eighteen data corruption types from three main categories defined in the RoboDepth benchmark \cite{kong2023robodepth_benchmark}. Examples shown are from the \textit{KITTI-C} dataset.}
\label{fig:taxonomy}
\end{figure}

Data distribution shifts often take various forms, such as adversarial attack \cite{cheng2022physical,duan2021advdrop,xie2023adv} and common corruptions \cite{ImageNet-C,geirhos2018imagenet-trained,kar20223d}. While the former aims to trick learning-based models by providing deceptive input, the latter cases -- which are caused by noises, blurs, illumination changes, perspective transformations, \textit{etc}. -- are more inclined to occur in practice. Recently, the RoboDepth benchmark \cite{kong2023robodepth_benchmark} established the first comprehensive study on the out-of-distribution (OoD) robustness of monocular depth estimation models under common corruptions. Specifically, a total of eighteen corruption types are defined, ranging from three main categories: 1) adverse weather and lighting conditions, 2) motion and sensor failure, and 3) noises during data processing. Following the taxonomy, two robustness probing datasets are constructed by simulating realistic data corruptions on images from the KITTI \cite{geiger2012kitti} and NYU Depth V2 \cite{silberman2012nyu2} datasets, respectively. More than forty depth estimation models are benchmarked and analyzed. The results show that existing depth estimation algorithms, albeit achieved promising performance on ``clean'' benchmarks, are at risk of being vulnerable to common corruptions. This study also showcases the importance of considering scenarios that are both in-distribution and OoD, especially for safety-critical applications.

The RoboDepth Challenge has been successfully hosted at the 40th IEEE Conference on Robotics and Automation (ICRA 2023), London, UK. This academic competition aims to facilitate and advance robust monocular depth estimation under OoD corruptions and perturbations. Specifically, based on the newly established \textit{KITTI-C} and \textit{NYUDepth2-C} benchmarks \cite{kong2023robodepth_benchmark}, this competition provides a venue for researchers from both industry and academia to explore novel ideas on: 1) designing network structures that are robust against OoD corruptions, 2) proposing operations and techniques that improve the generalizability of existing depth estimation algorithms, and 3) rethinking potential detrimental components from data corruptions occur under depth estimation scenarios. We formed two stand-alone tracks: one focused on robust self-supervised depth estimation from outdoor scenes and another focused on robust fully-supervised depth estimation from indoor scenes. The evaluation servers of these two tracks were built upon the CodaLab platform \cite{Codalab}. To ensure fair evaluations, we set the following rules and required all participants to obey during this challenge:
\begin{itemize}
    \item All participants must follow the exact same data configuration when training and evaluating their depth estimation algorithms. The use of public or private datasets other than those specified for model training is prohibited.
    \item Since the theme of this challenge is to probe the out-of-distribution robustness of depth estimation models, any use of the eighteen corruption types designed in the RoboDepth benchmark \cite{kong2023robodepth_benchmark} is strictly prohibited, including any atomic operation that is comprising any one of the mentioned corruptions.
    \item To ensure the above rules are followed, each participant was requested to submit the code with reproducible results; the code was for examination purposes only and we manually verified the training and evaluation of each participant's model.
\end{itemize}

We are glad to have more than two hundred teams registered on the challenge servers. Among them, $66$ teams made a total of $1137$ valid submissions; $684$ attempts are from the first track, while the remaining $453$ attempts are from the second track. More detailed statistics are included in Section~\ref{sec:challenge_summary}. In this report, we present solutions from nine teams that have achieved top performance in this challenge. Our participants proposed novel network structures and pre-processing and post-processing techniques, ranging from the following topics:
\begin{itemize}
    \item \textit{Spatial- and frequency-domain augmentations}: Observing that the common data corruptions like blurs and noises contain distinct representations in both spatial and frequency domains \cite{li2022uncertainty,chen2021apr}, new data augmentation techniques are proposed to enhance the feature learning.
    \item \textit{Masked image modeling}: The masking-based image reconstruction approach \cite{he2022mae} exhibits potential for improving OoD robustness; this simple operation encourages the model to learn more robust representations by decoding masked signals from remaining ones.
    \item \textit{Image restoration and super-resolution}: The off-the-shelf restoration and super-resolution networks \cite{zamir2022restormer,liang2021swin-ir,chen2021ipt} can be leveraged to handle degradation during the test time, such as noise contamination,  illumination changes, and image compression.
    \item \textit{Adversarial training}: The joint adversarial objectives \cite{madry2018adversarial} between the depth estimation and a noise generator facilitate robust feature learning; such an approach also maintains the performance on in-distribution scenarios while tackling OoD cases.
    \item \textit{Diffusion-based noise suppression}: The denoising capability of diffusion is naturally suitable for handling OoD situations \cite{rombach2022stable}; direct use of the denoising step in the pre-trained diffusion model could help suppress the noises introduced by different data corruptions. 
    \item \textit{Vision-language pre-training}: Leveraging the pre-trained text features \cite{zhao2023unleashing} and aligning them to the extracted image features via an adapter is popular among recent studies and is proven helpful to improve the performance of various visual perception tasks \cite{chen2023clip2Scene,chen2023towards}.
    \item \textit{Learned model ensembling}: The fusion among multiple models is commonly used in academic competitions; an efficient, proper, and simple model ensembling strategy often combines the advantages of different models and largely improves the performance.
    \item \textit{Hierarchical feature enhancement}: Designing network architectures that are robust against common corruptions is of great value; it has been constantly verified that the CNN-Transformer hybrid structures \cite{zhao2021monovit,zhang2023litemono} are superior in handling OoD corruptions. 
\end{itemize}

The remainder of this paper is organized as follows: Section~\ref{sec:related-work} reviews recent advancements in depth estimation and out-of-distribution perception and summarizes relevant challenges and competitions. Section~\ref{sec:challenge_summary} elaborates on the key statistics, public resources, and terms and conditions of this challenge. Section~\ref{sec:challenge_results} provides the notable results from our participants that are better than the baselines. The detailed solutions of top-performing teams from the first track and the second track of this challenge are presented in Section~\ref{sec:track1} and Section~\ref{sec:track2}, respectively. Section~\ref{sec:conclusion} draws concluding remarks and points out some future directions. Section~\ref{sec:acknowledgements} and Section~\ref{sec:appendix} are acknowledgments and appendix.

\section{Related Work}
\label{sec:related-work}

\subsection{Depth Estimation}
As opposed to some 3D perception tasks that rely on the LiDAR sensor, \textit{e.g.} LiDAR segmentation \cite{behley2019semanticKITTI,caesar2020nuScenes,kong2023conDA,kong2023rethinking,liu2023segment} and 3D object detection \cite{lang2019pointpillars,second,centerpoint,kong2023robo3d}, monocular depth estimation aims to predict 3D structural information from a single image, which is a more affordable solution in existing perception systems. Based on the source of supervision signals, this task can be further categorized into supervised \cite{silberman2012nyu2,bhat2021adabins}, self-supervised \cite{geiger2012kitti,godard2017unsupervised}, and semi-supervised \cite{kuznietsov2017semi,ji2019semi} depth estimation. Ever since the seminar works \cite{eigen2014depth,garg2016unsupervised,zhou2017sfm,godard2019monodepth2,lee2019big} in this topic, a diverse range of ideas has been proposed, including new designs on network architectures \cite{ranftl2021dpt,zhao2021monovit,zhang2023litemono,johnston2020self,li2022depthformer,li2022binsformer}, optimization functions \cite{zhang2022dynadepth,chen2023tridepth,xie2023revealing,schellevis2019maskocc}, internal feature constraints \cite{yan2021cadepth,zhou2021diffnet,ning2023ait,xue2020dnet}, semantic-aided learning \cite{wang2020sdc,jung2021fsre,li2022simipu}, geometry constraint \cite{watson2021manydepth,tosi2019learning}, mixing-source of depth supervisions \cite{ranftl2022towards,sun2022scdepthv3,li2018megadepth}, and unsupervised model pre-training \cite{caron2021dino,oquab2023dinov2}. Following the conventional ``training-testing'' paradigm, current depth estimation methods are often trained and tested on datasets within similar distributions, while neglecting the natural corruptions that commonly occur in real-world situations. This challenge aims to fill this gap: we introduce the first academic competition for robust out-of-distribution (OoD) depth estimation under corruptions. By shedding light on this new perspective of depth estimation, we hope this challenge could enlighten follow-up research in designing novel network architectures and techniques that improve the reliability of depth estimation systems to meet safety-critical requirements.

\subsection{Out-of-Distribution Perception}
The ability to be generalized across unseen domains and scenarios is crucial for a learning-based system \cite{wang2023survey_robustness}. To pursue superior OoD performance under commonly occurring data corruptions, various benchmarks affiliated with different perception tasks have been established. ImageNet-C \cite{ImageNet-C} represented the first attempt at OoD image classification; the proposed corruption types, such as blurs, illumination changes, perspective transformations, and noise contamination, have been widely adopted by following works in OoD dataset construction. Michaelis \textit{et al.} \cite{michaelis2019dragon} built the large-scale Robust Detection Benchmark upon PASCAL VOC \cite{VOC}, COCO \cite{COCO}, and Cityscapes \cite{cordts2016cityscapes} for OoD object detection. Subsequent works adopt a similar paradigm in benchmarking and analyzing OoD semantic segmentation \cite{Cityscapes-C}, video classification \cite{Kinetics-C}, pose estimation \cite{AdvMix}, point cloud perception \cite{ren2022modelnet-c,PointCloud-C}, LiDAR perception \cite{kong2022laserMix,kong2023robo3d,fong2022panoptic-nuScenes,liu2023segment_any_point_cloud}, bird's eye view perception \cite{xie2023robobev,xie2023robobev_codebase}, and robot navigation \cite{RobustNav}. All the above works have incorporated task-specific corruptions that mimic real-world situations, facilitating the development of robust algorithms for their corresponding tasks. To achieve a similar goal, in this challenge, we resort to the newly-established \textit{KITTI-C} and \textit{NYUDepth2-C} benchmarks \cite{kong2023robodepth_benchmark} to construct our OoD depth estimation datasets. We form two stand-alone tracks, with an emphasis on robust self-supervised and robust fully-supervised depth estimation, respectively, to encourage novel designs for robust and reliable OoD depth estimation.

\subsection{Relevant Competitions}
It is worth mentioning that several previous depth estimation competitions have been successfully held to facilitate their related research areas. The Robust Vision Challenge (RVC) \cite{RVC} aimed to explore cross-domain visual perception across different scene understanding tasks, including reconstruction, optical flow estimation, semantic segmentation, single image depth prediction, \textit{etc}. The Dense Depth for Autonomous Driving (DDAD) Challenge \cite{DDAD} targeted long-range and dense depth estimation from diverse urban conditions. The Mobile AI Challenge \cite{MobileAI} focused on real-time depth estimation on smartphones and IoT platforms. The SeasonDepth Depth Prediction Challenge \cite{hu2022seasondepth} was specialized for estimating accurate depth information of scenes under different illumination and season conditions. The Monocular Depth Estimation Challenge (MDEC) \cite{MDEC,MDEC2} attracted broad attention from researchers and was tailored to tackle monocular depth estimation from complex natural environments, such as forests and fields. The Argoverse Stereo Competition \cite{ArgoverseStereo} encouraged real-time stereo depth estimation under self-driving scenarios. The NTIRE 2023 Challenge on HR Depth from Images of Specular and Transparent Surfaces \cite{NTIRE} mainly aimed at handling depth estimation of non-Lambertian surfaces characterizing specular and transparent materials. Different from previous pursuits, our RoboDepth Challenge is tailored to facilitate robust OoD depth estimation against real-world corruptions. A total of eighteen corruption types are considered, ranging from adverse weather conditions, sensor failure, and noise contamination. We believe this research topic is of great importance to the practical deployment of depth estimation algorithms, especially for safety-critical applications.

\section{Challenge Summary}
\label{sec:challenge_summary}

\subsection{Overall Statistics}
This is the first edition of the RoboDepth Challenge. The official evaluation servers\footnote{We built servers on CodaLab. More details of this platform are at \url{https://codalab.lisn.upsaclay.fr}.} of this competition were launched on 01 January 2023. During the five-month period of competition, $226$ teams registered on our servers; among them, $66$ teams attempted to make submissions. Finally, we received $1137$ valid submissions and selected six winning teams (three teams for each track) and three innovation prize awardees. The detailed information of the winning teams and innovation prize awardees is shown in Table~\ref{tab:summary} and Table~\ref{tab:summary_innovation}, respectively.

\begin{figure}[t]
    \centering
    \includegraphics[width=0.999\linewidth]{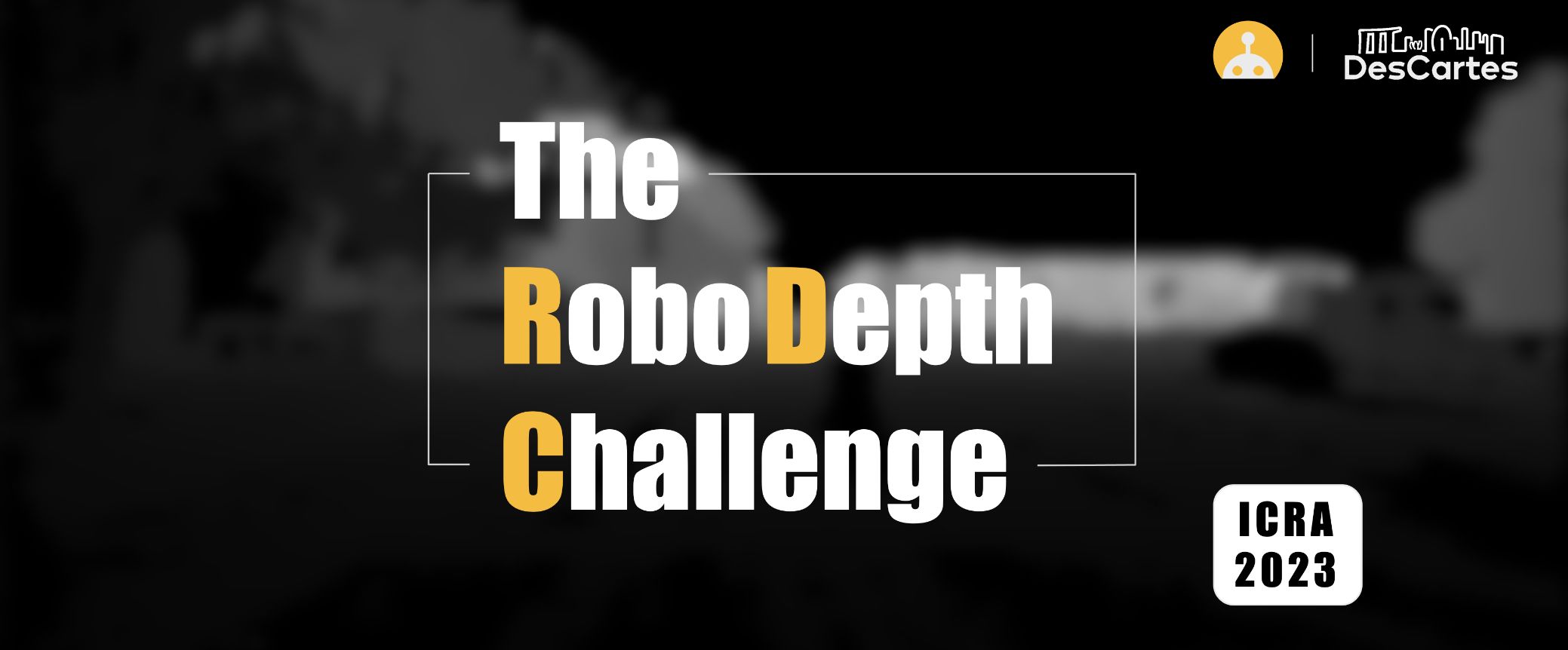}
    \caption{We successfully hosted the RoboDepth Challenge at ICRA 2023.}
\label{fig:competition}
\end{figure}

\begin{figure}
    \centering
    \begin{subfigure}[b]{0.495\textwidth}
         \centering
         \includegraphics[width=\textwidth]{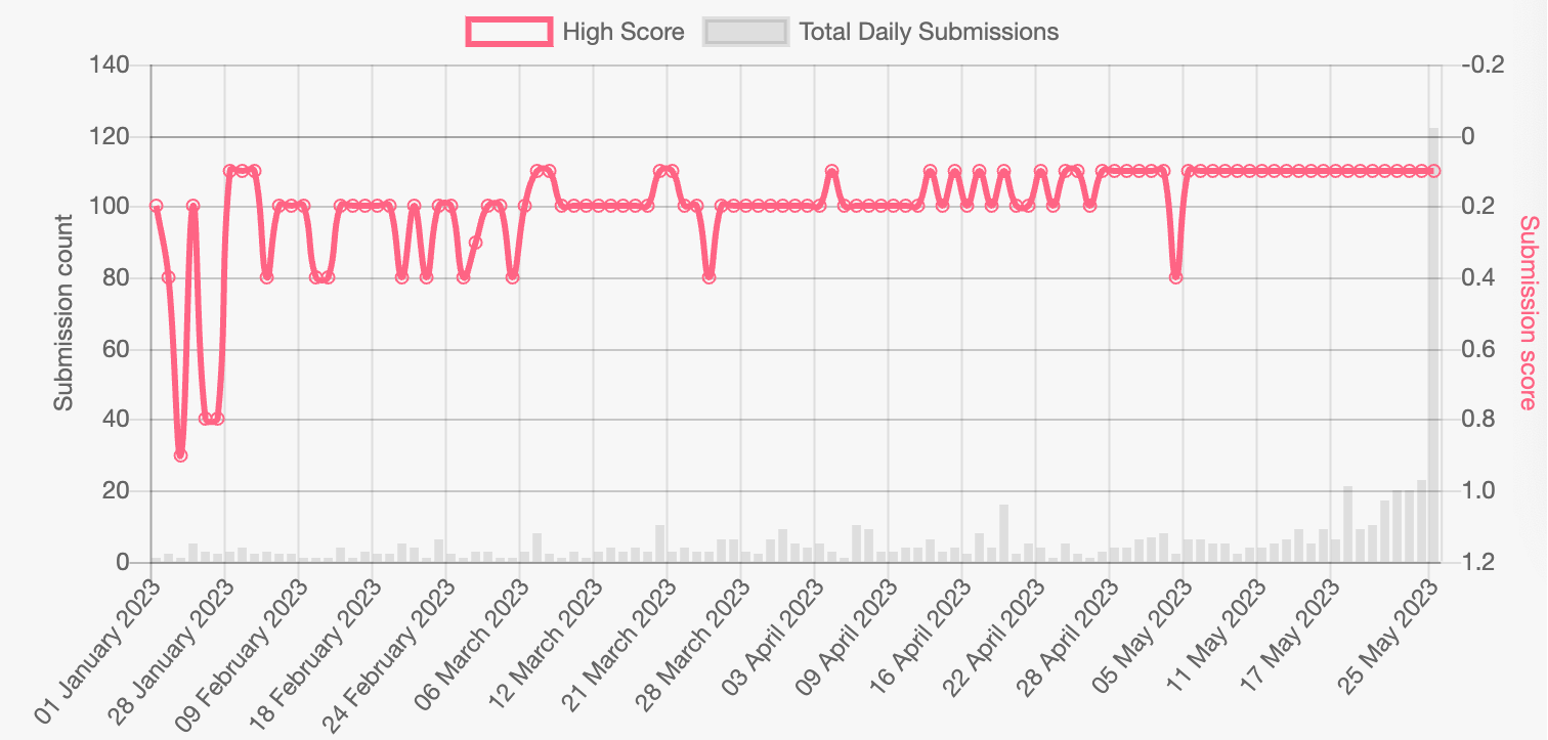}
         \caption{Track \# 1}
    \end{subfigure}
    \hfill
    \begin{subfigure}[b]{0.485\textwidth}
         \centering
         \includegraphics[width=\textwidth]{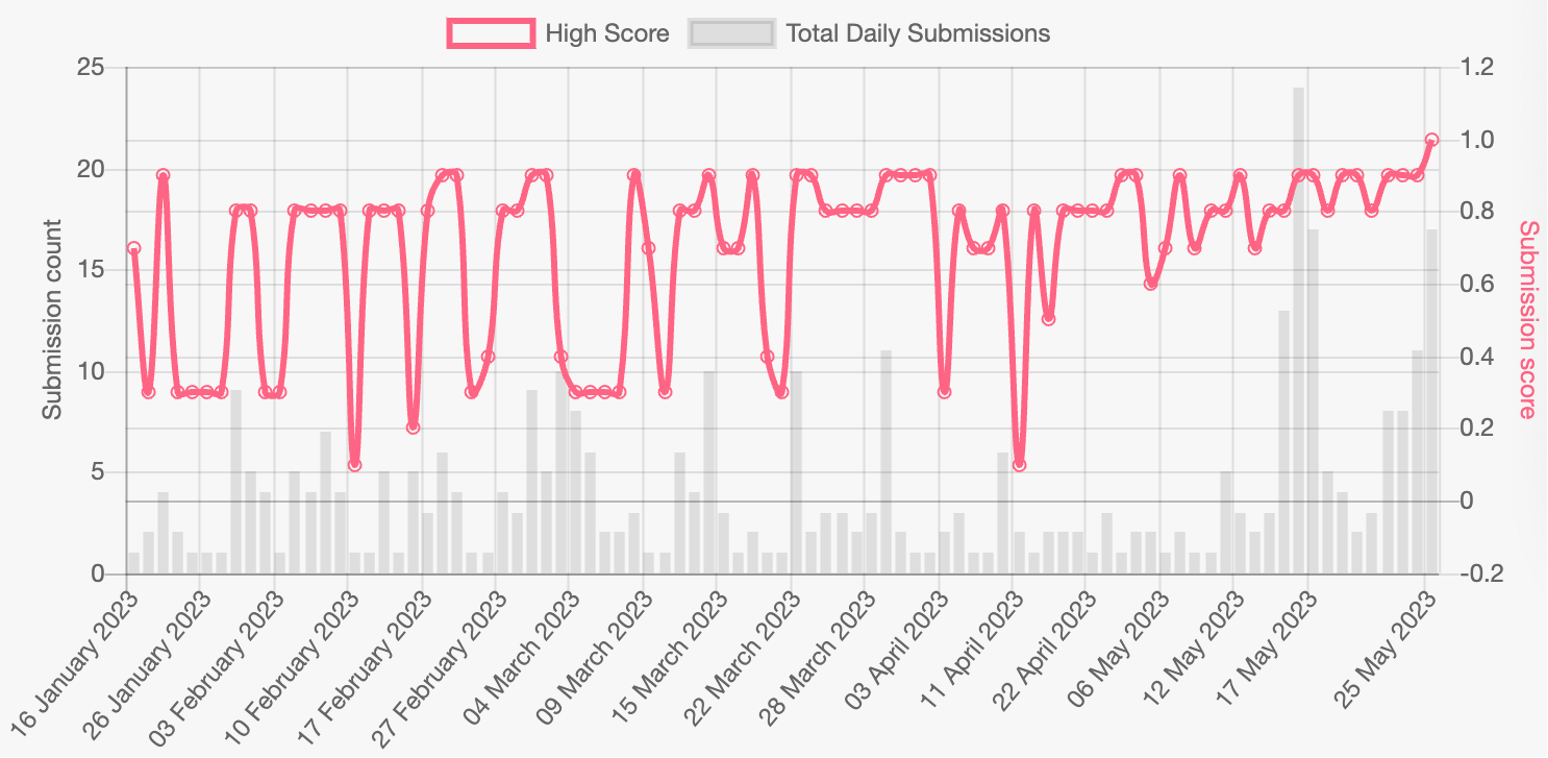}
         \caption{Track \# 2}
    \end{subfigure}
    \caption{The submission and scoring statistics for the two tracks in the RoboDepth Challenge.}
\label{fig:competition_stats}
\end{figure}

\subsection{Track \# 1: Robust Self-Supervised Depth Estimation}
\noindent\textbf{Evaluation Server}. The first track of the RoboDepth Challenge was hosted at \url{https://codalab.lisn.upsaclay.fr/competitions/9418}. The participants were requested to submit their depth disparity maps to our server for evaluation. Such depth predictions were expected to be generated by a learning-based model, in a self-supervised learning manner, trained on the official \textit{training} split of the KITTI dataset \cite{geiger2012kitti}.

\noindent\textbf{Statistics}. In the first track of the RoboDepth Challenge, a total number of $137$ teams registered at our evaluation server. We received $684$ valid submissions during the competition period. The top-three best-performing teams are \texttt{OpenSpaceAI}, \texttt{USTC-IAT-United}, and \texttt{YYQ}. Additionally, we selected the teams \texttt{Ensemble} and \texttt{Scent-Depth} as the innovation prize awardees of this track.

\subsection{Track \# 2: Robust Supervised Depth Estimation}
\noindent\textbf{Evaluation Server}. The second track of the RoboDepth Challenge was hosted at \url{https://codalab.lisn.upsaclay.fr/competitions/9821}. The participants were requested to submit their depth disparity maps to our server for evaluation. Such depth predictions were expected to be generated by a learning-based model, in a fully-supervised learning manner, trained on the official \textit{training} split of the NYUDepth V2 dataset \cite{silberman2012nyu2}.

\noindent\textbf{Statistics}. In the second track of the RoboDepth Challenge, a total number of $89$ teams registered at our evaluation server. We received $453$ valid submissions during the competition period. The top-three best-performing teams are \texttt{USTCxNetEaseFuxi}, \texttt{OpenSpaceAI}, and \texttt{GANCV}. Additionally, we selected the team \texttt{AIIA-RDepth} as the innovation prize awardee of this track.

\begin{table*}[t]
\caption{Summary of the top-performing teams in each track of the RoboDepth Challenge.}
\centering\scalebox{1}{
\begin{tabular}{c|p{5cm}|p{5cm}}
\toprule
\textbf{Rank} & \textbf{\#1: Robust Self-Supervised MDE} & \textbf{\#2: Robust Supervised MDE}
\\\midrule\midrule
\multirow{13}{*}{\textcolor{robo_blue}{\textbf{1st Place}}} & \textbf{Team Name} & \textbf{Team Name}
\\
& \textcolor{robo_blue}{OpenSpaceAI} & \textcolor{robo_blue}{USTCxNetEaseFuxi}
\\
\cmidrule{2-3}
& \textbf{Team Members} & \textbf{Team Members}
\\
& Ruijie Zhu$^1$, Ziyang Song$^1$, Li Liu$^1$, Tianzhu Zhang$^{1,2}$ & Jun Yu$^1$, Mohan Jing$^1$, Pengwei Li$^1$, Xiaohua Qi$^1$, Cheng Jin$^2$, Yingfeng Chen$^2$, Jie Hou$^2$
\\
\cmidrule{2-3}
& \textbf{Affiliations} & \textbf{Affiliations}
\\
& $^1$University of Science and Technology of China, $^2$Deep Space Exploration Lab & $^1$University of Science and Technology of China, $^2$NetEase Fuxi
\\\cmidrule{2-3}
& \textbf{Contact} $\textrm{\Letter}$ & \textbf{Contact} $\textrm{\Letter}$
\\
& \texttt{ruijiezhu@mail.ustc.edu.cn} & \texttt{USTC\_IAT\_United@163.com}
\\\midrule\midrule
\multirow{17}{*}{\textcolor{robo_red}{\textbf{2nd Place}}} & \textbf{Team Name} & \textbf{Team Name}
\\
& \textcolor{robo_red}{USTC-IAT-United} & \textcolor{robo_red}{OpenSpaceAI}
\\
\cmidrule{2-3}
& \textbf{Team Members} & \textbf{Team Members}
\\
& Jun Yu$^1$, Xiaohua Qi$^1$, Jie Zhang$^2$, Mohan Jing$^1$, Pengwei Li$^1$, Zhen Kan$^1$, Qiang Ling$^1$, Liang Peng$^3$, Minglei Li$^3$, Di Xu$^3$, Changpeng Yang$^3$ & Li Liu$^1$, Ruijie Zhu$^1$, Ziyang Song$^1$, Tianzhu Zhang$^{1,2}$
\\
\cmidrule{2-3}
& \textbf{Affiliations} & \textbf{Affiliations}
\\
& $^1$University of Science and Technology of China, $^2$Central South University, $^3$Huawei Cloud Computing Technology Co., Ltd & $^1$University of Science and Technology of China, $^2$Deep Space Exploration Lab
\\
\cmidrule{2-3}
& \textbf{Contact} $\textrm{\Letter}$ & \textbf{Contact} $\textrm{\Letter}$
\\
& \texttt{USTC\_IAT\_United@163.com} & \texttt{liu\_li@mail.ustc.edu.cn}
\\\midrule\midrule
\multirow{11}{*}{\textcolor{robo_green}{\textbf{3rd Place}}} & \textbf{Team Name} & \textbf{Team Name}
\\
& \textcolor{robo_green}{YYQ} & \textcolor{robo_green}{GANCV}
\\
\cmidrule{2-3}
& \textbf{Team Members} & \textbf{Team Members}
\\
& Yuanqi Yao$^1$, Gang Wu$^1$, Jian Kuai$^1$, Xianming Liu$^1$, Junjun Jiang$^1$ & Jiamian Huang$^1$, Baojun Li$^1$
\\
\cmidrule{2-3}
& \textbf{Affiliations} & \textbf{Affiliations}
\\
& $^1$Harbin Institute of Technology & $^1$Individual Researcher
\\
\cmidrule{2-3}
& \textbf{Contact} $\textrm{\Letter}$ & \textbf{Contact} $\textrm{\Letter}$
\\
& \texttt{yuanqiyao@stu.hit.edu.cn} & \texttt{huang176368745@gmail.com}
\\\bottomrule
\end{tabular}
}
\label{tab:summary}
\end{table*}
\begin{table*}[t]
\caption{Summary of innovation prize awardees (across two tracks) in the RoboDepth Challenge.}
\centering\scalebox{1}{
\begin{tabular}{p{4.1cm}|p{4.1cm}|p{4.1cm}}
\toprule
\textbf{Team 1} & \textbf{Team 2} & \textbf{Team 3}
\\\midrule\midrule
\textbf{Team Name} & \textbf{Team Name} & \textbf{Team Name}
\\
\textcolor{robo_blue}{Ensemble} & \textcolor{robo_red}{Scent-Depth} & \textcolor{robo_green}{AIIA-RDepth}
\\\midrule
\textbf{Track} & \textbf{Track} & \textbf{Track}
\\
\#1: Robust Self-Supervised MDE & \#1: Robust Self-Supervised MDE & \#2: Robust Supervised MDE
\\\midrule
\textbf{Team Members} & \textbf{Team Members} & \textbf{Team Members}
\\
Jiale Chen$^1$, Shuang Zhang$^1$ & Runze Chen$^{1,2}$, Haiyong Luo$^1$, Fang Zhao$^2$, Jingze Yu$^{1,2}$ & Sun Ao$^1$, Gang Wu$^1$, Zhenyu Li$^1$, Xianming Liu$^1$, Junjun Jiang$^1$
\\\midrule
\textbf{Affiliations} & \textbf{Affiliations} & \textbf{Affiliations}
\\
$^1$Tsinghua University & $^1$Beijing University of Posts and Telecommunications, $^2$Institute of Computing Technology, Chinese Academy of Sciences & $^1$Harbin Institute of Technology
\\\midrule
\textbf{Contact} $\textrm{\Letter}$ & \textbf{Contact} $\textrm{\Letter}$ & \textbf{Contact} $\textrm{\Letter}$
\\
cjl20@mails.tsinghua.edu.cn & chenrz925@bupt.edu.cn & sunao\_sz@163.com
\\\bottomrule
\end{tabular}
}
\label{tab:summary_innovation}
\end{table*}

\subsection{The RoboDepth Workshop}
We hosted the online workshop at ICRA 2023 on 02 June 2023 after the competition was officially concluded. Six winning teams and three innovation prize awardees attended and presented their approaches.

The video recordings of this workshop are publicly available at \url{https://www.youtube.com/watch?v=mYhdTGiIGCY&list=PLxxrIfcH-qBGZ6x_e1AT2_YnAxiHIKtkB}.

The slides used can be downloaded from \url{https://ldkong.com/talks/icra23_robodepth.pdf}.

\subsection{Terms and Conditions}
The RoboDepth Challenge is made freely available to academic and non-academic entities for non-commercial purposes such as research, teaching, scientific publications, or personal experimentation. Permission is granted to use the related public resources given that the participants agree:
\begin{itemize}
    \item That the data in this challenge comes “AS IS”, without express or implied warranty. Although every effort has been made to ensure accuracy, the challenge organizing team is not responsible for any errors or omissions.
    \item That the participants may not use the data in this challenge or any derivative work for commercial purposes as, for example, licensing or selling the data, or using the data with the purpose to procure a commercial gain.
    \item That the participants include a reference to RoboDepth (including the benchmark data and the specially generated data for this academic challenge) in any work that makes use of the benchmark. For research papers, please cite our preferred publications as listed on our webpage and GitHub repository.
\end{itemize}
\section{Challenge Results}
\label{sec:challenge_results}

\subsection{Evaluation Metrics}
In the RoboDepth Challenge, the two most conventional metrics were adopted: 1) error rate, including \texttt{Abs Rel}, \texttt{Sq Rel}, \texttt{RMSE}, and \texttt{log RMSE}; and 2) accuracy, including $\delta_1$, $\delta_2$, and $\delta_3$.

\noindent\textbf{Error Rate}.
The Relative Absolute Error (\texttt{Abs Rel}) measures the relative difference between the pixel-wise ground-truth (\texttt{gt}) and the prediction values (\texttt{pred}) in a depth prediction map $D$, as calculated by the following equation:
\begin{equation}
\text{Abs Rel} = \frac{1}{|D|}\sum_{pred\in D}\frac{|gt - pred|}{gt}~.
\end{equation}
The Relative Square Error (\texttt{Sq Rel}) measures the relative square difference between \texttt{gt} and \texttt{pred} as follows:
\begin{equation}
\text{Sq Rel} = \frac{1}{|D|}\sum_{pred\in D}\frac{|gt - pred|^2}{gt}~.
\end{equation}
\texttt{RMSE} denotes the Root Mean Square Error (in meters) of a scene (image), which can be calculated as $\sqrt{\sum|gt - pred|^2}$; while \texttt{log RMSE} is the log-normalized version of \texttt{RMSE}, \textit{i.e.}, $\sqrt{\sum|\log(gt) - \log(pred)|^2}$.

\noindent\textbf{Accuracy}.
The $\delta$ metric is the depth estimation accuracy given the threshold:
\begin{equation}
\delta_t = \frac{1}{|D|}|\{\ pred\in D | \max{(\frac{gt}{pred}, \frac{pred}{gt})< 1.25^t}\}| \times 100\%~,
\end{equation}
where $\delta_1 = \delta<1.25, \delta_2 = \delta<1.25^2, \delta_3 = \delta<1.25^3$ are the three conventionally used accuracy scores among prior works \cite{godard2019monodepth2,lidepthtoolbox2022}.

Following the seminar work MonoDepth2 \cite{godard2019monodepth2}, the \texttt{Abs Rel} metric was selected as the major indicator to compare among submissions in the first track of the RoboDepth Challenge.

Based on the Monocular-Depth-Estimation-Toolbox\footnote{\url{https://github.com/zhyever/Monocular-Depth-Estimation-Toolbox}.}, the $\delta_1$ score was used to rank different submissions in the second track of the RoboDepth Challenge.

\subsection{Track \# 1 Results}

In the first track of the RoboDepth Challenge, we received $684$ valid submissions. The top-performing teams in this track include \texttt{OpenSpaceAI}, \texttt{USTC-IAT-United}, and \texttt{YYQ}. The shortlisted submissions are shown in Table~\ref{tab:track1_results}; the complete results can be found on our evaluation server.

Specifically, the team \texttt{OpenSpaceAI} achieved a \texttt{Abs Rel} score of $0.121$, which is $0.100$ higher than the baseline MonoDepth2 \cite{godard2019monodepth2}. They also ranked first on the \texttt{log RMSE}, $\delta_1$, and $\delta_3$ metrics. Other top-ranked submissions are from: the team \texttt{USTC-IAT-United} (\texttt{Abs Rel}$=0.123$, $\delta_1=0.861$), team \texttt{YYQ} (\texttt{Abs Rel}$=0.123$, $\delta_1=0.848$), team \texttt{zs\_dlut} (\texttt{Abs Rel}$=0.124$, $\delta_1=0.852$), and team \texttt{UMCV} (\texttt{Abs Rel}$=0.124$, $\delta_1=0.847$). We refer readers to the solutions presented in Section~\ref{sec:track1} for additional comparative and ablation results and more detailed analyses.

\begin{table*}[t]
\caption{Leaderboard of Track \# 1 (robust self-supervised depth estimation) in the RoboDepth Challenge. The \textbf{best} and \underline{second best} scores of each metric are highlighted in \textbf{bold} and \underline{underline}, respectively. Only entries better than the baseline are included in this table. In Track \# 1, MonoDepth2 \cite{godard2019monodepth2} was adopted as the baseline. See our evaluation server for the complete results.}
\centering\scalebox{0.78}{
\begin{tabular}{c|c|cccc|cccc}
    \toprule
    \textbf{\#} & \textbf{Team Name} & \textbf{Abs Rel~$\downarrow$} & \textbf{Sq Rel~$\downarrow$} & \textbf{RMSE~$\downarrow$} & \textbf{log RMSE~$\downarrow$} & $\delta<1.25$~$\uparrow$ & $\delta<1.25^2$~$\uparrow$ & $\delta<1.25^3$~$\uparrow$
    \\\midrule\midrule
    \textcolor{robo_red}{\textbf{1}} & \textcolor{robo_red}{OpenSpaceAI} & $\mathbf{0.121}$ & $0.919$ & $4.981$ & $\mathbf{0.200}$	& $\mathbf{0.861}$ & \underline{$0.953$} & $\mathbf{0.980}$
    \\
    \textcolor{robo_blue}{\textbf{2}} & \textcolor{robo_blue}{USTC-IAT-United} & \underline{$0.123$} & $0.932$ & $\mathbf{4.873}$ & $0.202$ & $\mathbf{0.861}$ & $\mathbf{0.954}$ & \underline{$0.979$}
    \\
    \textcolor{robo_green}{\textbf{3}} & \textcolor{robo_green}{YYQ} & \underline{$0.123$} & $0.885$ & $4.983$ & \underline{$0.201$} & $0.848$ & $0.950$ & \underline{$0.979$}
    \\\midrule
    4 & zs\_dlut & $0.124$ & $0.899$ & $4.938$ & $0.203$ & $0.852$ & $0.950$ & \underline{$0.979$}
    \\
    5 & UMCV & $0.124$ & $\mathbf{0.845}$ & \underline{$4.883$} & $0.202$ & $0.847$ & $0.950$ & $\mathbf{0.980}$
    \\
    6 & THU\_ZS & $0.124$ & $0.892$ & $4.928$ & $0.203$ & $0.851$ & $0.951$ & $\mathbf{0.980}$
    \\
    7 & THU\_Chen & $0.125$ & \underline{$0.865$} & $4.924$ & $0.203$ & $0.846$ & $0.950$ & $\mathbf{0.980}$
    \\
    8 & seesee & $0.126$ & $0.990$ & $4.979$ & $0.206$ & $0.857$ & $0.952$ & $0.978$
    \\
    9 & namename & $0.126$ & $0.994$ & $4.950$ & $0.204$ & \underline{$0.860$} & \underline{$0.953$} & \underline{$0.979$}
    \\
    10 & USTCxNetEaseFuxi & $0.129$ & $0.973$ & $5.100$ & $0.208$ & $0.846$ & $0.948$ & $0.978$
    \\
    11 & Tutu & $0.131$ & $0.972$ & $5.085$ & $0.207$ & $0.835$ & $0.946$ & \underline{$0.979$}
    \\
    12 & Cai & $0.133$ & $1.017$ & $5.282$ & $0.214$ & $0.837$ & $0.945$ & $0.976$
    \\
    13 & Suzally & $0.133$ & $1.023$ & $5.285$ & $0.215$ & $0.835$ & $0.943$ & $0.976$
    \\
    14 & waterch & $0.137$ & $0.904$ & $5.276$ & $0.214$ & $0.813$ & $0.941$ & \underline{$0.979$}
    \\
    15 & hust99 & $0.139$ & $1.057$ & $5.302$ & $0.220$ & $0.826$ & $0.939$ & $0.975$
    \\
    16 & panzer & $0.141$ & $0.953$ & $5.429$ & $0.221$ & $0.804$ & $0.936$ & $0.976$
    \\
    17 & lyle & $0.142$ & $0.981$ & $5.590$ & $0.225$ & $0.806$ & $0.936$ & $0.974$
    \\
    18 & SHSCUMT & $0.142$ & $1.064$ & $5.155$ & $0.215$ & $0.821$ & $0.943$ & $0.977$
    \\
    19 & hanchenggong & $0.142$ & $1.064$ & $5.155$ & $0.215$ & $0.821$ & $0.943$ & $0.977$
    \\
    20 & king & $0.160$ & $1.230$ & $5.927$ & $0.244$ & $0.769$ & $0.921$ & $0.966$
    \\
    21 & xujianyao & $0.172$ & $1.340$ & $6.177$ & $0.258$ & $0.743$ & $0.910$ & $0.963$
    \\
    22 & Wenhui\_Wei & $0.172$ & $1.340$ & $6.177$ & $0.258$ & $0.743$ & $0.910$ & $0.963$
    \\
    23 & jerryxu & $0.192$ & $1.594$ & $6.506$ & $0.279$ & $0.709$ & $0.895$ & $0.956$
    \\\midrule
    - & MonoDepth2 \cite{godard2019monodepth2} & $0.221$ & $1.988$ & $7.117$ & $0.312$ & $0.654$ & $0.859$ & $0.938$
\\\bottomrule
\end{tabular}
}
\label{tab:track1_results}
\end{table*}

\subsection{Track \# 2 Results}

In the second track of the RoboDepth Challenge, we received $453$ valid submissions. The top-performing teams in this track include \texttt{USTCxNetEaseFuxi}, \texttt{OpenSpaceAI}, and \texttt{GANCV}. The shortlisted submissions are shown in Table~\ref{tab:track2_results}; the complete results can be found on our evaluation server.

Specifically, the team \texttt{USTCxNetEaseFuxi} achieved a $\delta_1$ score of $0.940$, which is $0.285$ higher than the baseline DepthFormer-SwinT \cite{li2022depthformer}. They also ranked first on the \texttt{Abs Rel} and \texttt{log RMSE} metrics. Other top-ranked submissions are from: the team \texttt{OpenSpaceAI} (\texttt{Abs Rel}$=0.095$, $\delta_1=0.928$), team \texttt{GANCV} (\texttt{Abs Rel}$=0.104$, $\delta_1=0.898$), team \texttt{shinonomei} (\texttt{Abs Rel}$=0.123$, $\delta_1=0.861$), and team \texttt{YYQ} (\texttt{Abs Rel}$=0.125$, $\delta_1=0.851$). We refer readers to the solutions presented in Section~\ref{sec:track2} for additional comparative and ablation results and more detailed analyses.

\section{Winning Solutions from Track \# 1}
\label{sec:track1}

\subsection{The \textcolor{robo_blue}{1st} Place Solution: \textcolor{robo_blue}{OpenSpaceAI}}
\noindent\textbf{Authors:} \textcolor{gray}{Ruijie Zhu, Ziyang Song, Li Liu, and Tianzhu Zhang.}

\begin{framed}
    \textbf{Summary} - Though existing self-supervised monocular depth estimation methods achieved high accuracy on standard benchmarks, few works focused on their OoD generalizability under real-world corruptions. The \texttt{OpenSpaceAI} team proposes IRUDepth to improve the robustness and uncertainty estimation of depth estimation systems. It takes a CNN-Transformer hybrid architecture as the baseline and applies simple yet effective data augmentation chains to enforce consistent depth predictions under diverse corruption scenarios.
\end{framed}

\begin{table*}[t]
\caption{Leaderboard of Track \# 2 (robust supervised depth estimation) in the RoboDepth Challenge. The \textbf{best} and \underline{second best} scores of each metric are highlighted in \textbf{bold} and \underline{underline}, respectively. Only entries better than the baseline are included in this table. In Track \# 2, DepthFormer-SwinT \cite{li2022depthformer} was adopted as the baseline. See our evaluation server for the complete results.}
\label{tab:supp-track2}
\centering\scalebox{0.78}{
\begin{tabular}{c|c|cccc|cccc}
    \toprule
    \textbf{\#} & \textbf{Team Name} & \textbf{Abs Rel~$\downarrow$} & \textbf{Sq Rel~$\downarrow$} & \textbf{RMSE~$\downarrow$} & \textbf{log RMSE~$\downarrow$} & $\delta<1.25$~$\uparrow$ & $\delta<1.25^2$~$\uparrow$ & $\delta<1.25^3$~$\uparrow$
    \\\midrule\midrule
    \textcolor{robo_red}{\textbf{1}} & \textcolor{robo_red}{USTCxNetEaseFuxi} & $\mathbf{0.088}$ & \underline{$0.046$} & \underline{$0.347$} & $\mathbf{0.115}$ & $\mathbf{0.940}$ & \underline{$0.985$} & \underline{$0.996$}
    \\
    \textcolor{robo_blue}{\textbf{2}} & \textcolor{robo_blue}{OpenSpaceAI} & \underline{$0.095$} & $\mathbf{0.045}$ & $\mathbf{0.341}$ & \underline{$0.117$} & \underline{$0.928$} & $\mathbf{0.990}$ & $\mathbf{0.998}$
    \\
    \textcolor{robo_green}{\textbf{3}} & \textcolor{robo_green}{GANCV} & $0.104$ & $0.060$ & $0.391$ & $0.131$ & $0.898$ & $0.982$ & $0.995$
    \\\midrule
    4 & AIIA-RDepth & $0.123$ & $0.080$ & $0.450$ & $0.153$ & $0.861$ & $0.975$ & $0.993$
    \\
    5 & YYQ & $0.125$ & $0.085$ & $0.470$ & $0.159$ & $0.851$ & $0.970$ & $0.989$
    \\
    6 & Hyq & $0.124$ & $0.089$ & $0.474$ & $0.158$ & $0.851$ & $0.967$ & $0.990$
    \\
    7 & DepthSquad & $0.137$ & $0.085$ & $0.462$ & $0.158$ & $0.845$ & $0.976$ & \underline{$0.996$}
    \\
    8 & kinda & $0.146$ & $0.095$ & $0.480$ & $0.165$ & $0.831$ & $0.973$ & $0.993$
    \\
    9 & dx3 & $0.131$ & $0.095$ & $0.507$ & $0.170$ & $0.825$ & $0.963$ & $0.989$
    \\
    10 & uuht & $0.150$ & $0.100$ & $0.492$ & $0.168$ & $0.822$ & $0.973$ & $0.993$
    \\
    11 & myungwoo & $0.147$ & $0.099$ & $0.496$ & $0.168$ & $0.820$ & $0.972$ & $0.994$
    \\
    12 & kamir\_t & $0.134$ & $0.100$ & $0.528$ & $0.176$ & $0.815$ & $0.959$ & $0.986$
    \\
    13 & daicver & $0.137$ & $0.100$ & $0.517$ & $0.175$ & $0.808$ & $0.962$ & $0.988$
    \\
    14 & THUZS & $0.156$ & $0.117$ & $0.555$ & $0.190$ & $0.785$ & $0.953$ & $0.988$
    \\
    15 & fnahua88 & $0.163$ & $0.129$ & $0.579$ & $0.193$ & $0.767$ & $0.952$ & $0.986$
    \\
    16 & wallong & $0.198$ & $0.167$ & $0.624$ & $0.222$ & $0.710$ & $0.927$ & $0.981$
    \\\midrule
    - & DepthFormer~\cite{li2022depthformer} & $0.190$ & $0.179$ & $0.717$ & $0.248$ & $0.655$ & $0.898$ & $0.970$
    \\\bottomrule
    \end{tabular}
}
\label{tab:track2_results}
\end{table*}

\subsubsection{Overview}
Depth estimation is a fundamental task in 3D vision with vital applications, such as autonomous driving \cite{schon2021mgnet}, augmented reality \cite{yucel2021real}, virtual reality \cite{li2021panodepth}, and 3D reconstruction \cite{yang2020mobile3d}. Though many specialized depth sensors, \textit{e.g.} LiDAR and Time-of-Flight (ToF) cameras, can generate accurate raw depth data, they have certain limitations compared to the learning-based monocular depth estimation systems, such as higher hardware cost and limited usage scenarios.

To meet the high requirement of the challenging OoD depth estimation, we propose IRUDepth, a novel framework that focuses on improving the robustness and uncertainty of current self-supervised monocular depth estimation systems. Following MonoViT \cite{zhao2021monovit}, we use MPViT \cite{lee2022mpvit} as the depth encoder, which is a CNN-Transformer hybrid architecture that fuses multi-scale image features. We use PoseNet \cite{xiang2018posecnn} to jointly optimize the camera parameters and predicted depth maps.

To improve the robustness of the self-supervised monocular depth estimation model under OoD situations, we design an image augmentation module and a triplet loss function motivated by AugMix \cite{hendrycks2020augmix}. For the image augmentation module, we utilize stochastic and diverse augmentations to generate random augmented pairs for input images. After predicting the corresponding depth maps, a triplet loss is applied to constrain the Jensen-Shannon divergence between the predicted depth of the clean image and its augmented version.

The proposed IRUDepth ranks first in the first track of the RoboDepth Challenge. Extensive experimental results on the KITTI-C benchmark also demonstrate that IRUDepth significantly outperforms state-of-the-art methods and exhibits satisfactory OoD robustness.

\subsubsection{Technical Approach}

Given an RGB image $I_t\in\mathbb{R}^{H\times W\times 3}$, the IRUDepth framework aims to predict its corresponding depth map $D_t\in\mathbb{R}^{H\times W}$. To improve the robustness and uncertainty estimation, we use random image augmentation inspired by AugMix \cite{hendrycks2020augmix} to generate two augmented views of an image. Then both clean and augmented images are used as the input to train the depth network.

Following MonoViT \cite{zhao2021monovit}, we use the MPViT module \cite{lee2022mpvit} as the encoder of the depth network to extract the local and global context from images. This module is a multi-path CNN-Transformer hybrid architecture. With a disparity head, we generate pixel-aligned depth maps for all input images. The triplet loss is proposed to encourage consistency between the predicted depth maps of the clean and augmented images. 

During training, to obtain supervisory signals from adjacent image frames, we input the adjacent image frames $I_t^\prime=\{I_{t-1}, I_{t+1}\}$ and feed them into the pose estimation network together with $I_t$ to estimate the relative camera pose $T_{t\rightarrow t^\prime}$. The synthesized counterpart to $I_t$ is then generated by:
\begin{equation}
\label{eq:track1_1st_1}
I_{t\rightarrow t^\prime} = T_{t^\prime}<\texttt{proj}(D_t, T_{t\rightarrow t^\prime}, K)>~,
\end{equation}
where the $\texttt{proj}(\cdot)$ operator returns the 2D coordinates when reprojecting the point cloud generated using $D_t$ onto $I_{t^\prime}$; $K$ denotes the camera parameter matrix.

\begin{figure}
    \centering
    \includegraphics[width=1.0\linewidth]{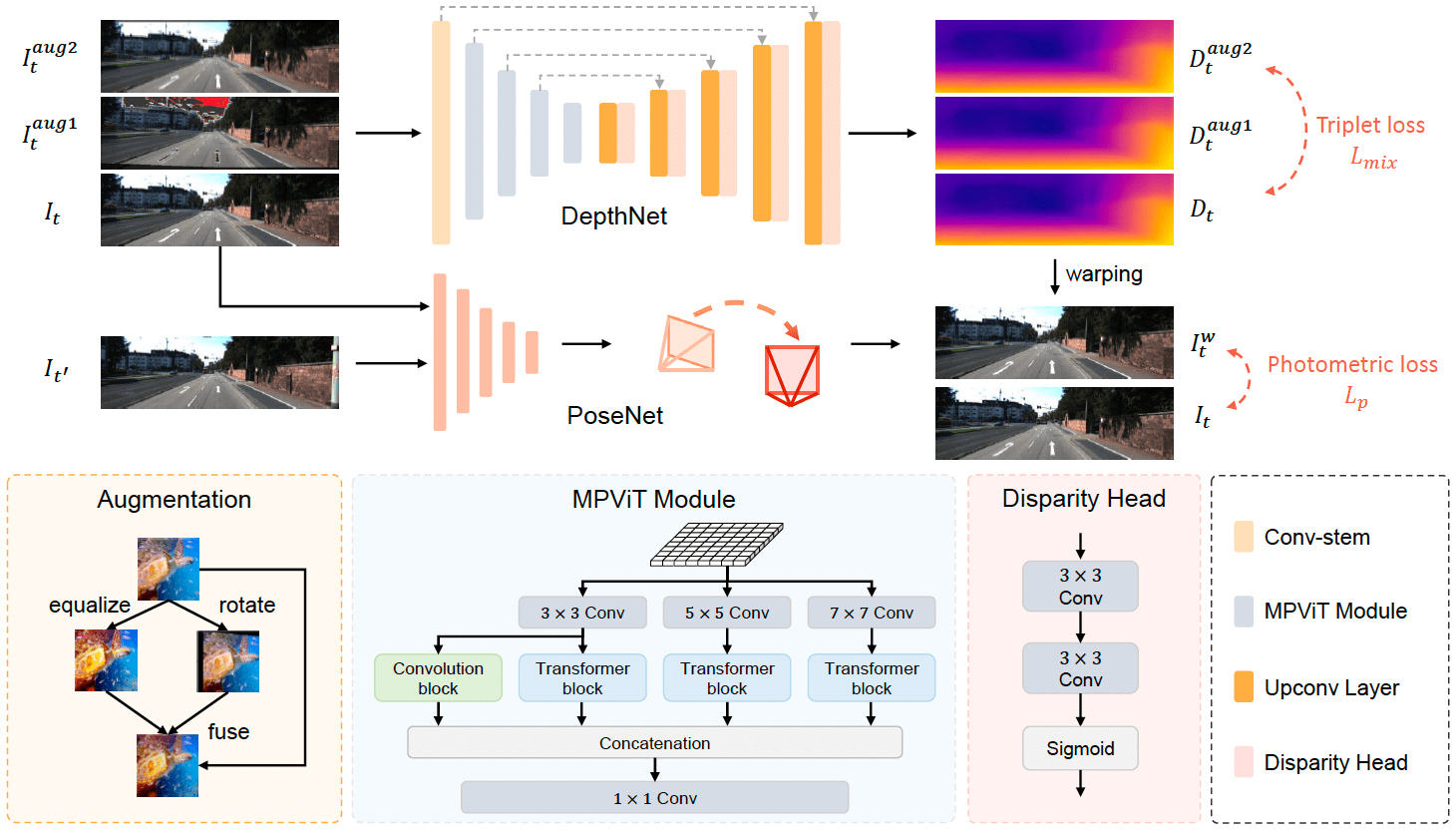}
    \caption{Overview of the IRUDepth framework designed for robust self-supervised depth estimation.}
\end{figure}

\noindent\textbf{Augmentation Module}.
We believe a proper data augmentation technique can significantly improve the generalizability of monocular depth estimation models. Various data augmentation methods \cite{devries2017cutout,yun2019cutmix,zhang2018mixup} have been proposed to enhance the robustness of the model during training. Recently, adding adversarial losses during training has been shown to be an effective way of improving model robustness \cite{madry2018adversarial}. Training with these approaches, however, often greatly increases the training time and GPU memory consumption. It is thus desirable to design a cost-effective augmentation that can be easily plugged into the training pipeline to balance model performance and training consumption. In particular, IRUDepth combines operations from AutoAugment \cite{cubuk2019autoaugment} and AugMix \cite{hendrycks2020augmix} as the main components of our augmentation chain.

\noindent\textbf{Augmentation Protocol}.
To ensure the designed augmentations are disjoint with simulated evaluation data, we exclude operations that constitute or are similar to the $18$ corruption types in KITTI-C. Specifically, we remove the \textit{`contrast'}, \textit{`color'}, \textit{`brightness'}, \textit{`sharpness'}, and \textit{`cutout'} operations from the original augmentation types in \cite{cubuk2019autoaugment,hendrycks2020augmix}. Also, to avoid any potential overlap with the KITTI-C testing set, we do not use any image noising or image blurring operations.

\noindent\textbf{Augmentation Chain}. We randomly sample $k=3$ augmentation chains to combine different augmentation operations. Following AugMix \cite{hendrycks2020augmix}, we mix the resulting images from these augmentation chains via element-wise convex combinations. In particular, we sample convex coefficients from a Dirichlet distribution for the first stage mixing on augmentation chains. Next, we use a second stage mixing sampled from a Beta distribution to mix the clean and the augmented images. In this way, we can obtain final images generated by an arbitrary combination of data augmentation operations with
random mixing weights. We use such images in the training phase of IRUDepth.

\noindent\textbf{Loss Function}.
Following MonoDepth2 \cite{godard2019monodepth2}, we minimize the photometric reprojection error $L_p$. This loss can be calculated as follows:
\begin{equation}
\label{eq:track1_1st_2}
L_p=\min_{t^\prime}\texttt{pe}(I_t, I_{t^\prime\rightarrow t})~,
\end{equation}
\begin{equation}
\label{eq:track1_1st_3}
\texttt{pe}(I_a, I_b)=\frac{\alpha}{2}(1 - \texttt{SSIM}(I_a, I_b))+(1-\alpha)||I_a, I_b||_1~.
\end{equation}
Here we set $\alpha = 0.85$. Additionally, as in \cite{godard2017unsupervised}, we apply the following smoothness loss:
\begin{equation}
\label{eq:track1_1st_4}
L_s=|\partial_x d^*_t|e^{-|\partial_x I_t|} + |\partial_y d^*_t|e^{-|\partial_y I_t|}~,
\end{equation}
where $d^*_t = dt/\bar{d}t$ is the normalized inverse depth as proposed in \cite{wang2018learning}.

To constrain the consistency between the predicted depth maps of the clean and augmented images, we apply the Jensen-Shannon divergence consistency loss used in \cite{hendrycks2020augmix}. This loss aims to enforce smoother neural network responses. Firstly, we mix the depth result as the mixed depth center:
\begin{equation}
\label{eq:track1_1st_5}
D_{mix} = \frac{1}{3}(D_t + D_t^{aug1} + D_t^{aug2})~,
\end{equation}
where $D_t$, $D_t^{aug1}$, and $D_t^{aug2}$ are the depth maps of the clean and the two augmented images, respectively. Next, we compute the triplet loss listed as follows:
\begin{equation}
\label{eq:track1_1st_6}
L_{mix} = \frac{1}{3}\big(\texttt{KL}(D_t || D_{mix}) + \texttt{KL}(D_t^{aug1} || D_{mix}) + \texttt{KL}(D_t^{aug2} || D_{mix})\big)~,
\end{equation}
where the KL divergence (\texttt{KL}) is used to measure the degree of difference between two depth distributions. Note that we use the mixed depth instead of the depth map of clean images in \texttt{KL}, which is proven to perform better in experiments. As in \cite{kannan2018pairing,hendrycks2020augmix}, the triplet loss function in the form of the Jensen-Shannon divergence impels models to be stable, consistent, and insensitive across the input images from diverse scenarios.

Finally, during training, the total loss sums up the above three losses computed from outputs at the
scale $s\in\{1, \frac{1}{2}, \frac{1}{4}, \frac{1}{8}\}$, as computed in the following form:
\begin{equation}
\label{eq:track1_1st_7}
L_{total} = \frac{1}{N}\sum^N_{s=i}(\alpha L_p + \beta L_s + \gamma L_{mix})~,
\end{equation}
where $\alpha$, $\beta$, and $\gamma$ are loss coefficients sampled from the scale set $s$.

\subsubsection{Experimental Analysis}

\noindent\textbf{Implementation Details}.
The IRUDepth framework is implemented using Pytorch. Four NVIDIA GTX 3090 GPUs are used for model training, each with a batch size of $6$. We take MPViT \cite{lee2022mpvit} as the backbone, which is pre-trained on ImageNet-1K, and further fine-tuned on low-resolution ($640\times192$) images from the KITTI dataset \cite{geiger2012kitti} following splits and
data processing in \cite{eigen2015predicting,zhou2017sfm}. The overall framework is optimized end-to-end with the AdamW optimizer \cite{loshchilov2018adamw} for $30$ epochs. The learning rate of the pose network and depth decoder is initially set as $1$e-$4$, while the initial learning rate of the MPViT module is set as $1$e-$5$. Both the learning rates decay by a factor of $10$ for the final $5$ epochs.

\begin{table*}[t]
\caption{Quantitative results on the Robodepth competition leaderboard (Track \# 1). The \textbf{best} and \underline{second best} scores of each metric are highlighted in \textbf{bold} and \underline{underline}, respectively.}
\centering\scalebox{0.78}{
\begin{tabular}{l|cccc|cccc}
    \toprule
    \textbf{Method} & \cellcolor{blue!10}\textbf{Abs Rel~$\downarrow$} & \cellcolor{blue!10}\textbf{Sq Rel~$\downarrow$} & \cellcolor{blue!10}\textbf{RMSE~$\downarrow$} & \cellcolor{blue!10}\textbf{log RMSE~$\downarrow$} & \cellcolor{red!10}$\delta<1.25$~$\uparrow$ & \cellcolor{red!10}$\delta<1.25^2$~$\uparrow$ & \cellcolor{red!10}$\delta<1.25^3$~$\uparrow$
    \\\midrule\midrule
    Ensemble & $0.124$ & $0.899$ & $4.938$ & $0.203$ & \underline{$0.852$} & $0.950$ & \underline{$0.979$}
    \\
    UMCV & $0.124$ & $\mathbf{0.845}$ & \underline{$4.883$} & $0.202$ & $0.847$ & $0.950$ & $\mathbf{0.980}$
    \\
    YYQ & \underline{$0.123$} & \underline{$0.885$} & $4.983$ & \underline{$0.201$} & $0.848$ & $0.950$ & \underline{$0.979$}
    \\
    USTC-IAT-United & \underline{$0.123$} & $0.932$ & $\mathbf{4.873}$ & $0.202$ & $\mathbf{0.861}$ & $\mathbf{0.954}$ & \underline{$0.979$}
    \\\midrule
    \textbf{IRUDepth~(Ours)} & $\mathbf{0.121}$ & $0.919$ & $4.981$ & $\mathbf{0.200}$ & $\mathbf{0.861}$ & \underline{$0.953$} & $\mathbf{0.980}$
    \\\bottomrule
    \end{tabular}
}
\vspace{0.1cm}
\label{tab:track1_1st_comparative}
\end{table*}

\begin{table*}[t]
\caption{Ablation results of IRUDepth on the RoboDepth competition leaderboard. Notations: \texttt{Aug} denotes the proposed image augmentations; $L_{mix}$ denotes the proposed triplet loss. For methods only with \texttt{Aug}, we use augmented images instead of clean images as the input. The \textbf{best} and \underline{second best} scores of each metric are highlighted in \textbf{bold} and \underline{underline}, respectively.}
\centering\scalebox{0.78}{
\begin{tabular}{l|cccc|cccc}
    \toprule
    \textbf{Method} & \cellcolor{blue!10}\textbf{Abs Rel~$\downarrow$} & \cellcolor{blue!10}\textbf{Sq Rel~$\downarrow$} & \cellcolor{blue!10}\textbf{RMSE~$\downarrow$} & \cellcolor{blue!10}\textbf{log RMSE~$\downarrow$} & \cellcolor{red!10}$\delta<1.25$~$\uparrow$ & \cellcolor{red!10}$\delta<1.25^2$~$\uparrow$ & \cellcolor{red!10}$\delta<1.25^3$~$\uparrow$
    \\\midrule\midrule
    MPViT-S & $0.172$ & $1.340$ & $6.177$ & $0.258$ & $0.743$ & $0.910$ & $0.963$
    \\
    MPViT-S + \texttt{Aug} + $L_{mix}$ & \underline{$0.123$} & \underline{$0.946$} & \underline{$5.011$} & \underline{$0.203$} & \underline{$0.855$} & \underline{$0.950$} & \underline{$0.979$}
    \\
    MPViT-B & $0.170$ & $1.212$ & $5.816$ & $0.319$ & $0.753$ & $0.912$ & $0.961$
    \\
    MPViT-B + \texttt{Aug} & $0.146$ & $1.166$ & $5.549$ & $0.226$ & $0.806$ & $0.936$ & $0.974$
    \\\midrule
    MPViT-B + \texttt{Aug} + $L_{mix}$ & $\mathbf{0.121}$ & $\mathbf{0.919}$ & $\mathbf{4.981}$ & $\mathbf{0.200}$ & $\mathbf{0.861}$ & $\mathbf{0.953}$ & $\mathbf{0.980}$
    \\\bottomrule
    \end{tabular}
}
\label{tab:track1_1st_ablation}
\end{table*}

\begin{figure}
    \centering
    \includegraphics[width=1.0\linewidth]{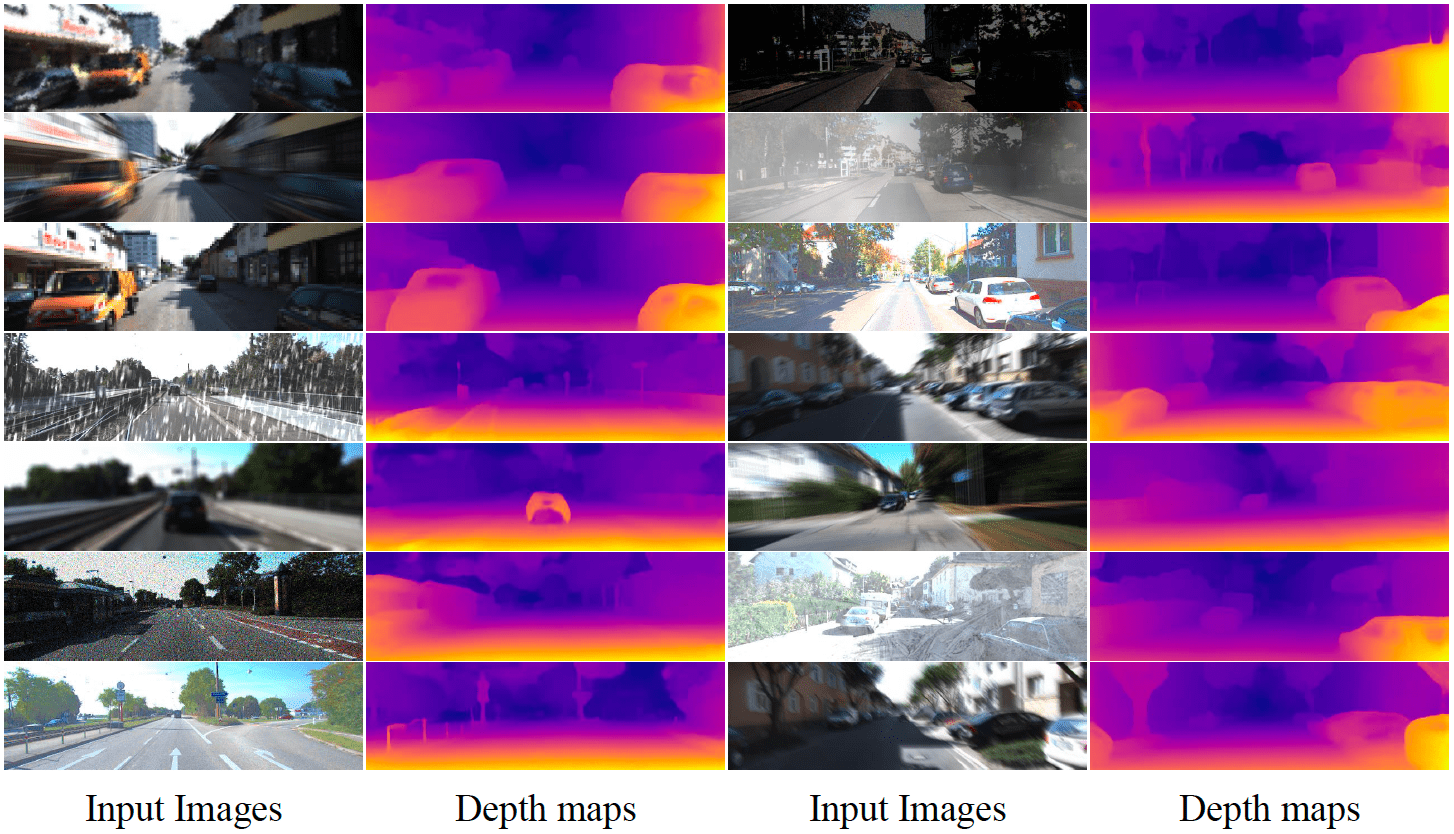}
    \caption{Qualitative results of IRUDepth in the RoboDepth benchmark under different corruptions.}
\label{fig:track1_1st_qualitative}
\end{figure}

\noindent\textbf{Comparative Study}.
The benchmark takes the depth maps of clean images as the ground truth for evaluation. Table~\ref{tab:track1_1st_comparative} compares the model performance with other methods on the RoboDepth competition leaderboard. Our method ranks first on the leaderboard and outperforms other methods across four depth evaluation metrics. Figure~\ref{fig:track1_1st_qualitative} reports some challenging examples from the RoboDepth benchmark. Even for severely corrupted images, our approach could predict accurate and consistent depth maps, which demonstrates the strong robustness of the proposed IRUDepth.

\noindent\textbf{Ablation Study}.
To further validate the effectiveness of the proposed data augmentation module and the triplet loss, we report some ablation results in Table~\ref{tab:track1_1st_ablation}. We assess the impact of different model backbones, image augmentation techniques, and loss functions. As shown in this table, the proposed data augmentation and triplet loss function play an important role in improving the model's performance under OoD corruptions.

\subsubsection{Solution Summary}
In this work, we proposed IRUDepth, a method that aims to improve the robustness and uncertainty estimation of self-supervised monocular depth estimation. With the novel image augmentation and proposed triplet loss function, IRUDepth achieved better generalization performance than state-of-the-art methods for self-supervised depth estimation on the KITTI-C dataset. Moreover, our IRUDepth ranked first in the first track of the RoboDepth Challenge, which demonstrates its superior robustness under different kinds of OoD situations.

\subsection{The \textcolor{robo_red}{2nd} Place Solution: \textcolor{robo_red}{USTC-IAT-United}}
\noindent\textbf{Authors:} \textcolor{gray}{Jun Yu, Xiaohua Qi, Jie Zhang, Mohan Jing, Pengwei Li, Zhen Kan, Qiang Ling, Liang Peng, Minglei Li, Di Xu, and Changpeng Yang.}

\begin{framed}
    \textbf{Summary} - Although current self-supervised depth estimation methods have achieved satisfactory results on ``clean'' data, their performance often disappoints when encountering damaged or unseen data, which are cases that frequently occur in the real world. To address these limitations, the \texttt{USTC-IAT-United} team proposes a solution that includes an MAE mixing augmentation during training and an image restoration module during testing. Both comparative and ablation results verify the effectiveness and superiority of the proposed techniques in handling various types of corruptions that a depth estimation system has to encounter in practice.
\end{framed}

\subsubsection{Overview}
Self-supervised depth estimation aims to estimate the depth map of a given image without the need for explicit supervision. This task is of great importance in computer vision and robotics, as it enables machines to perceive the 3D structure of the environment without the need for expensive depth sensors. Various self-supervised learning methods have been proposed to achieve this task, such as monocular, stereo, and multi-view depth estimation. These methods leverage the geometric and photometric constraints among multiple views of the same scene to learn depth representation.

In recent years, deep learning techniques have been widely adopted in self-supervised depth estimation tasks. Garg \textit{et al.} \cite{garg2016unsupervised} reformulated depth estimation into a view synthesis problem and proposed a photometric loss across stereo pairs to enforce view consistency. Godard \textit{et al.} \cite{godard2017unsupervised} proposed to leverage differentiable bilinear interpolation \cite{jaderberg2015spatial}, virtual stereo prediction, and an SSIM $+$ L1 reconstruction loss to better encourage the left-right consistency. Utilizing solely supervision signals from monocular video sequences, SfM-Learner \cite{zhou2017sfm} relaxed the stereo constraint by replacing the known stereo transform with a pose estimation network. These techniques have shown promising results in various visual perception applications, such as autonomous driving \cite{geiger2012kitti}, augmented reality \cite{yucel2021real}, and robotics \cite{ming2021survey}.

Despite the significant role that monocular and stereo depth estimation play in real-world visual perception systems and the remarkable achievements that have been made, current deep learning-based self-supervised monocular depth estimation models are mostly trained and tested on ``clean'' datasets, neglecting OoD scenarios. Common corruptions, however, often occur in practical scenes, which are crucial for the safety of applications such as autonomous driving and robot navigation. In response to this concern, recent research has focused on developing robust self-supervised depth estimation models that can handle OoD scenarios. The challenge of artifacts arising from dynamic objects has been addressed by integrating uncertainty estimation \cite{klodt2018supervising,poggi2020uncertainty,yang2020d3vo}, motion masks \cite{casser2019depth}, optical flow \cite{luo2019every}, or the minimum reconstruction loss. Simultaneously, to enhance robustness against unreliable photometric appearance, strategies such as feature-based reconstructions \cite{spencer2020general} and proxy-depth supervision \cite{klodt2018supervising} have been introduced.

In recent advancements of network architecture design, several techniques have been incorporated, such as 3D packing and unpacking blocks, positional encoding \cite{zhao2021monovit}, sub-pixel convolution for depth super-resolution \cite{pillai2019superdepth}, progressive skip connections, and self-attention decoders \cite{johnston2020self}. Moreover, some researchers have proposed to use synthetic data to augment the training dataset and improve the model’s generalization ability to OoD scenarios. For example, domain randomization techniques are used to generate diverse synthetic data with various levels of perturbations, which can help the model learn to handle different types of corruption.

In this work, to address this challenging task, we propose a solution with novel designs ranging from the following aspects: 1) an augmented training process and 2) a more stable testing pipeline. For the former stage, we resort to masked autoencoders (MAE) \cite{he2022mae} and image mixing techniques to enhance representation learning of self-supervised depth estimation models. For the latter, we explore off-the-shelf image restoration networks for obtaining images with better visual cues at the test time. Through comparative and ablation experiments, we demonstrate and verify the effectiveness and satisfactory performance of the proposed techniques under challenging OoD scenarios.

\begin{figure}
    \centering
    \includegraphics[width=1.0\linewidth]{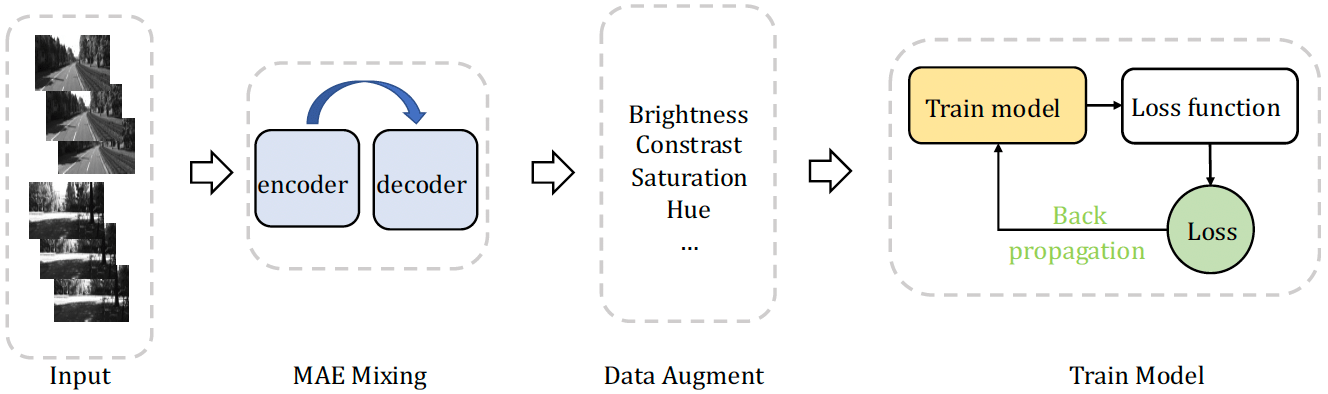}
    \caption{Illustration of the training pipeline in our proposed robust depth estimation solution.}
\label{fig:track1_2nd_training}
\end{figure}

\subsubsection{Technical Approach}
Our proposed solution consists of the following major components: 1) the training pipeline, 2) the test pipeline, 3) an MAE mixing operation, and 4) an image restoration module.

\noindent\textbf{Training Pipeline}. Our model training pipeline, as shown in Figure~\ref{fig:track1_2nd_training}, is composed of four main parts: 1) the input of image sequences or image pairs, 2) an MAE reconstruction with image mixing, 3) a data augmentation module, and 4) the overall model training objectives. The input of our model follows MonoDepth2 \cite{godard2019monodepth2}, where image sequences are used for single-view training and left-right image pairs are used for stereo training. The MAE mixing operation will be elaborated on later. We also adopt the data augmentation settings of MonoDepth2 \cite{godard2019monodepth2} during training. We use the MonoViT \cite{zhao2021monovit} architecture as our backbone model, which consists of MPViT \cite{lee2022mpvit} encoder blocks and a self-attention decoder. The training is conducted in a self-supervised manner on the KITTI depth estimation dataset \cite{geiger2012kitti}, using the photometric loss \cite{godard2019monodepth2} for both monocular frames and stereo pairs, as well as a proxy depth regression objective. The regularization is achieved by incorporating edge-aware disparity smoothness \cite{godard2017unsupervised} and depth gradient consistency with respect to the proxy labels.

\begin{figure}
    \centering
    \includegraphics[width=1.0\linewidth]{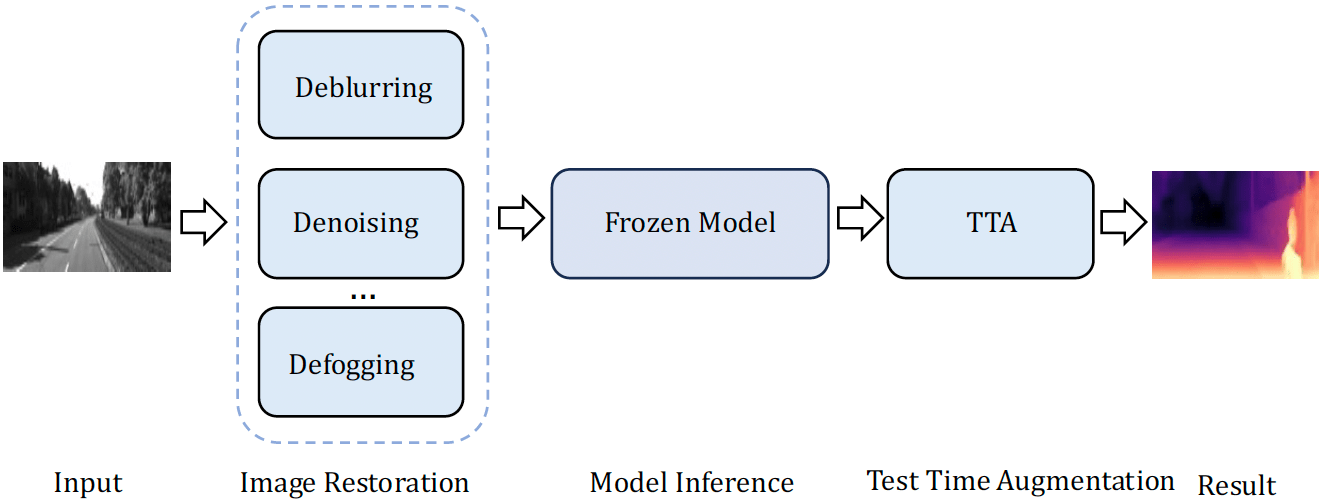}
    \caption{Illustration of the testing pipeline in our proposed robust depth estimation solution.}
\label{fig:track1_2nd_testing}
\end{figure}

\noindent\textbf{Testing Pipeline}.
The proposed solution consists of five components in the testing phase, as shown in Figure~\ref{fig:track1_2nd_testing}: 1) the input image for inference, 2) an image restoration module, 3) model inference, 4) a test-time augmentation (TTA) technique, and 5) the final depth prediction result. The specific process is as follows: we use a single image as the input for inference, which is then enhanced with the image restoration module to be described later. The restored image is then fed into the depth estimation model for feature extraction and prediction. Finally, a TTA approach based on MonoDepth2 \cite{godard2019monodepth2} is applied as a post-processing technique to produce the final result. The entire process is mathematically and academically rigorous.

\noindent\textbf{MAE Reconstruction}.
The masking-based image reconstruction method aims for reconstructing masked regions in an image by minimizing the mean absolute error between the original input and its reconstruction. Mathematically, given an image $x$ and its reconstruction $\hat{x}$, the MAE reconstruction process can be formulated as follows:
\begin{equation}
\label{eq:track1_2nd_1}
\hat{x} = \arg\min_{\hat{x}} \frac{1}{n}\sum^{n}_{i=1}|x_i - \Tilde{x}_i|~,
\end{equation}
where $n$ is the number of pixels in the image, and $x_i$ and $\Tilde{x}_i$ represent the $i$-th pixel of the original image and its reconstruction, respectively.

MAE is a type of network that can be used for unsupervised learning of visual features, which is particularly well-suited for learning from large-scale datasets as they can be trained efficiently on distributed computing systems. The basic idea of MAE is to learn a compressed representation of an image by encoding it into a lower-dimensional space and then decoding it back to its original size. Unlike traditional autoencoders which use fully connected layers for both the encoder and decoder, MAE uses convolutional layers to capture spatial information and reduce the number of parameters.

The MAE reconstruction process not only preserves semantic information similar to the original image but also introduces blurriness and distortion, making it a suitable method for enhancing robustness under various OoD corruptions. In this challenge, we directly load a pre-trained MAE model \cite{he2022mae} for image reconstruction of the input image $x$. Specifically, the pre-trained model $f$ can be represented as a function that maps the input image $x$ to its reconstructed image $\hat{x}$, \textit{i.e.}, $\hat{x}=f(x)$.

\noindent\textbf{Image Mixing}.
Blending different images is a commonly-used data augmentation technique. It can be used to generate new training samples by mixing two or more images together. The basic idea is to combine the content of two or more images in a way that preserves the semantic information while introducing some degree of variability. This can help the model learn to be more robust to changes in the input data and improve its generalization performance.

One common approach for image mixing is to conduct a weighted sum of the pixel values from different input images. Given two images $I_A$ and $I_B$, we can generate a mixed image $I_C$ as follows:
\begin{equation}
\label{eq:track1_2nd_2}
I_C = (1-\alpha)I_A + \alpha I_B~,
\end{equation}
where $\alpha$ is a mixing coefficient that controls the degree of influence of each of the two images. For example, when $\alpha = 0.5$, the resulting image is an equal blend of the two inputs. When $\alpha$ is closer to $0$ or $1$, the resulting image is more similar to one of these two candidate input images.

To introduce a certain degree of randomness and diversity into the mixing process, we can use different values of $\alpha$ for each pair of images. This can further increase the variability of the generated samples and improve the model’s ability to handle different types of input data. Image mixing has been shown to be an effective data augmentation technique for various computer vision tasks, including image classification, object detection, and semantic segmentation. It can help the model learn to be more robust to changes in the input data and improve its generalization performance.

\noindent\textbf{MAE Mixing}.
Different from the aforementioned image mixing, our MAE mixing operation refers to the mixing of the MAE-reconstructed image and the original image. This mixing process can be mathematically described as follows:
\begin{equation}
\label{eq:track1_2nd_3}
x_{mix} = (1-\alpha)x + \alpha \hat{x}~,
\end{equation}
where $x$ and $\hat{x}$ represent the original image and the MAE-reconstructed image, respectively, and $\alpha$ is a hyperparameter representing the mixing ratio.

\begin{figure}
    \centering
    \includegraphics[width=0.9\linewidth]{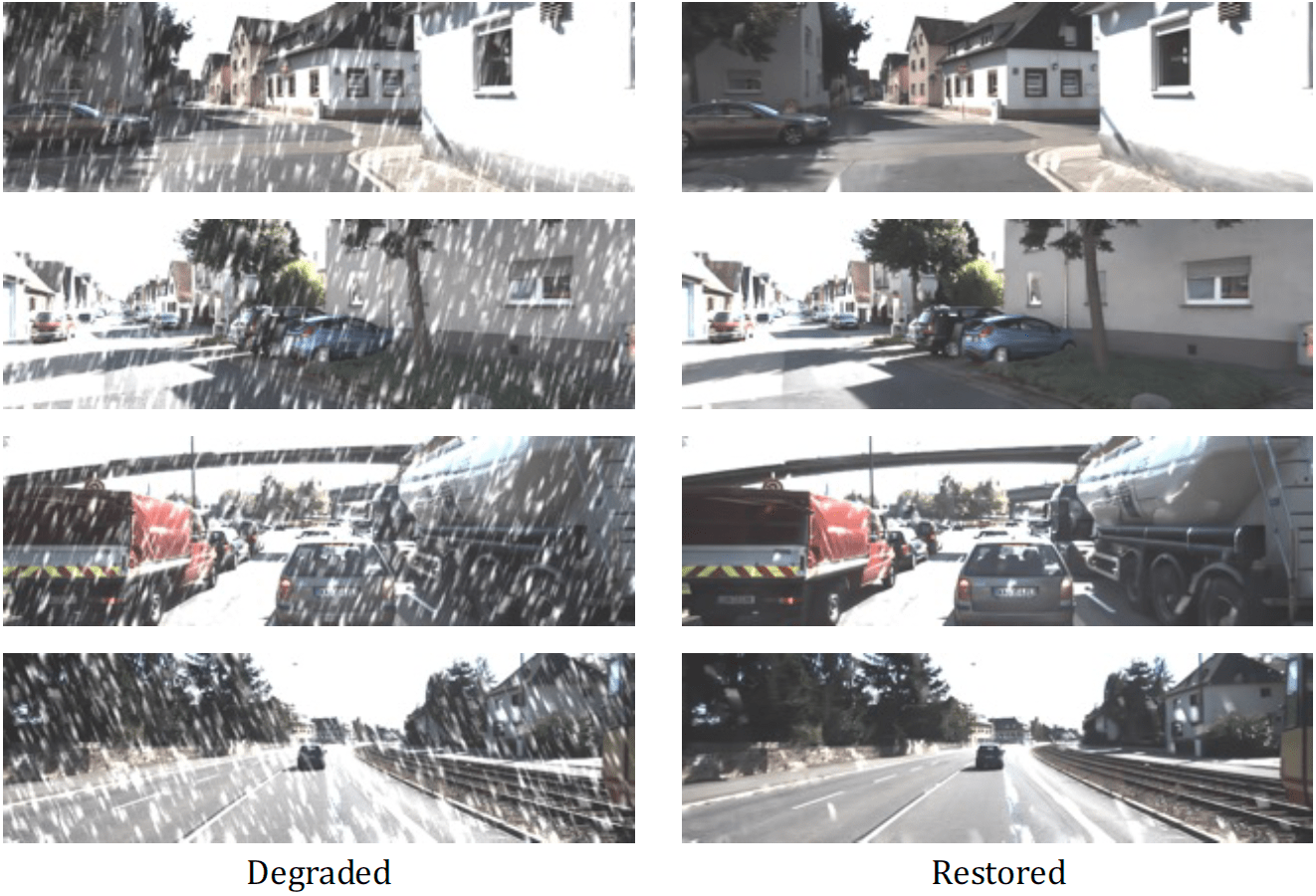}
    \caption{Visualizing the effectiveness of using the Restormer network for snow removal.}
\label{eq:track1_2nd_restore_snow}
\end{figure}

\begin{figure}
    \centering
    \includegraphics[width=0.9\linewidth]{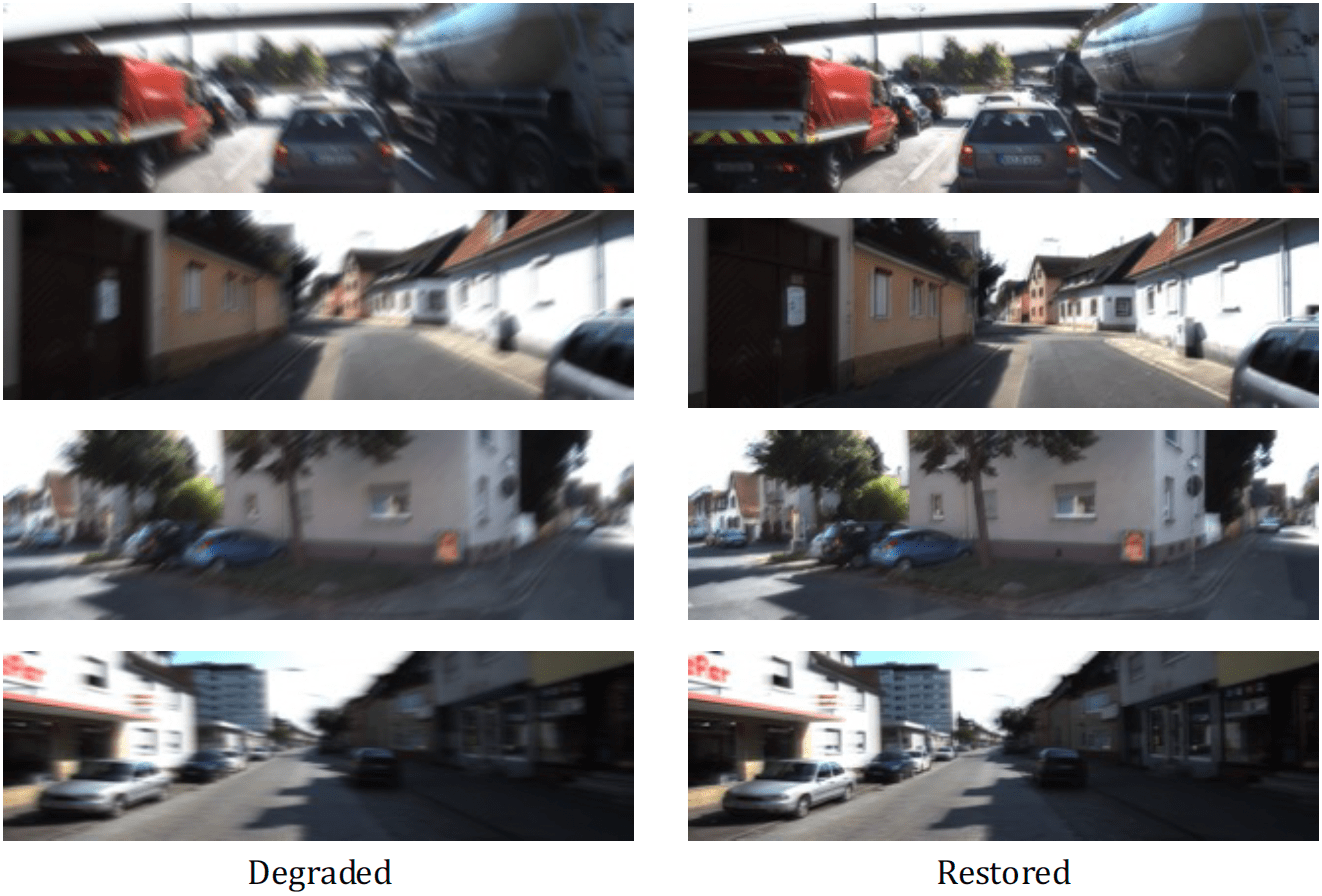}
    \caption{Visualizing the effectiveness of using the Restormer network for motion deblurring.}
\label{eq:track1_2nd_restore_motion}
\end{figure}

By combining the reconstructed images with the original ones, the diversity of the training data can be greatly enriched, thereby enhancing the robustness of the depth estimation model. Without the need for altering the supervision signal, we achieve such mixing and control its degree using weighted image interpolation, as described earlier. The resulting mixed image $x_{mix}$ can be used as the input to the depth estimation model, thereby increasing the diversity of the training data and improving the model’s ability to generalize to unseen data.

\noindent\textbf{Image Restoration}.
The goal of image restoration is to recover a blurred or noisy image without changing its size and content. To perform such a restoration, we use an efficient image restoration network called Restormer \cite{zamir2022restormer}. This model is based on the Transformer backbone to restore damaged images. In this challenge, we did not further fine-tune the network but directly loaded the pre-trained Restormer \cite{zamir2022restormer} checkpoint to restore the corrupted images.

\begin{figure}
    \centering
    \includegraphics[width=0.9\linewidth]{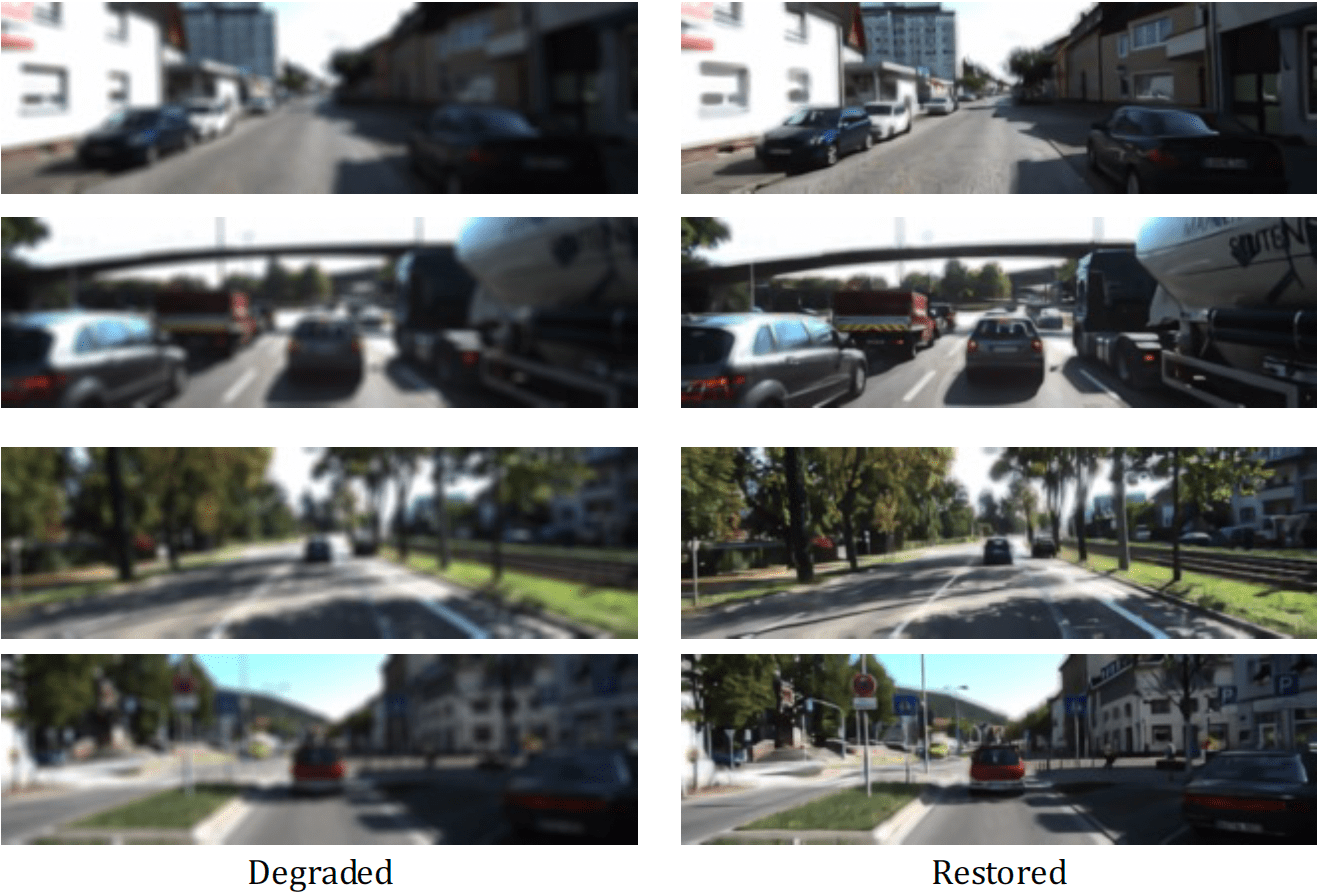}
    \caption{Visualizing the effectiveness of using the Restormer network for defocus deblurring.}
\label{eq:track1_2nd_restore_defocus}
\end{figure}

\begin{figure}
    \centering
    \includegraphics[width=0.9\linewidth]{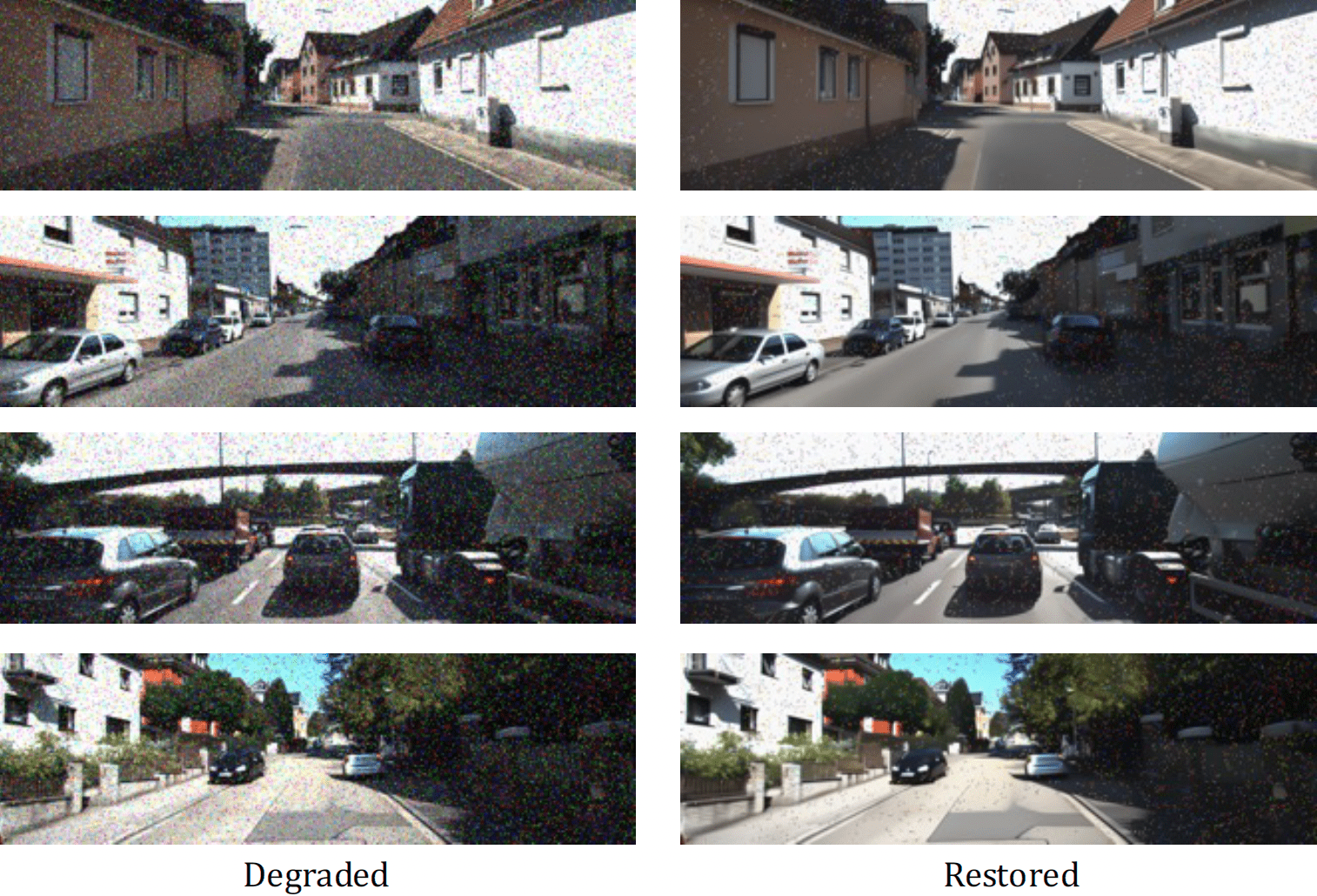}
    \caption{Visualizing the effectiveness of using the Restormer network for de-noising.}
\label{eq:track1_2nd_restore_denoise}
\end{figure}

As shown in Figure~\ref{fig:track1_2nd_testing}, before feeding the test images into the depth estimation model, we perform image restoration to enhance the image quality. Specifically, we first restore the damaged images using the Restormer network, which is pre-trained on various restoration tasks including `image de-raining', `single-image motion de-blurring', `defocus de-blurring', and `image de-noising'. After the restoration process, we use the restored images as the input of our depth estimation model for further processing. Mathematically, the restoration process can be formulated as follows:
\begin{equation}
\label{eq:track1_2nd_4}
\hat{I} = \texttt{Restormer}(I)~,
\end{equation}
where $I$ denotes the input image of the image restoration network and $\hat{I}$ denotes the restored image. Subsequently, the depth estimation process can be formulated as follows:
\begin{equation}
\label{eq:track1_2nd_5}
D = \texttt{DepthEstimate}(\hat{I})~,
\end{equation}
where $D$ denotes the estimated depth map. Figure~\ref{eq:track1_2nd_restore_snow} to Figure~\ref{eq:track1_2nd_restore_denoise} provide representative results of various types of corrupted images and their restored versions from Restormer \cite{zamir2022restormer}.

Specifically, Figure~\ref{eq:track1_2nd_restore_snow} displays the restoration results of images degraded by \textit{`snow'}; while Figure~\ref{eq:track1_2nd_restore_motion} shows the restoration results of images degraded by \textit{`motion blur'}. Figure~\ref{eq:track1_2nd_restore_defocus} and Figure~\ref{eq:track1_2nd_restore_denoise} present the restoration results of images degraded by \textit{`defocus blur'} and by \textit{`noises'}, respectively. In each figure, the left-hand-side images represent the inputs that are degraded by different kinds of real-world corruptions, while the right-hand-side images are the restored outputs. The results demonstrate the effectiveness of the Restormer network in restoring images degraded by various types of distortions.

\subsubsection{Experimental Analysis}

\noindent\textbf{Implementation Details}.
We use the standard Eigen split \cite{eigen2015predicting} of the KITTI depth estimation dataset \cite{geiger2012kitti} as our training dataset and trained our models with the corresponding hyperparameters specified in the MonoDepth2 paper \cite{godard2019monodepth2}. We then fine-tuned the pre-trained models using our MAE mixing data augmentation, with the starting learning rate being one-fifth of the original learning rate.

\noindent\textbf{Baselines}.
We evaluated the robustness of multiple self-supervised depth estimation models on the corrupted dataset and identified four models with superior depth performance under OoD scenarios: CADepth \cite{yan2021cadepth}, MonoDepth2 \cite{godard2019monodepth2}, Lite-Mono \cite{zhang2023litemono}, and MonoViT \cite{zhao2021monovit}. Their depth estimation results are shown in Table~\ref{tab:track1_2nd_baselines}. We can observe from this table that MonoViT \cite{zhao2021monovit} achieves the best OoD depth estimation performance when trained with the Mono+Stereo modality and with an input resolution of $640\times192$. Therefore, all subsequent experiments are conducted with this configuration.

\noindent\textbf{MAE Mixing}.
We conducted experiments to investigate the impact of the mixing ratio hyperparameter $\alpha$ in our MAE mixing data augmentation. The ablation results are shown in Table~\ref{tab:track1_2nd_mixing_ratio}. Specifically, we varied the mixing ratio $\alpha$ between the original image $x$ and the MAE-reconstructed image $\hat{x}$, where the mixed image is given by Eq.~\ref{eq:track1_2nd_3}. As can be seen from the results, the mixing ratio $\alpha$ should be carefully selected for our MAE mixing data augmentation, as excessively high or low values of $\alpha$ can negatively impact the model’s performance.

\begin{table*}[t]
\caption{The performance of multiple models trained on the standard Eigen split of the KITTI dataset.}
\centering\scalebox{0.78}{
\begin{tabular}{c|c|c|c|c|c}
    \toprule
    \textbf{Method} & \textbf{Ref} & \textbf{Input Modality} & \textbf{Input Resolution} & \cellcolor{blue!10}\textbf{Abs Rel~$\downarrow$} & $\Delta$
    \\\midrule\midrule
    \multirow{6}{*}{MonoDepth2} & \multirow{6}{*}{\cite{godard2019monodepth2}} & Mono & $640\times192$ & $0.149$ & \textcolor{gray}{$+0.000$}
    \\
    & & Stereo & $640\times192$ & $0.153$ & \textcolor{robo_red}{$+0.004$}
    \\
    & & Mono+Stereo & $640\times192$ & $0.146$ & \textcolor{robo_blue}{$-0.003$}
    \\
    & & Mono & $1024\times320$ & $0.153$ & \textcolor{robo_red}{$+0.004$}
    \\
    & & Stereo & $1024\times320$ & $0.154$ & \textcolor{robo_red}{$+0.005$}
    \\
    & & Mono+Stereo & $1024\times320$ & $0.240$ & \textcolor{robo_red}{$+0.091$}
    \\\midrule
    \multirow{5}{*}{CADepth} & \multirow{5}{*}{\cite{yan2021cadepth}} & Mono & $640\times192$ & $0.149$ & \textcolor{gray}{$+0.000$}
    \\
    & & Mono & $1024\times320$ & $0.151$ & \textcolor{robo_red}{$+0.002$}
    \\
    & & Mono & $1280\times384$ & $0.157$ & \textcolor{robo_red}{$+0.008$}
    \\
    & & Mono+Stereo & $640\times192$ & $0.147$ & \textcolor{robo_blue}{$-0.002$}
    \\
    & & Mono+Stereo & $1024\times320$ & $0.143$ & \textcolor{robo_blue}{$-0.006$}
    \\\midrule
    Lite-Mono-L & \cite{zhang2023litemono} & Mono & $1024\times320$ & $0.148$ & \textcolor{robo_blue}{$-0.001$}
    \\\midrule
    \multirow{5}{*}{MonoViT} & \multirow{5}{*}{\cite{zhao2021monovit}} & Mono & $640\times192$ & $0.143$ & \textcolor{robo_blue}{$-0.006$}
    \\
    & & Mono+Stereo & $640\times192$ & $0.134$ & \textcolor{robo_blue}{$-0.015$}
    \\
    & & Mono & $1024\times320$ & $0.149$ & \textcolor{gray}{$+0.000$}
    \\
    & & Mono+Stereo & $1024\times320$ & $0.138$ & \textcolor{robo_blue}{$-0.011$}
    \\
    & & Mono & $1280\times384$ & $0.147$ & \textcolor{robo_blue}{$-0.002$}
    \\\bottomrule
    \end{tabular}
}
\label{tab:track1_2nd_baselines}
\end{table*}

\begin{table*}[t]
\caption{Ablation results of MAE mixing ratio $\alpha$ on the Robodepth competition leaderboard (Track \# 1). The \textbf{best} and \underline{second best} scores of each metric are highlighted in \textbf{bold} and \underline{underline}, respectively.}
\centering\scalebox{0.78}{
\begin{tabular}{c|c|c|c|c|c}
    \toprule
    Mixing Ratio & $\alpha=0.1$ & $\alpha=0.3$ & $\alpha=0.5$ & $\alpha=0.7$ & $\alpha=0.9$
    \\\midrule
    \cellcolor{blue!10}\textbf{Abs Rel~$\downarrow$} & \underline{$0.125$} & $\mathbf{0.123}$ & $0.128$ & $0.132$ & $0.137$
    \\\bottomrule
    \end{tabular}
}
\vspace{0.1cm}
\label{tab:track1_2nd_mixing_ratio}
\end{table*}

Without loss of generalizability, our experimental results indicate that a mixing ratio of $\alpha = 0.3$ achieves the optimal performance on the OoD testing set and better enhances the model’s generalization ability. This suggests that a balanced mixture of the original image and the MAE-reconstructed image is beneficial for the model’s representation learning. On the other hand, excessively high values of $\alpha$ can lead to overfitting issues, where the model becomes too specialized to the training data and performs poorly on new data. Conversely, excessively low values of $\alpha$ may not provide enough variation in the augmented data, leading to underfitting and poor performance. One possible reason for the sensitivity of the MAE mixing method to the mixing ratio hyperparameter $\alpha$ is the use of image restoration techniques. The restoration algorithm may introduce artifacts or distortions in the reconstructed image, which can affect the performance of the MAE mixing method. Furthermore, the restoration algorithm only operates on the testing set, while the MAE mixing data augmentation is used during training, making it necessary to carefully tune the hyperparameter $\alpha$.

To address this issue, an end-to-end training approach can be explored in future work. This would involve jointly training the restoration algorithm and the downstream task model, allowing for better integration of the restoration and augmentation processes. By incorporating the restoration algorithm into the training process, the sensitivity of the MAE mixing method to the mixing ratio hyperparameter $\alpha$ can potentially be reduced, leading to improved performance and generalization ability.

\noindent\textbf{Image Restoration}.
In the final stage of our experiments, we applied the image restoration process described in previous sections to the testing images before depth inference. This resulted in an improved absolute relative error (in terms of the \texttt{Abs Rel} score) of $0.123$. The image restoration process helps to reduce the negative impact of artifacts and distortions in corrupted images, leading to more accurate predictions by the depth estimation model. By incorporating this step into the testing pipeline, we are able to achieve better performance over the baselines. Furthermore, the use of image restoration techniques can also improve the generalization ability of the depth estimation model, as it helps to reduce the impact of variations and imperfections across a wide range of test images.

\subsubsection{Solution Summary}
In this work, we have attempted various strategies to address the challenging OoD self-supervised monocular depth estimation. We first demonstrated that the CNN-Transformer hybrid networks exhibit excellent robustness over plain CNN-based ones. We designed and employed an efficient data augmentation method -- MAE mixing -- which can serve as a strong enhancement for depth estimation. Additionally, we have shown that the image restoration network can effectively handle common distortions at test time, such as blur, noise, rain, and snow, and can significantly improve depth prediction scores. Ultimately, our solution achieved an absolute relative error of $0.123$ and ranked second in the first track of the RoboDepth Challenge.

\subsection{The \textcolor{robo_green}{3rd} Place Solution: \textcolor{robo_green}{YYQ}}
\noindent\textbf{Authors:} \textcolor{gray}{Yuanqi Yao, GangWu, Jian Kuai, Xianming Liu, and Junjun Jiang.}

\begin{framed}
    \textbf{Summary} - The \texttt{YYQ} team proposes to enhance the OoD robustness of self-supervised depth estimation models via joint adversarial training. Adversarial samples are introduced during training to reduce the sensitivity of depth prediction models to minimal perturbations in the corrupted input data. This approach also ensures the depth estimation models maintain their performance on the in-distribution scenarios while being more robust to different types of data corruptions. Extensive ablation results showcase the effectiveness of the proposed approach.
\end{framed}

\subsubsection{Overview}

Self-supervised depth estimation has emerged as a crucial technique in visual perception tasks, enabling the inference of depth information from 2D images without the use of expensive 3D sensors. However, like conventional depth estimation algorithms, self-supervised depth estimation models trained on ``clean'' datasets often lack robustness and generalization ability when faced with naturally corrupted data. This issue is particularly relevant in real-world scenarios where it is often difficult to ensure that the input data at test time matches the ideal image distribution of the training dataset. Additionally, adversarial attacks can also lead to incorrect depth estimation results, posing safety hazards in applications such as autonomous driving.

To address the above challenges, we propose a method for enhancing the robustness of existing self-supervised depth estimation models via adversarial training. Specifically, adversarial samples are introduced during training to force the depth estimation model to process modified inputs that aim to deceive the discriminator model. By doing so, we can reduce the sensitivity of the self-supervised depth estimation model to minimal perturbations in the input data, ensuring that the model can be trained on a ``clean'' dataset while maintaining a certain degree of robustness to common types of corruptions in the real world.

We believe that our approach will play a significant role in future vision perception applications, providing more reliable depth estimation algorithms for various fields, including autonomous driving \cite{geiger2012kitti}, augmented reality \cite{yucel2021real}, and robot navigation \cite{RobustNav}.  Furthermore, our approach also provides a new aspect to improve the robustness of learning-based models in other self-supervised learning tasks with no extra cost. Experimental results in the RoboDepth competition leaderboard demonstrate that our proposed method can improve the depth estimation scores over existing models by $23\%$ on average while still maintaining their original performance on the ``clean'' KITTI dataset \cite{geiger2012kitti}. These results verify the effectiveness and practicality of our proposed approach.

\subsubsection{Technical Approach}
\begin{figure}
  \centering
  \includegraphics[width=1.0\textwidth]{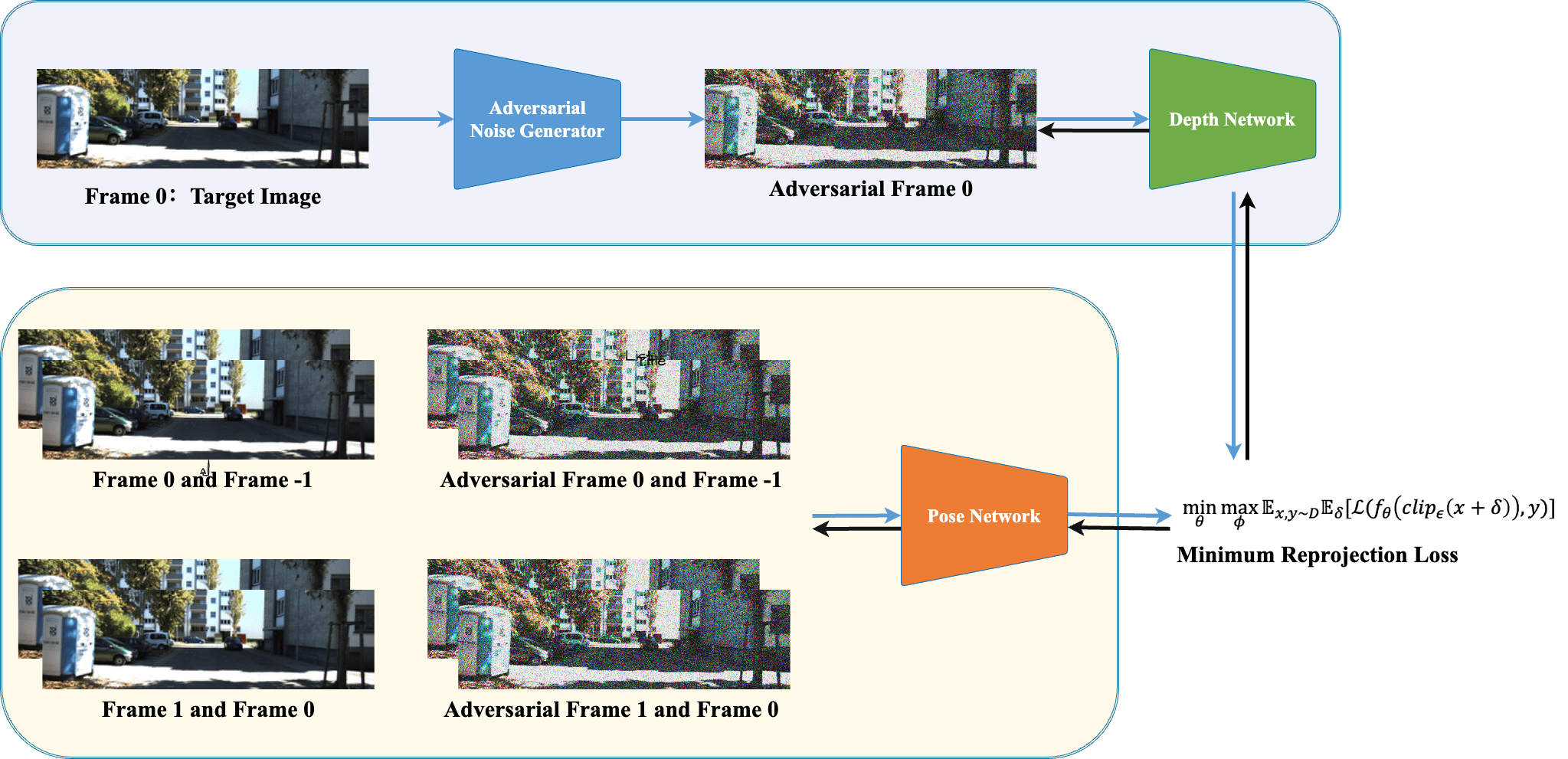}
  \caption{The overall architecture of our framework. We employ a subset of multi-frame images for adversarial training, which incorporates both ``clean'' and adversarial images into the encoder, decoder, and pose network, without altering the model structure. The image reprojection loss serves as a constraint for the corresponding adversarial noise generator, providing a simple yet effective way to enhance the self-supervised depth estimation model's robustness.}
\label{fig:3rd_framework}
\end{figure}

Our approach can be divided into two main parts as shown in Figure~\ref{fig:3rd_framework}. In the first part, we propose a constrained adversarial training method for self-supervised depth estimation, which allows us to jointly train the depth estimation model and the adversarial noise generator. The adversarial noise generator is designed to produce spatially uncorrelated noises with adversarial properties to counter a specified depth estimation network. The second part is a model ensemble, where we improve the robustness of individual models by fusing between different training settings and model sizes.

\noindent\textbf{Joint Adversarial Training}.
In the joint adversarial training stage, we use a simple method to jointly train an adversarial noise generator and the depth estimation model, making it useful to enhance the robustness of any existing self-supervised depth estimation model. 

Specifically, we first initialize an adversarial noise generator for adding adversarial noise to the depth estimation model, and then jointly train the depth estimation model with the adversarial noise generator. This encourages the trained depth estimation model to be robust to adversarial noise perturbations. In actual implementations, we use the common reprojection loss in self-supervised depth estimation as the supervision loss for optimizing the adversarial noise generator.

To facilitate robustness during feature learning, we now train the depth estimation model $f_\theta$ to minimize the risk under adversarial noise distributions jointly with the noise generator as follows:
\begin{equation}
\label{eq:track1_3rd_eq1}
\min _{\theta} \max _{\phi} \mathbb{E}_{x, y \sim D} \mathbb{E}_{\delta}\left[\mathcal{L}\left(f_{\theta}\left(\operatorname{clip}_{\epsilon}(\boldsymbol{x}+\boldsymbol{\delta})\right), y\right)\right]~,
\end{equation}
where $\boldsymbol{x}+\boldsymbol{\delta} \in[0,1]^{N}$ and $\|\boldsymbol{\delta}\|_{2}=\epsilon$. Here $\mathcal{L}$ represents the photometric reprojection error $L_p$ in MonoDepth2 \cite{godard2019monodepth2}, which can be formulated as follows:
\begin{equation}
\label{eq:track1_3rd_eq2}
L_{p}=\sum_{t^{\prime}} p e\left(I_{t}, I_{t^{\prime} \rightarrow t}\right)~.
\end{equation}
The noise generator $g_\phi$ consists of four $1\times1$ convolutional layers that use Rectified Linear Unit (ReLU) activations and include a residual connection that connects the input directly to the output. To ensure accurate depth estimation on ``clean'' KITTI images, we adopt a strategy that samples mini-batches comprising $50\%$ ``clean'' data and $50\%$ perturbed data. Out of the perturbed data, we use the current state of the noise generator to perturb $30\%$ of images from this source, while the remaining $20\%$ is augmented with samples from previous distributions selected randomly. To facilitate this process, we save the noise generator's states at regular intervals.

The overall framework of our approach is shown in Figure~\ref{fig:3rd_framework}. The network architecture we adopted remained consistent with MonoViT \cite{zhao2021monovit} except for the adversarial network. Firstly, we use ``clean'' multi-frame images as the input to the adversarial noise generator to obtain adversarial multi-frame images. Next, we feed the adversarial and ``clean'' images with a certain proportion into the encoder, decoder, and pose network without changing the original model architecture. We use the image reprojection loss as a constraint for optimizing the corresponding adversarial noise generator. 

\noindent\textbf{Model Ensemble}.
To further enhance the robustness of individual depth estimation models, we use a model ensemble strategy separately on both the small and base variants of MonoViT \cite{zhao2021monovit}, \textit{i.e.} MonoViT-S and MonoViT-B. Specifically, we verify the performance of MonoViT-S with and without a model ensemble, as well as MonoViT-B, which are improved by $3\%$ and $6\%$, respectively. Finally, considering that different model sizes could also affect the model's representation learning by focusing on different features, we ensemble the MonoViT-S and MonoViT-B models to achieve the best possible performance in our final submission. 

\subsubsection{Experimental Analysis}

\noindent\textbf{Implementation Details}.
We implement our proposed approach using PyTorch. The MonoViT-S model is trained on a single NVIDIA GeForce RTX 3090 GPU, while the MonoViT-B model is trained on a single NVIDIA A100 GPU. During training, only images from the training split of the KITTI depth estimation dataset \cite{geiger2012kitti} are used.

\noindent\textbf{Comparative Study}.
As shown in Table~\ref{tab:track1_3rd_at}, Table~\ref{tab:track1_3rd_at2}, and Table~\ref{tab:track1_3rd_ensemble}, our proposed approach improves the performance of existing self-supervised depth estimation models by $23\%$ on average under corrupted scenarios, while still maintaining good performance on the ``clean'' testing dataset.

\noindent\textbf{Joint Adversarial Training}.
Table~\ref{tab:track1_3rd_at} shows the evaluation results of the proposed joint adversarial training. It can be seen that such a training enhancement approach significantly improves the robustness of existing depth estimation models under OoD corruptions. The results from Table~\ref{tab:track1_3rd_at2} further validate that our method not only brings a positive impact on OoD settings but also maintains excellent performance on the ``clean'' testing set. We believe this advantage ensures the accurate estimation of depth information for images in any scenario.

\begin{table*}
  \caption{Quantitative results of the baseline and our proposed joint adversarial training approach on the RoboDepth competition leaderboard (Track \# 1). The \textbf{best} and \underline{second best} scores of each metric are highlighted in \textbf{bold} and \underline{underline}, respectively.}
  \centering\scalebox{0.78}{
  \begin{tabular}{l|cccc|ccc}
    \toprule
    \textbf{Method} & \cellcolor{blue!10}\textbf{Abs Rel~$\downarrow$} & \cellcolor{blue!10}\textbf{Sq Rel~$\downarrow$} & \cellcolor{blue!10}\textbf{RMSE~$\downarrow$} & \cellcolor{blue!10}\textbf{log RMSE~$\downarrow$} & \cellcolor{red!10}$\delta<1.25$~$\uparrow$ & \cellcolor{red!10}$\delta<1.25^2$~$\uparrow$ & \cellcolor{red!10}$\delta<1.25^3$~$\uparrow$
    \\\midrule\midrule
    MonoViT-S & \multirow{2}{*}{$0.160$} & \multirow{2}{*}{$1.238$} & \multirow{2}{*}{$5.935$} & \multirow{2}{*}{$0.245$} & \multirow{2}{*}{$0.768$} & \multirow{2}{*}{$0.920$} & \multirow{2}{*}{$0.967$}
    \\
    (Baseline) & & & & & & 
    \\\midrule
    MonoViT-S & \multirow{2}{*}{\underline{$0.135$}} & \multirow{2}{*}{\underline{$1.066$}} & \multirow{2}{*}{$\mathbf{5.258}$} & \multirow{2}{*}{$0.215$} & \multirow{2}{*}{$0.829$} & \multirow{2}{*}{\underline{$0.942$}} & \multirow{2}{*}{$\mathbf{0.976}$}
    \\
    + Adversarial Training & & & & & & & 
    \\\midrule
    MonoViT-B & \multirow{2}{*}{$\mathbf{0.130}$} & \multirow{2}{*}{$\mathbf{1.027}$} & \multirow{2}{*}{\underline{$5.281$}} & \multirow{2}{*}{$\mathbf{0.213}$} & \multirow{2}{*}{$\mathbf{0.839}$} & \multirow{2}{*}{$\mathbf{0.945}$} & \multirow{2}{*}{\underline{$0.975$}}
    \\
    + Adversarial Training & & & & & & & 
    \\\bottomrule
  \end{tabular}
}
\label{tab:track1_3rd_at}
\end{table*}

\begin{table*}
  \caption{Quantitative results f the baseline and our proposed joint adversarial training approach on the testing set of the KITTI dataset \cite{geiger2012kitti}. The \textbf{best} and \underline{second best} scores of each metric are highlighted in \textbf{bold} and \underline{underline}, respectively.}
  \centering\scalebox{0.78}{
  \begin{tabular}{l|cccc|ccc}
    \toprule
    \textbf{Method} & \cellcolor{blue!10}\textbf{Abs Rel~$\downarrow$} & \cellcolor{blue!10}\textbf{Sq Rel~$\downarrow$} & \cellcolor{blue!10}\textbf{RMSE~$\downarrow$} & \cellcolor{blue!10}\textbf{log RMSE~$\downarrow$} & \cellcolor{red!10}$\delta<1.25$~$\uparrow$ & \cellcolor{red!10}$\delta<1.25^2$~$\uparrow$ & \cellcolor{red!10}$\delta<1.25^3$~$\uparrow$
    \\\midrule\midrule
    MonoViT-S & \multirow{2}{*}{$0.104$} & \multirow{2}{*}{\underline{$0.747$}} & \multirow{2}{*}{$4.461$} & \multirow{2}{*}{$0.177$} & \multirow{2}{*}{$0.897$} & \multirow{2}{*}{$\mathbf{0.966}$} & \multirow{2}{*}{\underline{$0.983$}}
    \\
    + Adversarial Training & & & & & & & 
    \\\midrule
    MonoViT-B & \multirow{2}{*}{\underline{$0.100$}} & \multirow{2}{*}{\underline{$0.747$}} & \multirow{2}{*}{\underline{$4.427$}} & \multirow{2}{*}{\underline{$0.176$}} & \multirow{2}{*}{\underline{$0.901$}} & \multirow{2}{*}{$\mathbf{0.966}$} & \multirow{2}{*}{$\mathbf{0.984}$}
    \\
    (Baseline) & & & & & & & 
    \\\midrule
    MonoViT-B & \multirow{2}{*}{$\mathbf{0.099}$} & \multirow{2}{*}{$\mathbf{0.725}$} & \multirow{2}{*}{$\mathbf{4.356}$} & \multirow{2}{*}{$\mathbf{0.175}$} & \multirow{2}{*}{$\mathbf{0.902}$} & \multirow{2}{*}{$\mathbf{0.966}$} & \multirow{2}{*}{$\mathbf{0.984}$}
    \\
    + Adversarial Training & & & & & & & 
    \\\bottomrule
  \end{tabular}
}
\label{tab:track1_3rd_at2}
\end{table*}

\begin{table*}[t]
  \caption{Quantitative results of the baseline and the model ensemble strategy on the RoboDepth competition leaderboard (Track \# 1). Here AT denotes models trained with the proposed joint adversarial training approach. The \textbf{best} and \underline{second best} scores of each metric are highlighted in \textbf{bold} and \underline{underline}, respectively.}
  \label{table3}
  \centering\scalebox{0.78}{
  \begin{tabular}{l|cccc|ccc}
    \toprule
    \textbf{Method} & \cellcolor{blue!10}\textbf{Abs Rel~$\downarrow$} & \cellcolor{blue!10}\textbf{Sq Rel~$\downarrow$} & \cellcolor{blue!10}\textbf{RMSE~$\downarrow$} & \cellcolor{blue!10}\textbf{log RMSE~$\downarrow$} & \cellcolor{red!10}$\delta<1.25$~$\uparrow$ & \cellcolor{red!10}$\delta<1.25^2$~$\uparrow$ & \cellcolor{red!10}$\delta<1.25^3$~$\uparrow$
    \\\midrule\midrule
    \rowcolor{gray!10}\multicolumn{8}{c}{MonoViT-S} 
    \\\midrule
    + AT & $0.135$ & $1.066$ & $5.258$ & $0.215$ & $0.829$ & $0.942$ & $0.976$
    \\
    + AT + Ensemble & $0.127$ & $0.942$ & \underline{$5.043$} & \underline{$0.205$} & \underline{$0.844$} & \underline{$0.948$} & $\mathbf{0.979}$
    \\\midrule\midrule
    \rowcolor{gray!10}\multicolumn{8}{c}{MonoViT-B}
    \\\midrule
    + AT & $0.130$ & $1.027$  & $5.281$ & $0.213$ & $0.839$ & $0.945$ & $0.975$
    \\
    + AT + Ensemble & \underline{$0.126$} & \underline{$0.917$} & $5.115$ & $0.206$ & $0.842$ & \underline{$0.948$} & \underline{$0.978$}
    \\\midrule\midrule
    \rowcolor{gray!10}\multicolumn{8}{c}{MonoViT-S + MonoViT-B}
    \\\midrule
    + AT + Ensemble & $\mathbf{0.123}$ & $\mathbf{0.885}$ & $\mathbf{4.983}$ & $\mathbf{0.201}$ & $\mathbf{0.848}$ & $\mathbf{0.950}$ & $\mathbf{0.979}$
    \\\bottomrule
  \end{tabular}
}
\label{tab:track1_3rd_ensemble}
\end{table*}

\noindent\textbf{Model Ensemble}.
We evaluate the performance of MonoViT-S and MonoViT-B with and without model ensemble and show the results in Table~\ref{tab:track1_3rd_ensemble}. We observe that such a simple model fusion strategy introduces depth prediction improvements of $3\%$ and $6\%$, respectively. Furthermore, given that different model sizes could cause a model to focus on different features, we combined MonoViT-S and MonoViT-B through ensemble learning to achieve the best possible performance. This validates the effectiveness of the model ensemble in improving the robustness of depth estimation models. 

\subsubsection{Solution Summary}
In this work, we proposed a joint adversarial training approach along with a model ensemble strategy for enhancing the robustness of self-supervised depth estimation models. The adversarial samples introduced during training help reduce the sensitivity of the model to minimal perturbations in the input data, thereby improving the model performance on corrupted scenarios while still maintaining its original performance on ``clean'' datasets. Built upon the strong MonoViT baselines, our approaches achieved promising depth estimation results in this challenging competition. Our team ranked third in the first track of the RoboDepth Challenge.

\subsection{The Innovation Prize Solution: Ensemble}
\noindent\textbf{Authors:} \textcolor{gray}{Jiale Chen and Shuang Zhang.}

\begin{framed}
    \textbf{Summary} - Observing distinct behaviors of OoD corruptions in the frequency domain, the \texttt{Ensemble} team proposes two stand-alone models for robust depth estimation. The main idea is to improve the OoD generalizability of depth estimation models from two aspects: normalization and augmentation. To incorporate this, amplitude-phase recombination and feature interaction modules are proposed. The effectiveness of each model has been verified and analyzed. A further combination of both models contributes to an enhanced depth estimation robustness.
\end{framed}

\subsubsection{Overview}
Performing self-supervised depth estimation under common corruptions and sensor failure is of great value in practical applications. In this work, we propose two model variants built respectively upon MonoViT \cite{zhao2021monovit} and Lite-Mono \cite{zhang2023litemono} and improve their robustness to tackle OoD scenarios. We further propose a simple yet effective approach for the model ensemble to meet better performance on the challenging OoD depth estimation benchmark. It is worth noting that our method is the only one that trained without an extra pre-trained model; we also do not use any image pre-processing or post-processing operations in this competition.

\subsubsection{Technical Approach}

We contribute two stand-alone solutions for robust self-supervised depth estimation: \texttt{Model-I} and \texttt{Model-II}. The first model adopts MonoViT \cite{zhao2021monovit} as the backbone and is enhanced with a better normalization technique and an amplitude-phase recombination augmentation. The second model, on the other hand, is built upon Lite-Mono \cite{zhang2023litemono} and has been integrated with a double-path architecture for better feature extraction, a median-normalization for better OoD generalization, and a channel perturbation for stronger augmentation.

\noindent\textbf{Normalization}.
For \texttt{Model-I}, we change the normalization mechanism of the conventional CNN layers for robustness enhancement. The original network uses batch normalization (\texttt{BN}), which includes parameters containing information related to the batch dimension. Such in-distribution parameters, however, can have an impact on OoD testing. Inspired by AdaIN \cite{huang2017arbitrary}, which performs style transfer by controlling the normalization process at the channel level of the feature maps, we replace batch normalization with instance normalization (\texttt{IN}). It has been proven that \texttt{IN} performs individual normalization for each channel, thus achieving stable domain information for OoD testing.

\begin{figure}
    \centering
    \includegraphics[width=0.9\linewidth]{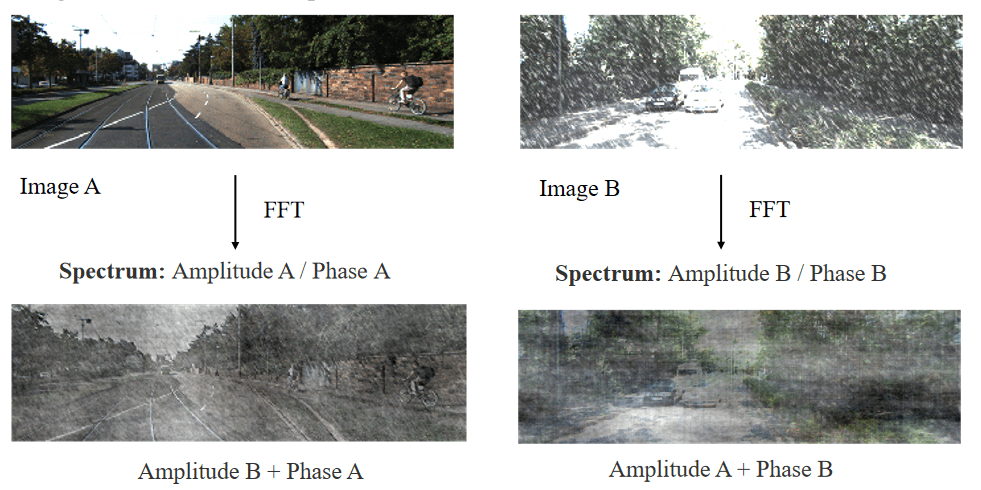}
    \caption{Illustrative examples of amplitude-phase exchange and recombination.}
\label{fig:track1_innov1_apr}
\end{figure}

\noindent\textbf{Amplitude-Phase Recombination}.
We employ the amplitude-phase recombination (APR) \cite{chen2021amplitude} as data augmentation for enhancing the model's robustness. Figure~\ref{fig:track1_innov1_apr} provides representative examples of this APR operation on the KITTI dataset \cite{geiger2012kitti}. By exchanging the magnitude and phase spectra between different style images and performing inverse Fourier transform, we discover that the phase spectra contain more shape information, while the magnitude spectra contain more texture and style information. We utilize this single-image transformation and magnitude-phase exchange to construct ARP samples during training.

\noindent\textbf{Lite Backbone}.
For \texttt{Model-II}, we aim at utilizing a lightweight model for robust depth estimation. Backbones with fewer parameters have lower capacity and weaker fitting capabilities. However, they may exhibit greater robustness and perform better in handling unknown data distributions. Our second model variant selects Lite-Mono \cite{zhang2023litemono} as the basic backbone and we make further changes to it to improve the overall robustness.

\begin{figure}
  \centering
  \subcaptionbox{}
    {\includegraphics[width=0.3\linewidth]{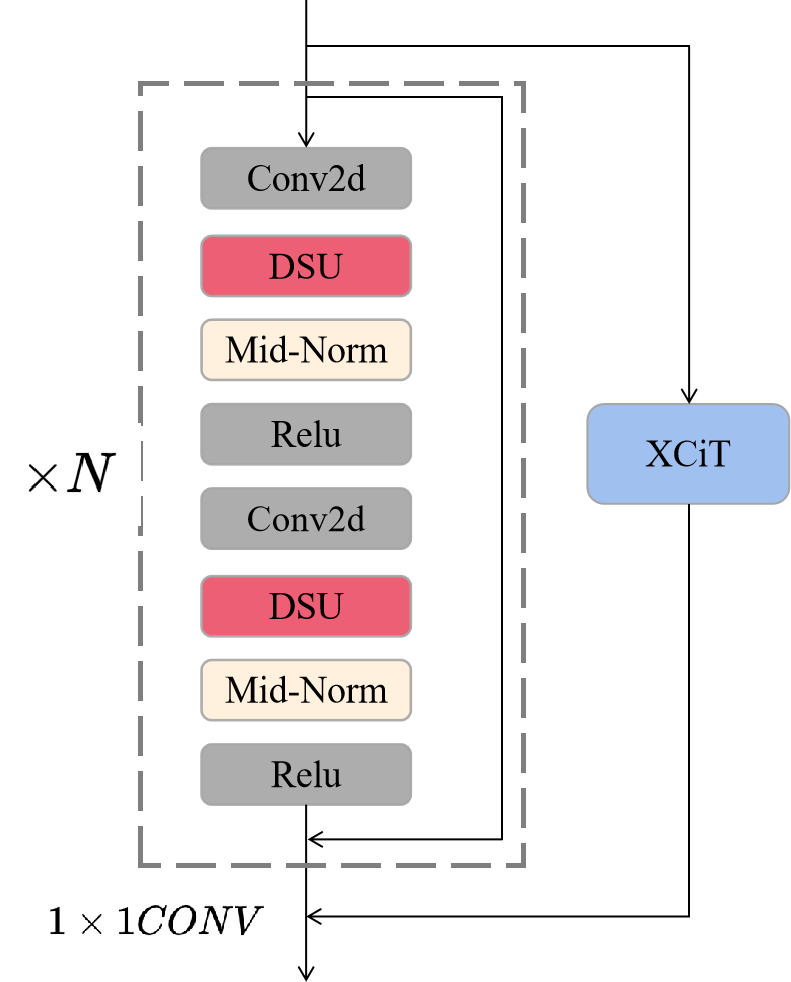}}
  \subcaptionbox{}
    {\includegraphics[width=0.69\linewidth]{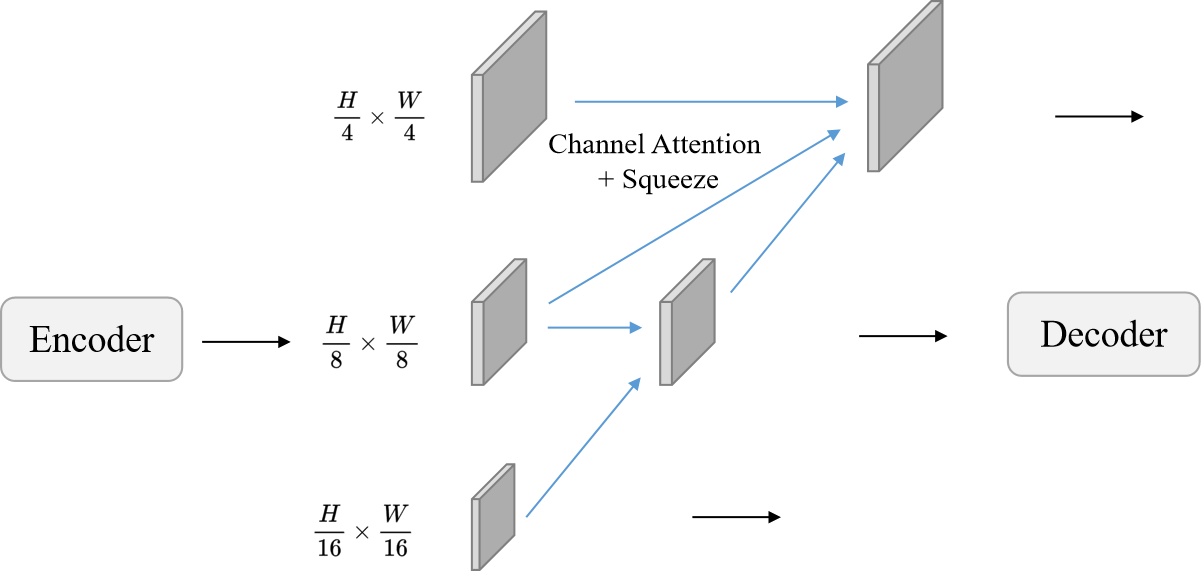}}
  \caption{Illustrative examples of the two main components in \texttt{Model-II}. (a) The double-path architecture. (b) The feature interaction module from semantics to texture.}
\label{fig:track1_innov1_framework}
\end{figure}

\noindent\textbf{Double-Path Architecture}.
CNNs have exhibited a heightened sensitivity towards local information, whereas visual Transformers demonstrate a greater aptitude for capturing global information. It is widely observed that various types of corruptions manifest significant dissimilarities in their frequency domain distributions. Consequently, a deliberate selection has been made to adopt a double-path architecture whereby distinct CNN and Transformer pathways are employed to extract features independently, followed by a subsequent feature aggregation step. Figure~\ref{fig:track1_innov1_framework}~(a) provides an example of the dual-path structure used in our network.

\noindent\textbf{Median-Normalization for OoD Generalization}.
In our framework, we propose a simple median-normalization method to facilitate better OoD generalizability. The feature map from the CNN layer is first divided into $4 \times 4$  patches, and the median value of each patch is selected for computing the mean and variance values of the channel.

\noindent\textbf{Domain \& Style Perturbation in Channel}.
For CNNs, the mean and variance of each channel represent domain and style information. Following DSU \cite{li2022uncertainty}, in the training process, we resample the mean and variance of the feature maps' channels outputted by the CNN. This allows the depth estimation model to utilize different domain and style distributions during training.

\noindent\textbf{Feature Interaction from Semantics to Texture}.
For a pyramid-shaped network architecture, shallow features contain more texture information, while deep features contain more semantic information. In the case of OoD corruptions, shallow texture features are often heavily affected, while deep semantic features exhibit higher robustness degrees. Therefore, we propose modules for aggregating information from semantics to texture before feeding the features into the depth decoder. Figure~\ref{fig:track1_innov1_framework}~(b) provides an example of our feature interaction modules. The high-level feature map is upsampled bilinearly and concatenated with the low-level one. Channel attention from CBAM \cite{woo2018cbam} and $1\times1$ convolution is adopted for fusion and channel squeeze.

\noindent\textbf{Model Ensemble}.
As will be discussed in the following section, \texttt{Model-II} is of a more stable performance compared with \texttt{Model-I}. To leverage the advantages of both models, we propose a simple yet effective approach for the model ensemble. The final depth prediction $D_{\text{final}}$ is the aggregation of predictions ($D_1$ and $D_2$) from both models, with fusion coefficients $\alpha$, $\beta$, and $\eta$. Specifically, the ensemble adopts the following formulation:
\begin{align}
D_{\text{final}}= \begin{cases}\frac{1}{\alpha \frac{1}{D_1}+\beta \frac{1}{D_2}}, & \text{where}~ \left|\frac{1}{D_1}-\frac{1}{D_2}\right| / \frac{1}{D_2}<\eta \\ D_2, & \text{where}~ \left|\frac{1}{D_1}-\frac{1}{D_2}\right| / \frac{1}{D_2} \geq \eta\end{cases}~.
\end{align}

\noindent\textbf{Training Loss}.
In addition to the conventional monocular self-supervised losses used in MonoDepth2 \cite{godard2019monodepth2}, our overall framework is trained with the proposed APR loss. The APR loss measures the L-1 distance between the disparities estimated from the raw image ($D$) and the augmented image ($D_{A P R}$) as follows:
\begin{align}
\mathcal{L}_{A P R}=\left\|\frac{1}{D}-\frac{1}{D_{A P R}}\right\|_1~.
\end{align}

\subsubsection{Experimental Analysis}

\noindent\textbf{Implementation Details}.
Our model is trained on four NVIDIA Tesla V100 GPUs. The AdamW optimizer \cite{loshchilov2018adamw} is adopted and the learning rate is set to $1$e-$4$. \texttt{Model-I} is fine-tuned for $40$ epochs with the pre-trained weights from MonoViT \cite{zhao2021monovit}. \texttt{Model-II} is trained without APR loss for $83$ epochs and a further $244$ epochs with APR loss. The parameters of model ensemble are set as $\alpha = \frac{2}{3}$,  $\beta = \frac{1}{3}$, and $\eta = 0.45$, respectively.

\begin{table*}[t]
\caption{Quantitative results of the baselines and our proposed approaches on the RoboDepth competition leaderboard (Track \# 1). The \textbf{best} and \underline{second best} scores of each metric are highlighted in \textbf{bold} and \underline{underline}, respectively.}
\centering\scalebox{0.78}{
    \begin{tabular}{l|cccc|ccc}
    \toprule
    \textbf{Method} & \cellcolor{blue!10}\textbf{Abs Rel~$\downarrow$} & \cellcolor{blue!10}\textbf{Sq Rel~$\downarrow$} & \cellcolor{blue!10}\textbf{RMSE~$\downarrow$} & \cellcolor{blue!10}\textbf{log RMSE~$\downarrow$} & \cellcolor{red!10}$\delta<1.25$~$\uparrow$ & \cellcolor{red!10}$\delta<1.25^2$~$\uparrow$ & \cellcolor{red!10}$\delta<1.25^3$~$\uparrow$
    \\\midrule\midrule
    \rowcolor{gray!10}\multicolumn{8}{c}{\texttt{Model-I}}
    \\\midrule
    MonoViT & $0.172$ & $1.340$ & $6.177$ & $0.258$ & $0.743$ & $0.910$ & $0.963$
    \\
    + APR & $0.140$ & $1.216$ & $5.448$ & $0.221$ & $0.830$ & $0.939$ & $0.974$
    \\
    + APR + BN $\rightarrow$ IN	& \underline{$0.129$} & $1.007$ & \underline{$5.066$} & \underline{$0.208$} & \underline{$0.849$} & \underline{$0.948$} & $0.977$
    \\\midrule\midrule
    \rowcolor{gray!10}\multicolumn{8}{c}{\texttt{Model-II}}
    \\\midrule
    Lite-Mono & $0.199$ & $1.642$ & $6.937$ & $0.293$ & $0.681$ & $0.880$ & $0.948$
    \\
    Lite-Mono-8m & $0.196$ & $1.569$ & $6.708$ & $0.287$ & $0.684$ & $0.884$ & $0.952$
    \\
    + Interact + Perturb & $0.133$ & \underline{$0.942$} & $5.115$ & $0.212$ & $0.832$ & $0.944$ & \underline{$0.978$}
    \\\midrule\midrule
    \rowcolor{gray!10}\multicolumn{8}{c}{\texttt{Model-I} \& \texttt{Model-II}}
    \\\midrule
    \textbf{Ensemble} & $\mathbf{0.124}$ & $\mathbf{0.871}$ & $\mathbf{4.904}$ & $\mathbf{0.202}$ & $\mathbf{0.851}$ & $\mathbf{0.951}$ & $\mathbf{0.980}$
    \\\bottomrule
\end{tabular}
}
\label{tab:track1_innov1_results}
\end{table*}

\noindent\textbf{Main Results}.
Table~\ref{tab:track1_innov1_results} shows the comparative and ablation results of \texttt{Model-I}, \texttt{Model-II}, and the fusion between them. For \texttt{Model-I}, we observe that the amplitude-phase recombination operation and statistical normalization help improve the depth estimation performance over the baseline MonoViT \cite{zhao2021monovit}. For \texttt{Model-II}, we can see that the double-path feature interaction and median-normalization modules are conducive to enhancing Lite-Mono \cite{zhang2023litemono} under OoD scenarios. Finally, an ensemble of both \texttt{Model-I} and \texttt{Model-II} brings a significantly positive impact on the robustness of self-supervised depth estimation models.

\subsubsection{Solution Summary}
In this work, we proposed two stand-alone models for robustness enhancement: \texttt{Model-I} adopted an amplitude-phase recombination operation and instance normalization for noise suppression; \texttt{Model-II} are equipped with a dual-path architecture with median-normalization, channel perturbation, and feature interaction for OoD generalization enhancement. As a result, our team achieved the innovative prize in the first track of the RoboDepth Challenge.

\subsection{The Innovation Prize Solution: Scent-Depth}
\noindent\textbf{Authors:} \textcolor{gray}{Runze Chen, Haiyong Luo, Fang Zhao, and Jingze Yu.}

\begin{framed}
    \textbf{Summary} - The lack of structural awareness in existing depth estimation systems can lead to significant performance degradation when faced with OoD situations. The \texttt{Scent-Depth} team resorts to structural knowledge distillation to tackle this challenge. A novel graph-based knowledge distillation framework is built, which is able to transfer structural knowledge from a large-scale semantic model to a monocular depth estimation model. Followed by an ensemble between semantic and depth models, the robustness of depth estimation is largely enhanced.
\end{framed}

\subsubsection{Overview}

Single-image depth estimation, also known as monocular depth estimation, is a popular area of research in computer vision due to its diverse range of applications in robotics, augmented reality, and autonomous driving \cite{dong2022survey,ming2021survey,zhao2020survey}. Despite considerable efforts made, accurately estimating the depth of objects from a single 2D image remains a challenging task due to the inherent ill-posed nature of this problem \cite{eigen2014depth}. Models that rely solely on pixel-level features struggle to capture the critical structural information of objects in a scene, which negatively impacts their performance in complex and noisy real-world environments. This lack of structural awareness can lead to significant performance degradation when faced with external disturbances such as occlusion, adverse weather, equipment malfunction, and varying lighting conditions. Therefore, effectively integrating structural information into the model becomes a crucial aspect of enhancing its depth estimation performance in various practical scenarios.

Recovering the 3D structure of a scene from just a single 2D image is difficult. However, researchers have developed unsupervised learning methods that leverage image reconstruction and view synthesis. Through the use of a warping-based view synthesis technique with either monocular image sequences or stereo pairs, the model can learn fundamental 3D object structural features, providing an elegant solution for monocular depth estimation. This approach has been previously described in academic literature \cite{zhou2017sfm,godard2017unsupervised}. Recent research has shown that the combination of Vision Transformers (ViT) \cite{dosovitskiy2020vit} and convolutional features can significantly enhance the modeling capacity for long-range structural features \cite{zhang2023litemono,zhao2021monovit}. The fusion approach leverages the strengths of both ViT \cite{dosovitskiy2020vit} and convolutional features in effectively capturing structural information from images. By incorporating both features, the model can leverage the benefits of the long-range attention mechanism and convolutional features’ ability to extract local features. This approach has shown promising results in improving the accuracy of monocular depth estimation models in complex and noisy real-world environments.

Currently, large-scale vision models demonstrate impressive generalization capabilities \cite{kirillov2023segment}, enabling the effective extraction of scene structural information in various visual contexts. Transferring scene structural knowledge through knowledge distillation from these vision models possesses significant research value. Building upon the RoboDepth Challenge, we aim to design a robust single-image depth estimation method based on knowledge distillation. Specifically, we have leveraged the ample scene structural knowledge provided by large-scale vision models to overcome the limitations of prior techniques. By incorporating these insights, our approach enhances robustness to OoD situations and improves overall performance in practical scenarios.

\subsubsection{Technical Approach}
\noindent\textbf{Task Formulation}.
The main objective of monocular depth estimation is to develop a learning-based model capable of accurately estimating the corresponding depth $\hat{D}_t$ from a monocular image frame $I_t$, within the context of a monocular image sequence $I = \{..., I_t\in\mathbb{R}^{W\times H}, ...\}$ with camera intrinsic determined by $K$. However, the challenge lies in obtaining the ground truth depth measurement $D_t$, which is both difficult and expensive to acquire.

To overcome this, we rely on unsupervised learning methods, which require our monocular depth estimation approach to leverage additional scene structural information in a single image to obtain more accurate results. In monocular depth estimation, our ultimate goal is to synthesize the view $I_{t^\prime \rightarrow t}$ by using the estimated relative pose $\hat{T}_{t \rightarrow t^\prime}$ and the estimated depth map $\hat{D}_t$ with respect to the source frame $I_t^\prime$ and the target frame $I_t$. This synthesis operation can be expressed as follows:
\begin{equation}
\label{track1_innov2_eq1}
I_{t^\prime \rightarrow t} = I_{t^\prime} < \texttt{proj}(\hat{D}_t, \hat{T}_{t \rightarrow t^\prime}, K) >~,
\end{equation}
where \texttt{proj}$(\cdot)$ projects the depth $D_t$ onto the image $I_{t^\prime}$ to obtain the two-dimensional positions, while $<\cdot>$ upsamples the estimation to match the shape of $I_{t^\prime}$ and is the approximation of $I_t$ obtained by projecting $I_{t^\prime}$. The crux of monocular depth estimation is the depth structure consistency; we need to leverage the consistency of depth structure between adjacent frames to accomplish view synthesis tasks. To achieve this, we refer to \cite{zhou2017sfm,zhao2016loss} and utilize $\mathcal{L}_p$ to impose constraints on the quality of re-projected views. This learning objective is defined as follows:
\begin{equation}
\label{track1_innov2_eq2}
\mathcal{L}_p^{u,v}(I_t, I_{t^\prime \rightarrow t}) = \frac{\alpha}{2}(1 - \texttt{ssim}(I_t, I_{t^\prime \rightarrow t})) + (1-\alpha)||I_t - I_{t^\prime \rightarrow t}||_1~,
\end{equation}
\begin{equation}
\label{track1_innov2_eq3}
\mathcal{L}_p = \sum_\mu\mathcal{L}_p^{u,v}(I_t, I_{t^\prime \rightarrow t})~,
\end{equation}
where \texttt{ssim}$(\cdot)$ computes the structural similarity index measure (SSIM) between $I_t$ and $I_{t^\prime \rightarrow t}$, $\mu$ is the auto-mask of dynamic pixels \cite{godard2019monodepth2}, and $\mathcal{L}_p$ calculates the distance measurement by taking the weighted sum of
$\sum_\mu\mathcal{L}_p^{u,v}$ over all pixels $(u, v)$.

Textures on object surfaces can vary greatly and are often not directly related to their three-dimensional structure. As a result, local textures within images have limited correlation with overall scene structure, and our depth estimation model must instead focus on higher-level, global structural features. To overcome this, we adopt the method proposed in \cite{ranjan2019competitive} to model local texture independence by
utilizing an edge-aware smoothness loss, denoted as follows:
\begin{equation}
\label{track1_innov2_eq4}
\mathcal{L}_e = \sum|| \mathbf{e}^{-\nabla I_t} \cdot \nabla \hat{D}_t ||~,
\end{equation}
where $\nabla$ denotes the spatial derivative. By incorporating $\mathcal{L}_e$, our model can better learn and utilize the overall scene structural information, irrespective of local texture variations in the object
surfaces.

\noindent\textbf{Structural Knowledge Distillation}.
The visual scene structural information is vital for a wide range of visual tasks. However, feature representations of different models for distinct tasks exhibit a certain degree of structural correlation in different channels, which is not necessarily one-to-one due to task specificity. We define $A(E, F)$ as the correlation between feature channels of $E$ and $F$, where \texttt{vec} flattens a 2D matrix into a 1D vector as follows:
\begin{equation}
\label{track1_innov2_eq5}
A(E, F) = \frac{|\texttt{vec}(E) \cdot \texttt{vec}(F)^\mathbf{T}|}{||\texttt{vec}(E)||_2 \cdot ||\texttt{vec}(F)^\mathbf{T}||_2}~.
\end{equation}
Here, $A(E, F)$ represents the edge adjacency matrix for state transitions from $E$ to $F$ in the graph space, where all $C$ channels are the nodes.

To leverage this correlation between features, we propose a structure distillation loss $\mathcal{S}$ based on isomorphic graph convolutions. We use graph isomorphic networks based on convolution operations to extract features from $E$ and $F$, resulting in $F^\prime$ and $E^\prime$, respectively. We then calculate the cosine distance between $E$ and $E^\prime$, as well as between $F$ and $F^\prime$, and include these calculations in $\mathcal{S}$ as:
\begin{equation}
\label{track1_innov2_eq6}
F^\prime = \texttt{gin}(\theta^{E\rightarrow F}, F, A(F, E))~,
\end{equation}
\begin{equation}
\label{track1_innov2_eq7}
E^\prime = \texttt{gin}(\theta^{F\rightarrow E}, E, A(E, F))~,
\end{equation}
\begin{equation}
\label{track1_innov2_eq8}
\mathcal{S}(E,F) = \texttt{cosdist}(E, E^\prime) + \texttt{cosdist}(F, F^\prime)~,
\end{equation}
where \texttt{gin}$(\cdot)$ represents the graph isomorphic network function and $\theta$ refers to the parameters of the
graph isomorphic network. This approach aggregates structured information across different tasks, which enables the transfer of such structural information to aid depth estimation.

Semantic objects in a scene carry crucial structural information; the depth of a semantic object in an image exhibits a degree of continuity. To extract image $I_t$’s encoding, we use separate depth and semantic expert encoders to obtain $F^{(d)}_t$ and $E^{(s)}_t$, respectively. The depth feature $F^{(d)}_t\in\mathbb{R}^{C^\prime \times W^\prime \times H^\prime}$ and the semantic feature $E^{(s)}_t\in\mathbb{R}^{C\times L}$ of frame $t$ exhibit a structural correlation that demonstrates graph-like characteristics in the feature embedding.

We align the depth feature $F^{(d)}_t$ with the semantic feature $E^{(s)}_t$ using the alignment function \texttt{align}$(\cdot)$, which is satisfying $E^{(s)}_t = \texttt{align}(F^{(d)}_t)$. To ensure consistent node feature dimensions before constructing the graph structure, we implement the alignment mapping \texttt{align}$(\cdot)$ using bilinear interpolation and convolution layers.

To distill the feature structural information of a powerful expert semantic model to the deep depth estimation model, we employ the structure graph distillation loss $\mathcal{L}_g$, which links the structural correlation between semantic embedding and depth embedding as follows:
\begin{equation}
\label{track1_innov2_eq9}
\mathcal{L}_g = \mathcal{S}(F^{(d)}_t, E^{(s)}_t)~.
\end{equation}
It is worth noting that $\mathcal{L}_g$ enables the cross-domain distillation of semantic structural information from the semantic expert model to the depth estimation model.

\noindent\textbf{Total Loss}.
We propose a method to train a monocular depth model using semantic and structural correlation of visual scenarios. To achieve this goal, we incorporate the idea of knowledge distillation into the design of training constraints for monocular depth estimation. The overall training objective is defined as follows:
\begin{equation}
\label{track1_innov2_eq10}
\min_{\theta}\mathcal{L} = \lambda_p\mathcal{L}_p + \lambda_e\mathcal{L}_e + \lambda_g\mathcal{L}_g~,
\end{equation}
where $\{\lambda_p, \lambda_e, \lambda_g\}$ are loss weights that balance the various constraints during training. We use these weights to determine the level of importance assigned to each constraint.

\noindent\textbf{Model Ensembling}.
To further improve the overall robustness, we use different single-stage monocular depth estimation backbones to train multiple models, resulting in different model configurations $C$ and corresponding depth estimations $D^{(C)}_t$. We then employ a model ensembling approach to improve the robustness of the overall depth estimation $D_t$. The ensembling process involves combining the predictions of each individual model $D^{(C)}_t$ with equal weight, resulting in an ensemble prediction $D_t$, which we use as the final prediction. This approach leverages the diversity of the individual models and improves the overall robustness of the depth estimation. we combine these depth maps using equal weights and obtain the ensemble depth map $D_t$ as follows:
\begin{equation}
\label{track1_innov2_eq11}
D_t = \frac{1}{N}\sum\frac{D^{(C)}_t}{\texttt{median}(D^{(C)}_t)}~,
\end{equation}
where $N$ denotes the total number of configurations, and the \texttt{median}$(\cdot)$ calculates the median value of each depth map $D^{(C)}_t$.

\subsubsection{Experimental Analysis}
\noindent\textbf{Implementation Details}.
We use the publicly accessible KITTI dataset \cite{geiger2012kitti} as the training dataset. This dataset consists of images with a resolution of $1242\times375$. We down-sample each image to $640\times192$ during training. To avoid introducing any corruptions as data augmentation during the training process, we use only raw images from KITTI \cite{geiger2012kitti} for training. During training, we use a batch size of 8 and train the model on two NVIDIA V100 GPUs. We employ a cosine annealing learning rate adjustment strategy with a period of $100$ iterations, setting the minimum and maximum learning rates to $1$e-$5$ and $1$e-$4$, respectively.

\begin{table*}[t]
\caption{Quantitative results of the baselines and our proposed approaches on the RoboDepth competition leaderboard (Track \# 1). The \textbf{best} and \underline{second best} scores of each metric are highlighted in \textbf{bold} and \underline{underline}, respectively.}
\centering\scalebox{0.78}{
\begin{tabular}{l|cccc|cccc}
    \toprule
    \textbf{Method} & \cellcolor{blue!10}\textbf{Abs Rel~$\downarrow$} & \cellcolor{blue!10}\textbf{Sq Rel~$\downarrow$} & \cellcolor{blue!10}\textbf{RMSE~$\downarrow$} & \cellcolor{blue!10}\textbf{log RMSE~$\downarrow$} & \cellcolor{red!10}$\delta<1.25$~$\uparrow$ & \cellcolor{red!10}$\delta<1.25^2$~$\uparrow$ & \cellcolor{red!10}$\delta<1.25^3$~$\uparrow$
    \\\midrule\midrule
    MonoDepth2 \cite{godard2019monodepth2} & $0.221$ & $1.988$ & $7.117$ & $0.312$ & $0.654$ & $0.859$ & $0.938$
    \\
    DPT \cite{ranftl2021dpt} & $0.151$ & $1.073$ & $5.988$ & $0.237$ & $0.782$ & $0.928$ & $0.970$
    \\\midrule
    Ours (MonoDepth2) & $0.156$ & $1.185$ & $5.587$ & $0.235$ & $0.787$ & $0.932$ & $0.973$
    \\
    Ours (MonoDepth2-E) & $0.151$ & $1.058$ & \underline{$5.359$} & \underline{$0.226$} & \underline{$0.794$} & \underline{$0.935$} & \underline{$0.976$}
    \\
    Ours (MonoViT) & \underline{$0.148$} & \underline{$1.030$} & $5.582$ & $0.230$ & $0.790$ & $0.930$ & $0.974$
    \\
    Ours (MonoViT-E) & $\mathbf{0.137}$ & $\mathbf{0.904}$ & $\mathbf{5.276}$ & $\mathbf{0.214}$ & $\mathbf{0.813}$ & $\mathbf{0.941}$ & $\mathbf{0.979}$
    \\\bottomrule
    \end{tabular}
}
\label{tab:track1_innov2_results}
\end{table*}

\noindent\textbf{Comparative \& Ablation Study}.
We adopt MonoDepth2 \cite{godard2019monodepth2}, DPT \cite{ranftl2021dpt}, and MonoViT \cite{zhao2021monovit} as the baseline models in our experiments and compare them with our own models trained in different ways. The quantitative results of different methods are shown in Table~\ref{tab:track1_innov2_results}. As can be seen from the comparative results, our knowledge distillation strategy can significantly improve the depth estimation performance over the original MonoDepth2 \cite{godard2019monodepth2} and MonoViT \cite{zhao2021monovit} under corrupted scenes. We also enable the performance of MonoViT \cite{zhao2021monovit} to exceed that of the large-scale depth estimation model, DPT \cite{ranftl2021dpt}, in the challenging OoD scenarios. Additionally, the model ensembling strategy also improves the performance of depth estimation, with all indicators achieving higher performance than those of a single model setting in our robustness evaluation.

\subsubsection{Solution Summary}
In this work, we explored robust single-image depth estimation and proposed a novel knowledge distillation approach based on graph convolutional networks. We integrated information from different visual tasks for robustness enhancement and showed that the proposed knowledge distillation strategy can effectively improve the performance of MonoDepth and MonoViT and surpassed that of DPT in corrupted scenes. Additionally, the fusion of multiple models further improved the depth estimation results. Our team achieved the innovative prize in the first track of the RoboDepth Challenge.

\section{Winning Solutions from Track \# 2}
\label{sec:track2}

\subsection{The \textcolor{robo_blue}{1st} Place Solution: \textcolor{robo_blue}{USTCxNetEaseFuxi}}
\noindent\textbf{Authors:} \textcolor{gray}{Jun Yu, Mohan Jing, Pengwei Li, Xiaohua Qi, Cheng Jin, Yingfeng Chen, and Jie Hou.}

\begin{framed}
    \textbf{Summary} - Most existing depth estimation models are trained solely on ``clean'' data, thereby lacking resilience against real-world interference. To address this limitation, the \texttt{USTCxNetEaseFuxi} team incorporates CutFlip and MAEMix as augmentations to enhance the model’s generalization capabilities during training. Additionally, appropriate inpainting methods, such as image restoration and super-resolution, are selected and tailored to handle specific types of corruptions during testing. Furthermore, a new classification-based fusion approach is proposed to leverage advantages from different backbones for robustness enhancement.
\end{framed}

\subsubsection{Overview}

To fulfill the needs of real-world perception tasks such as robot vision and robot autonomous driving, significant progress has been made in the field of image depth estimation in recent years. Many high-quality datasets have been constructed by using high-performance sensing elements, such as depth cameras for depth imaging \cite{silberman2012nyu2} and LiDAR sensors for 3D perception \cite{geiger2012kitti}.

However, the current learning-based depth estimation paradigm might become too ideal. Most existing models are trained and tested on clean datasets, without considering the fact that image acquisition often happens in real-world scenes. Even high-performance sensing devices are often affected by factors such as different lighting conditions, lens jittering, and noise perturbations. These factors can disrupt the contour information of objects in the image and interfere with the determination of relative depth. Traditional methods such as filtering cannot effectively eliminate these noise interference, and existing models often lack sufficient robustness to effectively overcome these problems.

In this work, to pursue robust depth estimation against corruptions, we propose an effective solution with specific contributions as follows. Firstly, we conducted experiments on various high-performance models to compare their robustness and ultimately selected the models with the best possible robustness at present \cite{zhao2023unleashing,bhat2023zero,ning2023ait,liu2022swin-v2}. Secondly, we attempted to find a group of non-pixel-level operational data enhancement methods without simulating noise in real conditions. Next, we have chosen some new and effective image restoration methods for reconstructing corrupted images \cite{zamir2022restormer}. Finally, our proposed approach achieved first place in the second track of the RoboDepth Challenge, which proves the effectiveness of our designs.

\begin{figure}
    \centering
    \includegraphics[width=1.0\linewidth]{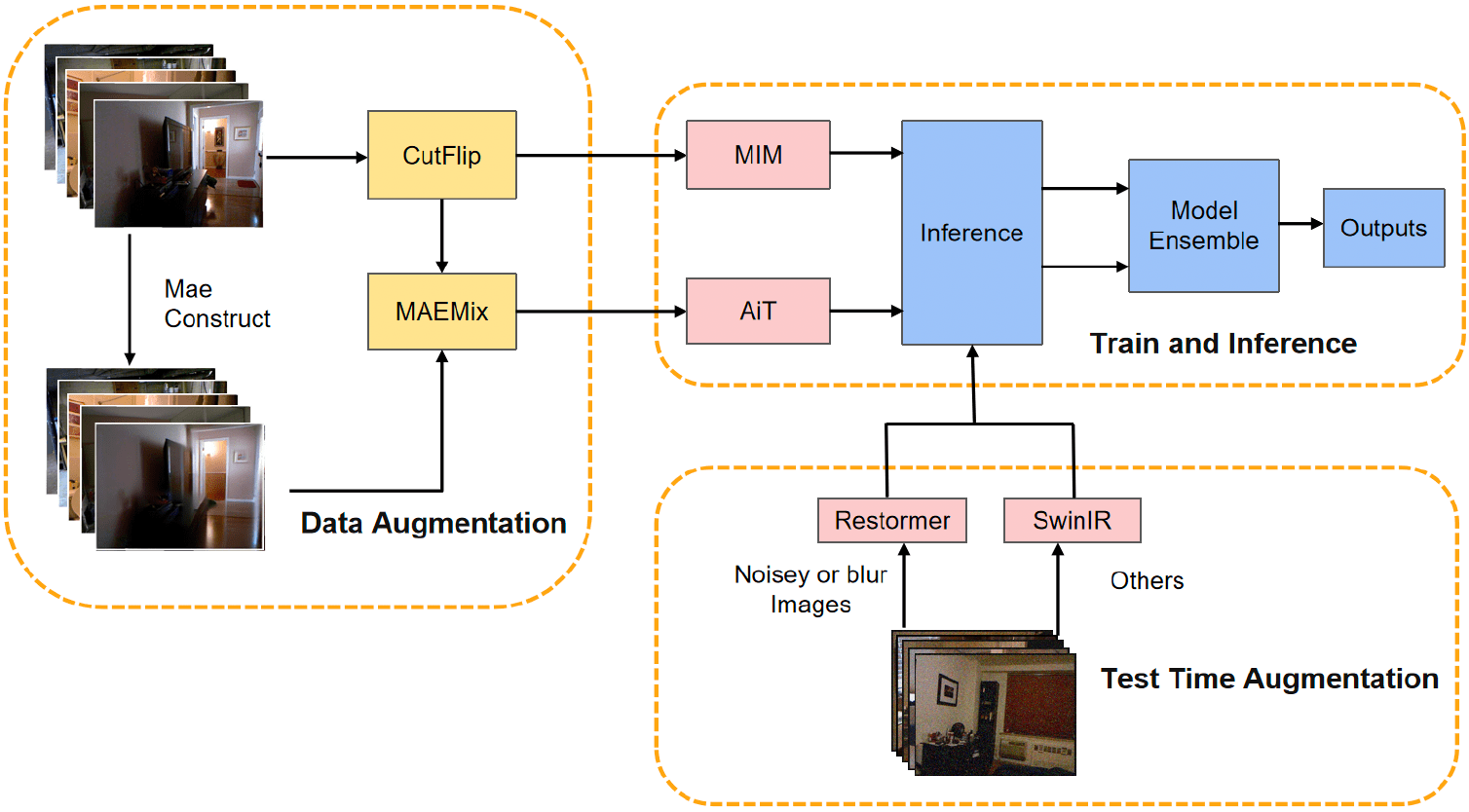}
    \caption{Overview of the proposed robust depth estimation solution. Our design consists of three main components: 1) a training and inference framework; 2) a data augmentation combo; and 3) a test-time augmentation module.}
\label{fig:track2_1st_framework}
\end{figure}

\subsubsection{Technical Approach}
Our proposed solution consists of three main components: 1) the overall framework, 2) training augmentations, and 3) post-processing techniques. We first elaborate on the overall architecture of our solution. We then describe the two data augmentation techniques involved in our solution. The third component, post-processing, is introduced, including a test-time data processing method and a model fusion approach.

\noindent\textbf{Robust Depth Estimation Framework}.
Our overall model architecture is depicted in Figure~\ref{fig:track2_1st_framework}. We select four existing depth estimation models that have demonstrated outstanding performance on the NYU Depth V2 \cite{silberman2012nyu2} dataset for comparative experiments. Among them, AiT \cite{ning2023ait} exhibited the highest robustness and was found to be highly suitable for tackling this challenging task. MIM-Depth \cite{xie2023revealing} also exhibited relatively superior performance. Consequently, these two models were chosen for subsequent experiments. VPD \cite{zhao2023unleashing} focused on learning advanced semantic information about the scene during training, which was lacking fine-grained capabilities in handling the corrupted images. On the other hand, Zoe \cite{bhat2023zero} emphasized learning relative depth, but the presence of certain corruption types, such as zoom blur, inevitably resulted in significant errors.

During training, we employed two data augmentation techniques: CutFlip and MAEMix. CutFlip involves horizontally cutting the input image and swapping the upper and lower halves. MAEMix, on the other hand, entails simple image blending between the original image and the reconstructed image using a mask. In the testing phase, we applied denoising, deblurring, and super-resolution techniques to the corrupted images to enhance their quality and mitigate the effects brought by noises and interference.

\noindent\textbf{CutFlip}.
We adopt a simple CutFlip operation during training to enhance data diversity. We vertically split the input images into the upper and lower halves, and then flip these two parts along the vertical axis. This process helps to weaken the correlation between the depth and the vertical position of the image. The probability of applying CutFlip is set to $0.5$ and the vertical splitting position is randomly sampled, allowing the model to adapt well to various types of training data.

\noindent\textbf{MAEMix}.
In fact, MAE-based data processing can serve as a powerful data augmentation technology \cite{he2022mae}. The realization of MAE is simple: masking out random patches on the input images and reconstructing the masked regions based on the remaining visual cues. Empirically, masking out most of the input images (such as $75\%$) will form an important and meaningful self-supervised learning task. Strictly speaking, the MAE method belongs to a denoising autoencoder (DAE). The denoising operation in DAE belongs to a kind of representation learning, which destroys the input signal and learns to reconstruct the original and undamaged signals. The encoder and decoder structures of MAE are different and asymmetric. The encoder often encodes the input as a latent representation, while the decoder reconstructs the original signal from this latent representation.

The reconstructed image will have a decrease in clarity compared to the original image, and the image content will also undergo certain changes, which to some extent aligns with our idea of enhancing the model's robustness. An effective approach is to mix the reconstructed image with the original image, thereby transferring the disturbance introduced by MAE reconstruction to the original image. This process helps to incorporate the variations and distortions captured by MAE into the original input, resulting in enhanced feature learning and improved overall robustness of the depth estimation model.

\noindent\textbf{Post-Processing}.
For the test time augmentation, our research focus lies on image restoration operations. The testing set comprises heavily interfered and damaged images. Observing the test set, it was found that noises and blurs accounted for a significant proportion of corruptions, while weather-related corruptions were rarely seen, with only a small number of images showing corruption effects similar to fog. Indeed, as the NYU Depth V2 dataset \cite{silberman2012nyu2} is mainly constructed for indoor scenes, such indoor environments are rarely affected by adverse weather conditions in practical situations. Hence we focus on noise corruptions and blur corruptions during the post-processing.

Before performing image reconstruction, we pre-classified the test set, categorizing different noises and blurs into pre-defined categories, while the remaining images were mainly compressed image quality and color corruptions, which were all classified into another category together.

For images with various types of noises and blurs, we utilized Restormer \cite{zamir2022restormer} for repairing. Restormer \cite{zamir2022restormer} has achieved state-of-the-art results in multiple image restoration tasks, including image de-snowing, single image motion de-blurring, defocus de-blurring (single image and dual pixel data), and image de-noising, outperforming networks such as SwinIR \cite{liang2021swin-ir} and IPT \cite{chen2021ipt}.

\begin{table*}[t]
\caption{Quantitative results of the baselines and different data augmentation techniques on the RoboDepth competition leaderboard (Track \# 2). The \textbf{best} and \underline{second best} scores of each metric are highlighted in \textbf{bold} and \underline{underline}, respectively.}
\centering\scalebox{0.78}{
\begin{tabular}{c|c|c|c|c}
    \toprule
    \textbf{Method} & \textbf{Ref} & \textbf{Data Augmentation} & \cellcolor{red!10}$\delta<1.25$~$\uparrow$ & $\Delta$
    \\\midrule\midrule
    VPD & \cite{zhao2023unleashing} & \textcolor{gray}{None} & $0.743$ & -
    \\\midrule
    ZoeD-M12-N & \cite{bhat2023zero} & \textcolor{gray}{None} & $0.875$ & -
    \\\midrule
    \multirow{4}{*}{AiT-P} & \multirow{4}{*}{\cite{ning2023ait}} & \textcolor{gray}{None} & $0.903$ & \textcolor{gray}{$+0.000$}
    \\
    & & CutFlip & $0.900$ & \textcolor{robo_blue}{$-0.003$}
    \\
    & & MAEMix & $0.902$ & \textcolor{robo_blue}{$-0.001$}
    \\
    & & CutFlip + MAEMix & $0.903$ & \textcolor{gray}{$+0.000$}
    \\\midrule
    \multirow{4}{*}{SwinV2-L 1K-MIM-Depth} & \multirow{4}{*}{\cite{liu2022swin-v2}} & \textcolor{gray}{None} & $0.887$ & \textcolor{gray}{$+0.000$}
    \\
    & & CutFlip & $0.897$ & \textcolor{robo_red}{$+0.010$}
    \\
    & & MAEMix & $\mathbf{0.915}$ & \textcolor{robo_red}{$+0.028$}
    \\
    & & CutFlip + MAEMix & \underline{$0.905$} & \textcolor{robo_red}{$+0.018$}
    \\\bottomrule
    \end{tabular}
}
\label{tab:track2_1st_augmentation}
\end{table*}

On the other hand, for other images, we employed SwinIR \cite{liang2021swin-ir} for super-resolution processing. SwinIR \cite{liang2021swin-ir} has exhibited excellent performance in dealing with image compression and corruption, which can significantly improve image quality. However, color destruction, due to its inherent difficulty in recovery, can only receive a small amount of improvement.

Furthermore, our attempts to utilize Mean Absolute Error for image reconstruction during inference yielded unsatisfactory results. Similarly, we conducted multi-scale testing using super-resolution techniques, but the outcomes were sub-optimal. We speculate that the underwhelming performance of the Mean Absolute Error metric and super-resolution techniques may be attributed to their reliance on algorithmic assumptions to generate image features, rather than capturing genuine content. We conjecture that the discrepancy between algorithmic assumptions and real content could have contributed to the cause of these sub-optimal results.

\subsubsection{Experimental Analysis}
\noindent\textbf{Baselines}.
We selected four state-of-the-art depth estimation models \cite{zhao2023unleashing,bhat2023zero,ning2023ait,liu2022swin-v2} in our experiments. The inference results of these models on the RoboDepth Challenge testing set are presented in Table~\ref{tab:track2_1st_augmentation}. Among all models tested, AiT \cite{ning2023ait} and MIM-Depth \cite{xie2023revealing} demonstrated relatively superior performance, making them highly suitable for handling the challenging OoD depth estimation task. AiT \cite{ning2023ait} itself adopts mask authentication techniques similar to that in natural language processing hence it has exhibited a certain degree of robustness. MIM-Depth \cite{xie2023revealing} balances the global attention and local attention well, making it perform more evenly on the entire image.

\noindent\textbf{Data Augmentation Techniques}.
The results of our ablation experiments on the proposed MAEMix and CutFlip are presented in Table~\ref{tab:track2_1st_augmentation}. We observe that the optimal data augmentation combinations differ for AiT \cite{ning2023ait} and MIM-Depth \cite{xie2023revealing}. After applying data augmentation, AiT \cite{ning2023ait} achieved a depth estimation performance (in terms of the $\delta_1$ score) of $90.3\%$, while MIM-Depth \cite{xie2023revealing} reached a $\delta_1$ accuracy of $91.5\%$.

\begin{table*}[t]
\caption{Quantitative results of the baselines and different post-processing techniques on the RoboDepth competition leaderboard (Track \# 2). The \textbf{best} and \underline{second best} scores of each metric are highlighted in \textbf{bold} and \underline{underline}, respectively.}
\centering\scalebox{0.78}{
\begin{tabular}{c|c|c|c|c}
    \toprule
    \textbf{Method} & \textbf{Ref} & \textbf{Post-Processing} & \cellcolor{red!10}$\delta<1.25$~$\uparrow$ & $\Delta$
    \\\midrule\midrule
    \multirow{3}{*}{AiT-P} & \multirow{3}{*}{\cite{ning2023ait}} & \textcolor{gray}{None} & $0.903$ & \textcolor{gray}{$+0.000$}
    \\
    & & Restormer & $0.921$ & \textcolor{robo_red}{$+0.018$}
    \\
    & & Restormer + SwinIR & $0.922$ & \textcolor{robo_red}{$+0.019$}
    \\\midrule
    \multirow{3}{*}{SwinV2-L 1K-MIM-Depth} & \multirow{3}{*}{\cite{liu2022swin-v2}} & \textcolor{gray}{None} & $0.887$ & \textcolor{gray}{$+0.000$}
    \\
    & & Restormer & \underline{$0.924$} & \textcolor{robo_red}{$+0.037$}
    \\
    & & Restormer + SwinIR & $\mathbf{0.929}$ & \textcolor{robo_red}{$+0.042$}
    \\\bottomrule
    \end{tabular}
}
\label{tab:track2_1st_post}
\end{table*}

\begin{table*}[t]
\caption{Quantitative results of the baselines and different model ensemble techniques on the RoboDepth competition leaderboard (Track \# 2). The \textbf{best} and \underline{second best} scores of each metric are highlighted in \textbf{bold} and \underline{underline}, respectively.}
\centering\scalebox{0.78}{
\begin{tabular}{c|c|c|c|c}
    \toprule
    \textbf{Method} & \textbf{Ref} & \textbf{Model Ensemble} & \cellcolor{red!10}$\delta<1.25$~$\uparrow$ & $\Delta$
    \\\midrule\midrule
    AiT-P & \cite{ning2023ait} & \textcolor{gray}{None} & $0.903$ & $+0.000$
    \\\midrule
    \multirow{2}{*}{AiT-P + MIM-Depth} & \multirow{2}{*}{\cite{xie2023revealing}} & Weighted Average Ensemble & \underline{$0.933$} & $+0.030$
    \\
    & & Classification Ensemble & $\mathbf{0.940}$ & $+0.037$
    \\\bottomrule
    \end{tabular}
}
\label{tab:track2_1st_ensemble}
\end{table*}

\noindent\textbf{Post-Processing Techniques}.
We found that the post-processing for interference reduction using image restoration and super-resolution techniques \cite{zamir2022restormer,liang2021swin-ir} led to an additional improvement in performance. As can be seen from Table~\ref{tab:track2_1st_post}, these two post-processing operations help improve the depth estimation accuracy (in terms of the $\delta_1$ score) with scores reaching $92.20\%$ and $92.87\%$ for AiT \cite{ning2023ait} and MIM-Depth \cite{xie2023revealing}, respectively.

\noindent\textbf{Model Ensemble Techniques}.
Finally, we perform model fusion on the previously obtained results from AiT \cite{ning2023ait} and MIM-Depth \cite{xie2023revealing} using the following strategies: 1) Weighted Average Ensemble, and 2) Classification Ensemble approaches. As can be seen from Table~\ref{tab:track2_1st_ensemble}, the former ensemble strategy achieved a final $\delta_1$ score of $93.3\%$ while the latter one yielded an accuracy $94.0\%$, which is $0.0037$ higher than the baseline.

\subsubsection{Solution Summary}
In this work, we have demonstrated through extensive experiments that incorporating the CutFlip and MAEMix data augmentation techniques during the training process brings a positive effect on enhancing the depth estimation model's robustness. Additionally, the simple and effective inpainting approaches, \textit{i.e.} Restormer and SwinIR, directly improve the depth prediction results under OoD corruptions. Our classification-based model fusion method fully considers the advantages of different depth estimation modes and achieves better results than the traditional fusion approaches, which helped us achieve satisfactory results in the challenge competition. Our team ranked first in the second track of the RoboDepth Challenge.

\subsection{The \textcolor{robo_red}{2nd} Place Solution: \textcolor{robo_red}{OpenSpaceAI}}
\noindent\textbf{Authors:} \textcolor{gray}{Li Liu, Ruijie Zhu, Ziyang Song, and Tianzhu Zhang.}

\begin{framed}
    \textbf{Summary} - The \texttt{OpenSpaceAI} team proposes a Robust Diffusion model for Depth estimation (RDDepth) to address the problem of single-image depth estimation on OoD datasets. RDDepth takes the use of VPD as the baseline for utilizing the denoising capability of the diffusion model, which is naturally suitable for handling such a problem. Additionally, the high-level scene priors provided by the text-to-image diffusion model are leveraged for robust predictions. Furthermore, the AugMix data augmentation is incorporated to further enhance the model’s robustness.
\end{framed}

\subsubsection{Overview}
Monocular depth estimation is a fundamental task in computer vision and is crucial for scene understanding and other downstream applications. In real practice, there are inevitably some corruptions (\textit{e.g.} rain), which hinder safety-critical applications. Many learning-based monocular depth estimation methods \cite{bhat2021adabins,li2022binsformer,yuan2022newcrfs,li2022depthformer,tang2021transdepth,liu2015deep} train and evaluate in the subsets of an individual benchmark. Therefore, they tend to overfit a specific dataset, which leads to poor performance on OoD datasets. The second track of the RoboDepth Challenge provides the necessary data and toolkit for the supervised learning-based model to handle OoD depth estimation. The objective is to accurately estimate the depth information while training only on the clean NYU Depth V2 \cite{silberman2012nyu2} dataset. Our goal is to improve the model’s generalization ability across real-world OoD scenarios.

To address this issue, we propose a Robust Diffusion model for Depth estimation (RDDepth). RDDepth takes VPD \cite{zhao2023unleashing} as the baseline, which aims to leverage the high-level knowledge learned in the text-to-image diffusion model for visual perception. We believe the knowledge from VPD \cite{zhao2023unleashing} can also benefit the robustness of depth predictors since the prior of scenes is given. Moreover, the denoising capability of diffusion is naturally suitable for handling OoD situations.

Instead of using the step-by-step diffusion pipeline, we simply employ the autoencoder as a backbone model to directly consume the natural images without noise and perform a single extra denoising step with proper prompts to extract the semantic information. Specifically, RDDepth takes the RGB image as input and extracts features by the pre-trained encoder of VQGAN \cite{esser2021vqgan}, which projects the image into the latent space. The text input is defined by the template of \textit{``a photo of a [CLS]''}, and then the CLIP \cite{radford2021clip} text encoder is applied to obtain text features.

To solve the domain gap when transferring the text encoder to depth estimation, we adopt an adapter to refine the text features obtained by the CLIP \cite{radford2021clip}. The latent feature map and the refined text features are then fed into UNet \cite{ronneberger2015unet} to obtain hierarchical features, which are used by the depth decoder to generate the final depth map. In addition, we employed the AugMix \cite{hendrycks2020augmix} data augmentation, which does not include any of the $18$ types of corruption and their atomic operations in the original RoboDepth benchmark. We find that, within a certain range, more complex data augmentation enables the model to learn more robust scene priors, thereby enhancing its generalization when tested on corrupted data.

\begin{figure}
    \centering
    \includegraphics[width=1.0\linewidth]{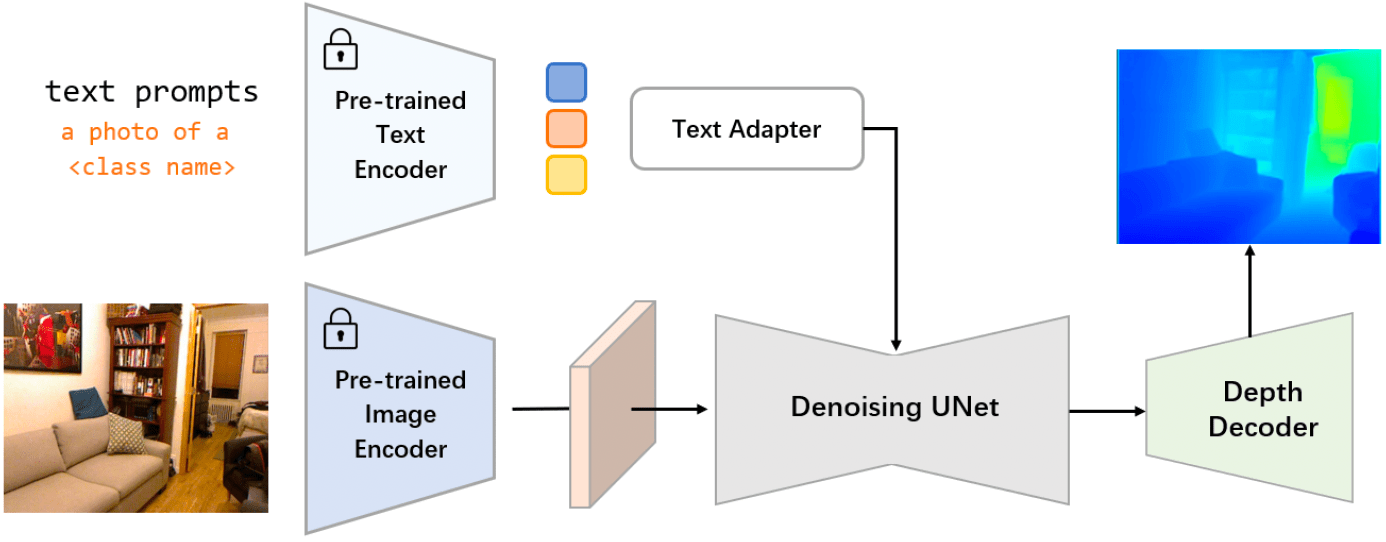}
    \caption{The architecture of RDDepth. The proposed RDDepth firstly uses a pre-trained image encoder to project RGB image into the latent space, meanwhile extracting the corresponding text feature. The text adapter is used to tackle the domain gap between the text and depth estimation tasks. UNet is considered a backbone to provide hierarchical features in our framework.}
    \label{fig:track2_2nd_framework}
\end{figure}

\subsubsection{Technical Approach}
In this section, we present RDDepth, a framework that achieves promising robustness in OoD depth estimation scenarios. RDDepth is built upon VPD \cite{zhao2023unleashing} -- a framework that exploits the semantic information of a pre-trained text-to-image diffusion model in visual perception tasks. The key idea of VPD \cite{zhao2023unleashing} is to investigate how to fully extract the pre-trained high-level knowledge in a pre-trained diffusion model. We find such knowledge also benefits the prediction of corrupted images thanks to their semantic information. The overall framework of our RDDepth is illustrated in Figire~\ref{fig:track2_2nd_framework}.

\noindent\textbf{Framework Overview}.
Our RDDepth is based on VPD \cite{zhao2023unleashing}, a model that builds upon the foundation of the popular Stable Diffusion \cite{rombach2022stable} and conducts the denoising process in a learned latent space with a UNet architecture. As for Stable Diffusion \cite{rombach2022stable}, there is adequate high-level knowledge due to the weak supervision of the natural language during pre-training. We believe that this high-level knowledge can, to some extent, mitigate the influence of the corruptions in the feature space, thereby guiding the recovery of more accurate depth maps in the depth prediction head. Therefore, the key idea is to investigate how to effectively leverage the advanced knowledge of the diffusion model to steer subsequent models in monocular depth estimation.

Specifically, RDDepth firstly uses encoder $\epsilon$ in VQGAN \cite{esser2021vqgan} to extract image features and obtain the representation of latent space. Then we hope to extract corresponding text features from class names by the simple template \textit{``a photo of a [CLS]''}. Moreover, We align the text features to the image features by an adapter. This design enables us to retain the pre-trained knowledge of the text encoder to the fullest extent while reducing the domain discrepancy between the pre-training task and the depth estimation task. After that, we feed the latent feature map and the conditioning inputs to the pre-trained network (usually implemented as a UNet \cite{ronneberger2015unet}). We do not use the step-by-step diffusion pipeline, which is common in other works. Instead, we simply consider it as a backbone. In other words, no noise is added to the latent feature map during the denoising process since we set $t = 0$. Then, we use only one denoising step by UNet \cite{ronneberger2015unet} to obtain the features.

The hierarchical feature $\mathcal{F}$ can be easily obtained from the last layer of each output block in different resolutions. Typically, the size of the input image is $512\times512$; the hierarchical feature maps $\mathcal{F}$  contain four sets, where the $i$-th feature map $F_i$ has the spatial size of $H_i = W_i = 2^{i+2}$, with $i = 1,2,3,4$. The final depth map is then generated by a depth decoder, which is implemented as a semantic FPN \cite{kirillov2019fpn}.

\noindent\textbf{Data Augmentation Module}.
We exploit new data augmentation designs that are overtly different from conventional ones. In general circumstances, models can only memorize the specific corruptions seen during training, which results in poor generalization ability against corruptions. AugMix \cite{hendrycks2020augmix} is proposed for helping models withstand unforeseen corruptions. Specifically, AugMix \cite{hendrycks2020augmix} involves blending the outputs obtained by applying chains or combinations of multiple augmentation operations. Inspired by it, we investigate the effect of different data augmentation on indoor scene corruptions in our work. The augmentation operations include rotation, translation, shear, \textit{etc}. Next, we randomly sample three augmentation chains; each augmentation chain is constructed by composing from one to three randomly selected augmentation operations. This operation can prevent the augmented image from veering too far from the original image.

\noindent\textbf{Loss Function}.
We adopt the Scale-Invariant Logarithmic (SILog) loss introduced in \cite{eigen2014depth} and denote it as $L$. We first calculate the logarithm difference between the predicted depth map and the ground-truth depth as follows:
\begin{equation}
\label{track2_2nd_eq1}
\Delta d_i = \log d^{\prime}_i - \log d^{*}_i~,
\end{equation}
where $d^{\prime}_i$ and $d^{*}_i$ are the predicted depth and ground-true depth, respectively, at pixel $i$. The SIlog loss is computed as:
\begin{equation}
\label{track2_2nd_eq2}
L = \sqrt{\frac{1}{K} \sum_i \Delta d^2_i - \frac{\lambda}{K}(\sum_i \Delta d_i)^2}~,
\end{equation}
where $K$ is the number of pixels with valid depth and $\lambda$ is a variance-minimizing factor. Following previous works \cite{bhat2021adabins,li2022binsformer}, we set $\lambda = 0.5$ in our experiments.

\subsubsection{Experimental Analysis}
\noindent\textbf{Implementation Details}.
We provide the common configurations of our baseline. We fix the VQGAN \cite{esser2021vqgan} encoder $\epsilon$ and the CLIP \cite{radford2021clip} text encoder during training. To fully preserve the pre-trained knowledge, we always set the learning rate $\epsilon_\theta$ of as $1$/$10$ of the base learning rate.

Our work is implemented using PyTorch on eight NVIDIA RTX 3090 GPUs. The network is optimized end-to-end with the Adam optimizer ($\beta_1 = 0.9$, $\beta_2 = 0.999$). We set the learning rate to $5$e-$4$ and train the model for $30$ epochs with a batch size of $24$. During training, we randomly crop the images to $480\times480$ and then use AugMix \cite{hendrycks2020augmix} to obtain the augmented image. We freeze the CLIP \cite{radford2021clip} text decoder and VQGAN \cite{esser2021vqgan} encoder. The version of stable diffusion is v1-5 by default. The decoder head and other experimental settings are the same as \cite{xie2023revealing}. We use the flip and sliding windows during testing.

\begin{table*}[t]
\caption{Quantitative results on the RoboDepth competition leaderboard (Track \# 2). The \textbf{best} and \underline{second best} scores of each metric are highlighted in \textbf{bold} and \underline{underline}, respectively.}
\centering\scalebox{0.78}{
\begin{tabular}{l|cccc|cccc}
    \toprule
    \textbf{Method} & \cellcolor{blue!10}\textbf{Abs Rel~$\downarrow$} & \cellcolor{blue!10}\textbf{Sq Rel~$\downarrow$} & \cellcolor{blue!10}\textbf{RMSE~$\downarrow$} & \cellcolor{blue!10}\textbf{log RMSE~$\downarrow$} & \cellcolor{red!10}$\delta<1.25$~$\uparrow$ & \cellcolor{red!10}$\delta<1.25^2$~$\uparrow$ & \cellcolor{red!10}$\delta<1.25^3$~$\uparrow$
    \\\midrule\midrule
    YYQ & $0.125$ & $0.085$ & $0.470$ & $0.159$ & $0.851$ & $0.970$ & $0.989$
    \\
    AIIA-RDepth & $0.123$ & $0.080$ & $0.450$ & $ 0.153$ & $0.861$ & $0.975$ & $0.993$
    \\
    GANCV & $0.104$ & $0.060$ & $0.391$ & $0.131$ & $0.898$ & $0.982$ & $0.995$
    \\
    USTCxNetEaseFuxi & $\mathbf{0.088}$ & \underline{$0.046$} & \underline{$0.347$} & $\mathbf{0.115}$ & $\mathbf{0.940}$ & \underline{$0.985$} & \underline{$0.996$}
    \\\midrule
    \textbf{OpenSpaceAI~(Ours)} & \underline{$0.095$} & $\mathbf{0.045}$ & $\mathbf{0.341}$ & \underline{$0.117$} & \underline{$0.928$} & $\mathbf{0.990}$ & $\mathbf{0.998}$
    \\\bottomrule
    \end{tabular}
}
\label{tab:track2_2nd_results}
\end{table*}

\noindent\textbf{Main Results}.
The results of RDDepth are presented in Table~\ref{tab:track2_2nd_results}. Our RDDepth framework ranks second in terms of the a1 metric and outperforms many other participants in other metrics, such as \texttt{Sq Rel}, \texttt{RMSE}, a2, and a3. The results demonstrate the superior robustness of our approach.

\begin{figure}
    \centering
    \includegraphics[width=1.0\linewidth]{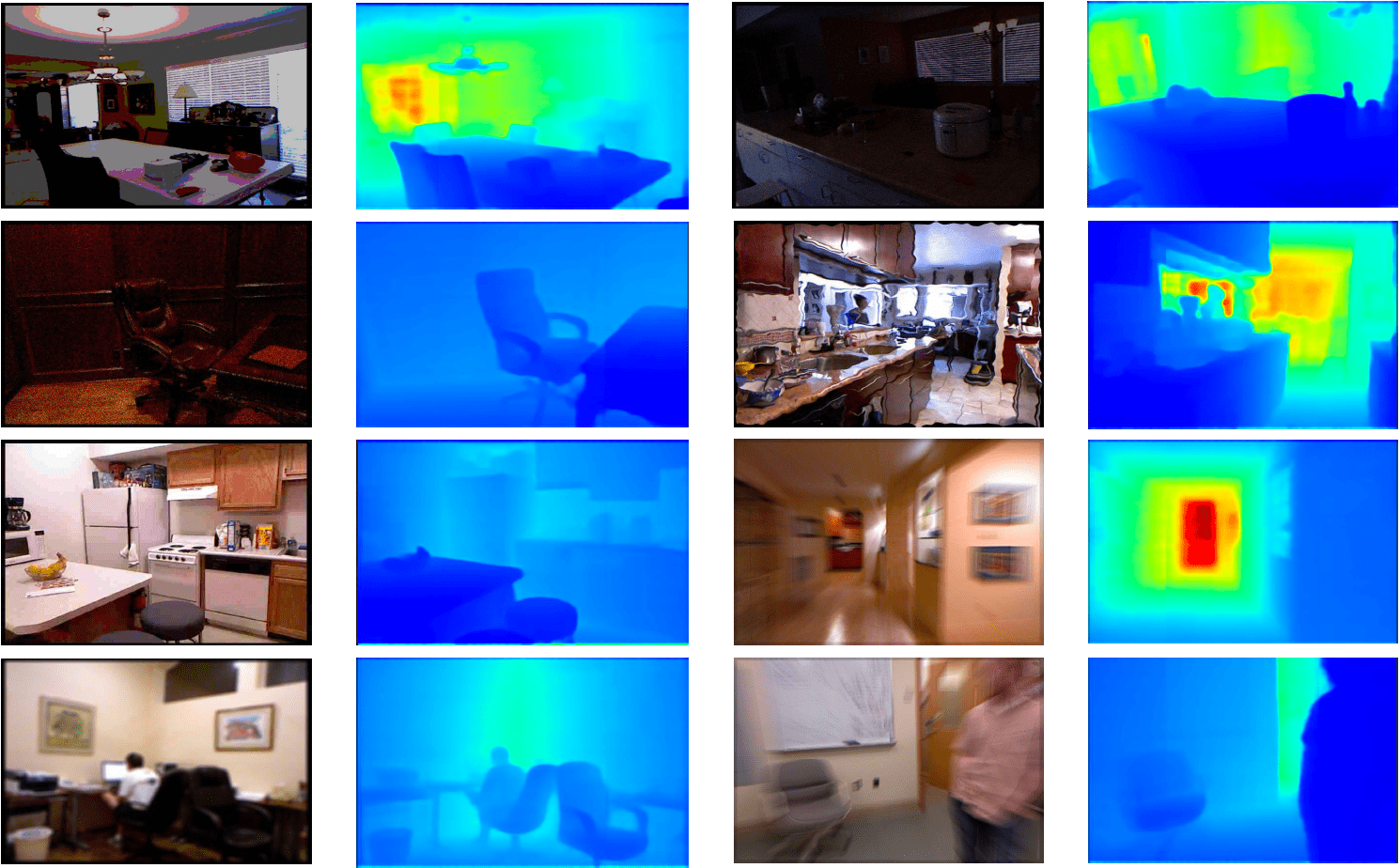}
    \caption{Qualitative results of RDDepth in the second track of the RoboDepth Challenge.}
\label{fig:track2_2nd_qualitative}
\end{figure}

Additionally, we show some visual examples under different corruption scenarios in Figure~\ref{fig:track2_2nd_qualitative}. It can be seen that our proposed RDDepth can predict reasonable and accurate depth maps under various corruptions, such as color quantization, low light, motion blur, zoom blur, \textit{etc}.

\subsubsection{Solution Summary}
In this work, we proposed RDDepth, a Robust Diffusion model for Depth estimation. With the VPD serving as the baseline, we exploited how to leverage the diffusion model’s high-level scene knowledge to guide depth estimation head to counteract the effects of corruptions, thus generalizing well in unseen OoD data. Furthermore, we found that proper data augmentations can benefit the model’s generalization ability. RDDepth achieved state-of-the-art performance on the RoboDepth benchmark and ranked second in the second track of the RoboDepth Challenge.

\subsection{The \textcolor{robo_green}{3rd} Place Solution: \textcolor{robo_green}{GANCV}}
\noindent\textbf{Authors:} \textcolor{gray}{Jiamian Huang and Baojun Li.}

\begin{framed}
    \textbf{Summary} - To better handle depth estimation under real-world corruptions, the \texttt{GANCV} team proposes a joint depth estimation solution that combines AiT with masked image modeling depth estimation (MIM-Depth). New techniques related to data augmentation and model ensemble are incorporated to further improve the depth estimation robustness. By combining the advantages of AiT and MIM-Depth, this solution achieves promising OoD depth prediction results and ranks third in the second track of the RoboDepth Challenge.
\end{framed}

\subsubsection{Overview}
Depth estimation plays a crucial role as one of the vital components in visual systems that capture 3D scene structure. Depth estimation models have been widely deployed in practical applications, such as the 3D reconstruction of e-commerce products, mobile robotics, and autonomous driving \cite{geiger2012kitti,silberman2012nyu2,dong2022survey,laga2020survey}. Compared to expensive and power-hungry LiDAR sensors that provide high-precision but sparse depth information, the unique advantages of low-cost and low-power cameras have made monocular depth estimation techniques a relatively popular choice.

Although promising depth estimation results have been achieved, the current learning-based models are trained and tested on datasets within the same distribution. These approaches often ignore the more commonly occurring OoD situations in the real world. The RoboDepth Challenge was recently established to raise attention among the community for robust depth estimation. To investigate the latest advancements in monocular depth estimation, we propose a solution that combines the AiT \cite{ning2023ait}  and masked image modeling (MIM) depth estimation \cite{xie2023revealing}.

\begin{figure}
    \centering
    \includegraphics[width=1.0\linewidth]{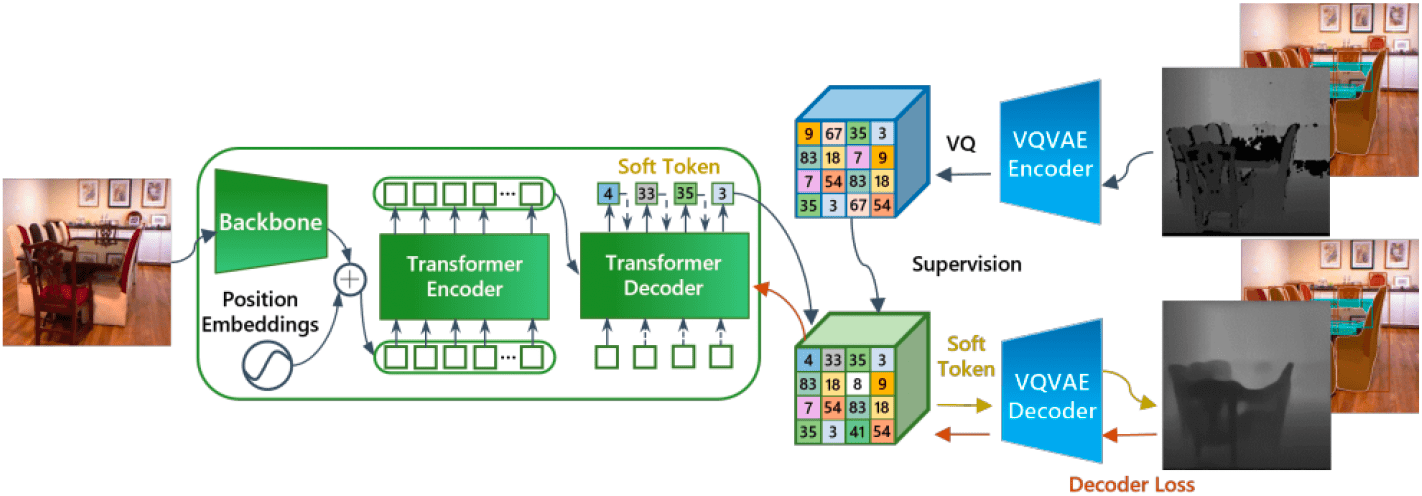}
    \caption{Overview of the architecture proposed in AiT \cite{ning2023ait}. This network structure includes a VQ-VAE \cite{oord2017vqvae} tokenizer and a task solver.}
\label{fig:track2_3rd_framework}
\end{figure}

AiT \cite{ning2023ait} consists of three components: the tokenizer, detokenizer, and
task solver, as shown in Figure~\ref{fig:track2_3rd_framework}. The tokenizer and detokenizer form a VQ-VAE \cite{oord2017vqvae}, which is primarily used for the automatic encoding and decoding of tokens. The task solver is implemented as an auto-regressive encoder-decoder network, where both the encoder and decoder components combine  Transformer blocks to generate
soft tokens. In summary, the task solver model takes images as inputs, predicts token sequences through autoregressive decoding, and employs VQ-VAE’s decoder to transform the predicted tokens
into the desired output results.

MIM is a sub-task of masked signal prediction, where a portion of input images is masked, and deep networks are employed to predict the masked signals conditioned on the visible ones. In this work, we utilized SimMIM \cite{xie2022simmim} model deep estimation training. SimMIM \cite{xie2022simmim} consists of four major components with simple designs: 1) random masking with a large masked patch size; 2) the masked tokens and image tokens are fed to the encoder together; 3) the prediction head is as light as a linear layer; 4) predicting raw pixels of RGB values as the target with the L1 loss of the direct regression. With these simple designs, SimMIM \cite{xie2022simmim} can achieve state-of-the-art performance on different downstream tasks.

In addition to the network architecture, we also explore model ensemble -- a commonly used technique in competitions, aiming to combine the strengths and compensate for the weaknesses of multiple models by integrating their results. For depth estimation, utilizing a model ensemble can effectively balance the estimated results, especially when there are significant differences in the estimated depth values. It can help mitigate the disparities and harmonize the variations among them.

Lastly, we investigate the choice of different backbones in the depth estimation model. The backbone refers to the selection of the underlying architecture or network as the foundational framework for monocular depth estimation. Currently, in the field of computer vision, the most commonly used backbones are Vision Transformers (ViT) \cite{dosovitskiy2020vit} and Swin Transformers \cite{liu2021swin}. The choice between ViT \cite{dosovitskiy2020vit} and Swin Transformers \cite{liu2021swin} depends on various factors. We use Swin Transformers \cite{liu2021swin} as the backbone of our framework due to its general-purpose nature.

\subsubsection{Technical Approach}
We propose a multi-model fusion approach with a primary focus on integrating the results of two depth estimation models: AiT \cite{ning2023ait} and MIM-Depth \cite{xie2023revealing}. Since the overall framework involves combining the results of two models, the training process is conducted in multiple stages. Therefore, we will first discuss the training of the AiT \cite{ning2023ait} algorithm, followed by the training of MIM-Depth \cite{xie2023revealing}. Finally, we will explain how to integrate the results of these two models together. When presenting the training stages of both models, we will also provide details related to training tricks and the relevant training parameters employed. 

\noindent\textbf{AiT Training}.
As per the competition organizers’ requirements, we train AiT \cite{ning2023ait} on the official training split of NYU Depth V2 \cite{silberman2012nyu2}. Moreover, we have strictly limited the utilization of various data augmentation techniques as specified. We initially employed only two data augmentation techniques: random horizontal flipping and random cropping. These methods were used to augment the training data while ensuring adherence to the specified guidelines. In addition to the aforementioned data augmentation techniques, we also employed two additional data augmentation methods: random brightness and random gamma. These augmentation techniques were sourced from the Monocular-Depth-Estimation-Toolbox \cite{lidepthtoolbox2022} and were implemented in compliance with the competition guidelines. We set all the probability of random augmentations to $0.5$.

As can be observed from Figure \ref{fig:track2_3rd_framework}, the training process of AiT \cite{ning2023ait} can be divided into two stages. In the first stage, we focus on training the VQ-VAE \cite{oord2017vqvae} network, which is a token encoder-decoder model. To enhance the robustness of the VQ-VAE \cite{oord2017vqvae}, we apply random masks to the original ground truth depth maps as inputs. The mask is performed with a masking ratio of $0.5$ and a patch size of $16$, allowing the model to better estimate monocular depth information in real-world scenarios. For monocular depth estimation, the input image size adopted is $480\times480$, with a batch size of $512$ in the first training stage. The AdamW \cite{loshchilov2018adamw} optimizer is used with the base learning rate of $3$e-$4$. During the pre-training process, we utilized a server equipped with eight A100 GPUs and trained the VQ-VAE model \cite{oord2017vqvae} for a total of $100$ epochs. In the fine-tuning process, we continued training VQ-VAE \cite{oord2017vqvae} for at least $50$ additional epochs with a learning rate of $1$e-$5$. This fine-tuning process aims to further optimize the model’s performance on the validation set of the NYU Depth V2 dataset \cite{silberman2012nyu2}. Ultimately, the validation loss of the VQ-VAE \cite{oord2017vqvae} model reduced to around $0.021$. For the data augmentation in this stage, we just use random horizontal flipping and random cropping.

In the second stage, we use the Swin Transformer V2 Large \cite{liu2022swin-v2} as the backbone, which is pre-trained with SimMIM \cite{xie2022simmim}. In the training process, we use the AdamW \cite{loshchilov2018adamw} optimizer with a base learning rate of $2$e-$4$; the weight decay is set to $0.075$. Furthermore, we set the layer decay value of the learning rate to $0.9$ in order to prevent the model from overfitting. This value helps to control the learning rate decay rate for different layers of the depth estimation model, ensuring a balanced optimization process during training. We also set the drop path rate to $0.1$. The total training steps are $15150$ with a batch size of $80$. The step learning rate schedule is used and the learning rate dropped to $2$e-$5$ and $2$e-$6$ at the $7575$-th step and the $12120$-th step, respectively. Regarding data augmentation, in addition to the conventional ones used in VQ-VAE \cite{oord2017vqvae}, we also append random brightness with a limit value from $0.75$ to $1.25$ and a random gamma.

\begin{figure}
    \centering
    \includegraphics[width=0.75\linewidth]{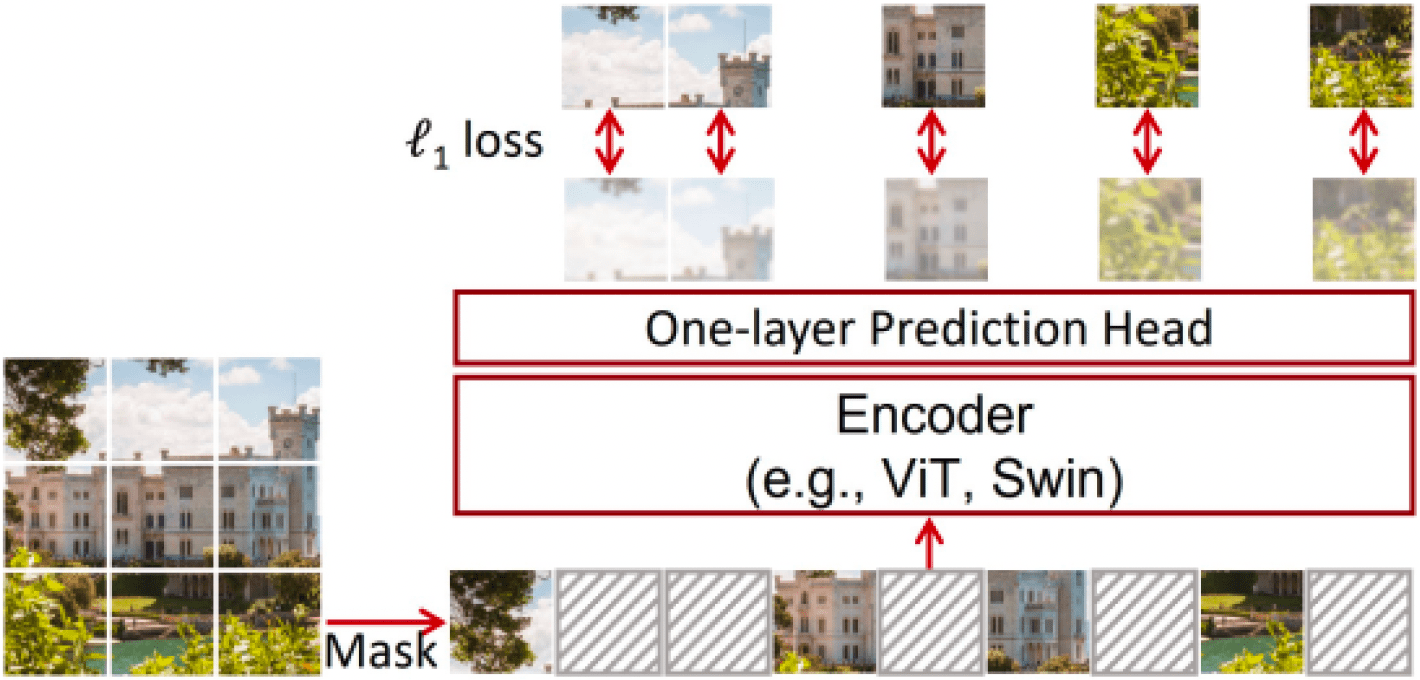}
    \caption{Illustration of MIM-Depth \cite{xie2023revealing}. The overall structure is from SimMIM \cite{xie2022simmim}.}
\label{fig:track2_3rd_mim}
\end{figure}

\noindent\textbf{MIM-Depth-Estimation Training}.
In addition to training AiT \cite{ning2023ait}, we also explored the application of MIM-Depth \cite{xie2023revealing}. Unlike the training strategy used for AiT \cite{ning2023ait}, the training of MIM-Depth \cite{xie2023revealing} is performed in an end-to-end manner, which makes the training process relatively simpler. The network architecture of MIM-Depth \cite{xie2023revealing} is depicted in Figure~\ref{fig:track2_3rd_mim}. As mentioned earlier, we select SimMIM \cite{xie2022simmim} as the backbone architecture.

During the training of MIM-Depth \cite{xie2023revealing}, we apply five data augmentation techniques to enhance the model performance. These methods include random masking, random horizontal flipping, random cropping, random brightness adjustment, and random gamma adjustment. For all the random augmentation, a probability of $0.5$ is employed.

\noindent\textbf{Integrating Both Models}.
After completing the training of both models, we experiment with various ensemble strategies to integrate the results of AiT \cite{ning2023ait} and MIM-Depth \cite{xie2023revealing}, aiming to achieve better performance than each individual model. Two ensemble strategies we used include plain averaging and weighted averaging. After conducting comparative experiments, we decide to opt for weighted averaging of the depth estimation results, with weights assigned to the AiT model’s result and the MIM-Depth-Estimation model’s result as $0.6$ and $0.4$, respectively.

\subsubsection{Experimental Analysis}
\noindent\textbf{Implementation Details}.
For the training of MIM-Depth \cite{xie2023revealing}, we select Swin Transformer V2 Large \cite{liu2022swin-v2} as the backbone architecture. Additionally, we apply a trained weight of Swin Transformer V2 Large \cite{liu2022swin-v2} pre-trained on the ImageNet classification dataset as the pre-trained model for MIM-Depth \cite{xie2023revealing}. For monocular depth estimation training, we maintain the same input image size as AiT \cite{ning2023ait}, which consists of $480\times480$ pixels. This consistency in input image size ensures compatibility and facilitates the comparison and integration of results between AiT \cite{ning2023ait} and MIM-Depth \cite{xie2023revealing}. We also apply layer decay during the training, but unlike AiT \cite{ning2023ait}, we set the value to $0.85$. For the drop path rate setting, we apply it with a value of $0.5$. Regarding the data augmentation for masking, we select the mask patch size of $32$ and the mask ratio of $0.1$. In terms of the optimizer, we use AdamW \cite{loshchilov2018adamw} with a learning rate of $5$e-$4$. We use the linear learning rate schedule and set a minimum learning rate to prevent the learning rate from decreasing too quickly. We train the entire model for approximately $25$ epochs on an $8$ V100 GPUs server. The batch size we set during training is $24$.

\begin{table*}[t]
\caption{Quantitative results of the candidate models \cite{ning2023ait,xie2023revealing} with different data augmentation strategies on the RoboDepth competition leaderboard (Track \# 2). \texttt{Aug1} indicates that only random horizontal flipping and random cropping are used during the training. \texttt{Aug2} refers to the addition of random brightness and random gamma as data augmentations on top of \texttt{Aug1}. The \textbf{best} and \underline{second best} scores of each metric are highlighted in \textbf{bold} and \underline{underline}, respectively.}
\centering\scalebox{0.78}{
    \begin{tabular}{l|cccc|ccc}
    \toprule
    \textbf{Method} & \cellcolor{blue!10}\textbf{Abs Rel~$\downarrow$} & \cellcolor{blue!10}\textbf{Sq Rel~$\downarrow$} & \cellcolor{blue!10}\textbf{RMSE~$\downarrow$} & \cellcolor{blue!10}\textbf{log RMSE~$\downarrow$} & \cellcolor{red!10}$\delta<1.25$~$\uparrow$ & \cellcolor{red!10}$\delta<1.25^2$~$\uparrow$ & \cellcolor{red!10}$\delta<1.25^3$~$\uparrow$
    \\\midrule\midrule
    MIM-Depth \cite{xie2023revealing} \textit{w/} \texttt{Aug1} & $0.132$ & $0.091$ & $0.458$ & $0.157$ & $0.849$ & $0.967$ & \underline{$0.990$}
    \\
    MIM-Depth \cite{xie2023revealing} \textit{w/} \texttt{Aug2} & \underline{$0.115$} & \underline{$0.070$} & \underline{$0.414$} & \underline{$0.141$} & \underline{$0.883$} & \underline{$0.976$} & $\mathbf{0.994}$
    \\\midrule
    AiT \cite{ning2023ait} \textit{w/} \texttt{Aug1} & \underline{$0.115$} & $0.076$ & $0.435$ & $0.146$ & $0.871$ & $0.973$ & \underline{$0.990$}
    \\
    AiT \cite{ning2023ait} \textit{w/} \texttt{Aug2} & $\mathbf{0.104}$ & $\mathbf{0.062}$ & $\mathbf{0.405}$ & $\mathbf{0.134}$ & $\mathbf{0.891}$ & $\mathbf{0.981}$ & $\mathbf{0.994}$
    \\\bottomrule
\end{tabular}
}
\label{tab:track2_3rd_aug}
\end{table*}

\begin{table*}[t]
\caption{Quantitative results of MIM-Depth \cite{xie2023revealing} with different masking ratios and patch sizes on the RoboDepth competition leaderboard (Track \# 2). \texttt{p} indicates patch size and \texttt{r} denotes masking ratio. The \textbf{best} and \underline{second best} scores of each metric are highlighted in \textbf{bold} and \underline{underline}, respectively.}
\centering\scalebox{0.78}{
    \begin{tabular}{l|cccc|ccc}
    \toprule
    \textbf{Method} & \cellcolor{blue!10}\textbf{Abs Rel~$\downarrow$} & \cellcolor{blue!10}\textbf{Sq Rel~$\downarrow$} & \cellcolor{blue!10}\textbf{RMSE~$\downarrow$} & \cellcolor{blue!10}\textbf{log RMSE~$\downarrow$} & \cellcolor{red!10}$\delta<1.25$~$\uparrow$ & \cellcolor{red!10}$\delta<1.25^2$~$\uparrow$ & \cellcolor{red!10}$\delta<1.25^3$~$\uparrow$
    \\\midrule\midrule
    MIM-Depth \textit{w/} \texttt{p16-r0.1} & $\mathbf{0.115}$ & $\mathbf{0.070}$ & \underline{$0.418$} & $\mathbf{0.141}$ & \underline{$0.881$} & \underline{$0.971$} & \underline{$0.990$}
    \\
    MIM-Depth \textit{w/} \texttt{p32-r0.0} & \underline{$0.169$} & \underline{$0.157$} & $0.535$ & \underline{$0.186$} & $0.794$ & $0.940$ & $0.978$
    \\
    MIM-Depth \textit{w/} \texttt{p32-r0.1} & $\mathbf{0.115}$ & $\mathbf{0.070}$ & $\mathbf{0.414}$ & $\mathbf{0.141}$ & $\mathbf{0.883}$ & $\mathbf{0.976}$ & $\mathbf{0.994}$
    \\\bottomrule
\end{tabular}
}
\label{tab:track2_3rd_mask}
\end{table*}

\begin{table*}[t]
\caption{Quantitative results of MIM-Depth \cite{xie2023revealing}, AiT \cite{ning2023ait}, our ensemble model, and other participants on the RoboDepth competition leaderboard (Track \# 2). The \textbf{best} and \underline{second best} scores of each metric are highlighted in \textbf{bold} and \underline{underline}, respectively.}
\centering\scalebox{0.78}{
    \begin{tabular}{l|cccc|ccc}
    \toprule
    \textbf{Method} & \cellcolor{blue!10}\textbf{Abs Rel~$\downarrow$} & \cellcolor{blue!10}\textbf{Sq Rel~$\downarrow$} & \cellcolor{blue!10}\textbf{RMSE~$\downarrow$} & \cellcolor{blue!10}\textbf{log RMSE~$\downarrow$} & \cellcolor{red!10}$\delta<1.25$~$\uparrow$ & \cellcolor{red!10}$\delta<1.25^2$~$\uparrow$ & \cellcolor{red!10}$\delta<1.25^3$~$\uparrow$
    \\\midrule\midrule
    USTCxNetEaseFuxi & $\mathbf{0.088}$ & \underline{$0.046$} & \underline{$0.347$} & $\mathbf{0.115}$ & $\mathbf{0.940}$ & \underline{$0.985$} & \underline{$0.996$}
    \\
    OpenSpaceAI & \underline{$0.095$} & $\mathbf{0.045}$ & $\mathbf{0.341}$ & \underline{$0.117$} & \underline{$0.928$} & $\mathbf{0.990}$ & $\mathbf{0.998}$
    \\
    AIIA-RDepth & $0.123$ & $0.088$ & $0.480$ & $0.162$ & $0.861$ & $0.975$ & $0.993$
    \\\midrule
    \textbf{MIM-Depth~(Ours)} & $0.115$ & $0.070$ & $0.414$ & $0.141$ & $0.883$ & $0.976$ & $0.994$
    \\
    \textbf{AiT~(Ours)} & $0.104$ & $0.062$ & $0.405$ & $0.134$ & $0.891$ & $0.981$ & $0.994$
    \\
    \textbf{Ensemble~(Ours)} & $0.104$ & $0.060$ & $0.391$ & $0.131$ & $0.898$ & $0.982$ & $0.995$
    \\\bottomrule
\end{tabular}
}
\label{tab:track2_3rd_leaderboard}
\end{table*}

We conduct several comparative experiments focusing on the selection of data augmentation methods, masking strategies, and ensemble strategies. All experimental results are obtained using the test set of the second track of the RoboDepth competition.

\noindent\textbf{Data Augmentations}.
Regarding the use of data augmentations, we compare multiple combinations and present the results in Table~\ref{tab:track2_3rd_aug}. We first establish a data augmentation combination that includes random horizontal flipping and random cropping, both with a probability of $0.5$, dubbed \texttt{Aug1}. We apply this combination to preprocess the training data for both MIM-Depth \cite{xie2023revealing} and AiT \cite{ning2023ait}. We also form another data augmentation combination by adding random brightness variation and random gamma adjustment to the previous combination; we denote this strategy as \texttt{Aug2}. Both of these augmentations are applied with a probability value of $0.5$.

\noindent\textbf{Masking Strategy}.
We conduct experiments with two sets of masking strategies on the MIM-Depth\cite{xie2023revealing}. In the first set, we select the patch size to $32$, while in the second set, the patch size is $16$. Both sets have a mask ratio of $0.1$. As shown in Table~\ref{tab:track2_3rd_mask}, we observe that the two different mask patch sizes have a minimal impact on the final depth estimation results. Additionally, we compare the scenarios where no masking ratio is set (baseline). It can be seen that masking-based modeling has a significant impact on the model's robustness.

\noindent\textbf{Ensemble Strategy}.
In terms of ensemble strategies, we compare the plain averaging and weighted averaging methods. As shown in Table~\ref{tab:track2_3rd_leaderboard}, we can see that the weighted averaging method outperforms the simple averaging method. We also observe that the optimal weights for the ensemble are $0.6$ for AiT \cite{ning2023ait} and $0.4$ for MIM-Depth \cite{xie2023revealing}. Furthermore, the ensemble approach achieves better performance compared to individual models.

\subsubsection{Solution Summary}
In this work, we presented a collaborative model ensemble solution for robust monocular depth estimation and conducted various experimental analyses to demonstrate its effectiveness. The proposed solution primarily consists of combining AiT and MIM-Depth, with an in-depth discussion on the use of data augmentation and model ensemble techniques. This solution is employed in the second track of the RoboDepth Challenge. Ultimately, we achieved a third-place ranking in the second track of this competition.

\subsection{The Innovation Prize Solution: AIIA-RDepth}
\noindent\textbf{Authors:} \textcolor{gray}{Sun Ao, Gang Wu, Zhenyu Li, Xianming Liu, and Junjun Jiang.}

\begin{framed}
    \textbf{Summary} - To enhance the resilience of deep depth estimation models, the \texttt{AIIA-RDepth} team introduces a multi-stage methodology that incorporates both spatial and frequency domain operations. Initially, several masks are employed to selectively occlude regions in the input image, followed by spatial domain enhancement techniques. Subsequently, robust attacks are applied to the high-frequency information of the image in the frequency domain. Finally, these two approaches are amalgamated into a unified framework called MRSF: Masking and Recombination in the Spatial and Frequency domains.
\end{framed}

\subsubsection{Overview}
Monocular depth estimation is a vital research area in the field of computer vision and finds wide-ranging applications in industries such as robotics \cite{dong2022survey}, autonomous driving \cite{geiger2012kitti,uhrig2017kitti}, virtual reality, and 3D reconstruction \cite{silberman2012nyu2}. Recently, deep learning has witnessed significant advancements and has gradually become the mainstream approach for addressing monocular depth estimation problems.

Existing models for supervised monocular depth estimation often train and test on datasets within the same distribution, yielding satisfactory performance on the corresponding testing sets. However, when there exists an incomplete distribution match or certain corruptions between the training and testing data, such as variations in weather and lighting conditions, sensor failures and movements, and data processing issues, the performance of these deep learning models tends to be significantly degraded. To address these challenges, the RoboDepth competition introduced novel datasets that include $18$ types of corruptions, aiming to probe the robustness of models against these corruptions. In light of these OoD issues, we propose a robust data augmentation method that enhances images in both spatial and frequency domains.

Among approaches for supervised monocular depth estimation, DepthFormer from Li \textit{et al.} \cite{li2022depthformer} stands out as a significant contribution. This model proposed to leverage the Transformer architecture to effectively capture the global context by integrating an attention mechanism. Furthermore, it employs an additional convolutional branch to preserve local information.

As for data augmentation, the CutOut technique from DeVries and Taylor \cite{devries2017cutout} is widely acknowledged, where square regions of the input are randomly masked out during training. This approach has been proven effective in improving the robustness and overall performance of convolutional neural networks. Additionally, in frequency domain enhancement, Amplitude Phase Reconstruction (APR) from Chen \textit{et al.} \cite{chen2021apr} is an important method. It directs the attention of CNN models toward the phase spectrum, enhancing their ability to extract meaningful information from the frequency domain.

\subsubsection{Technical Approach}
\begin{figure}
    \centering
    \includegraphics[width=0.9\linewidth]{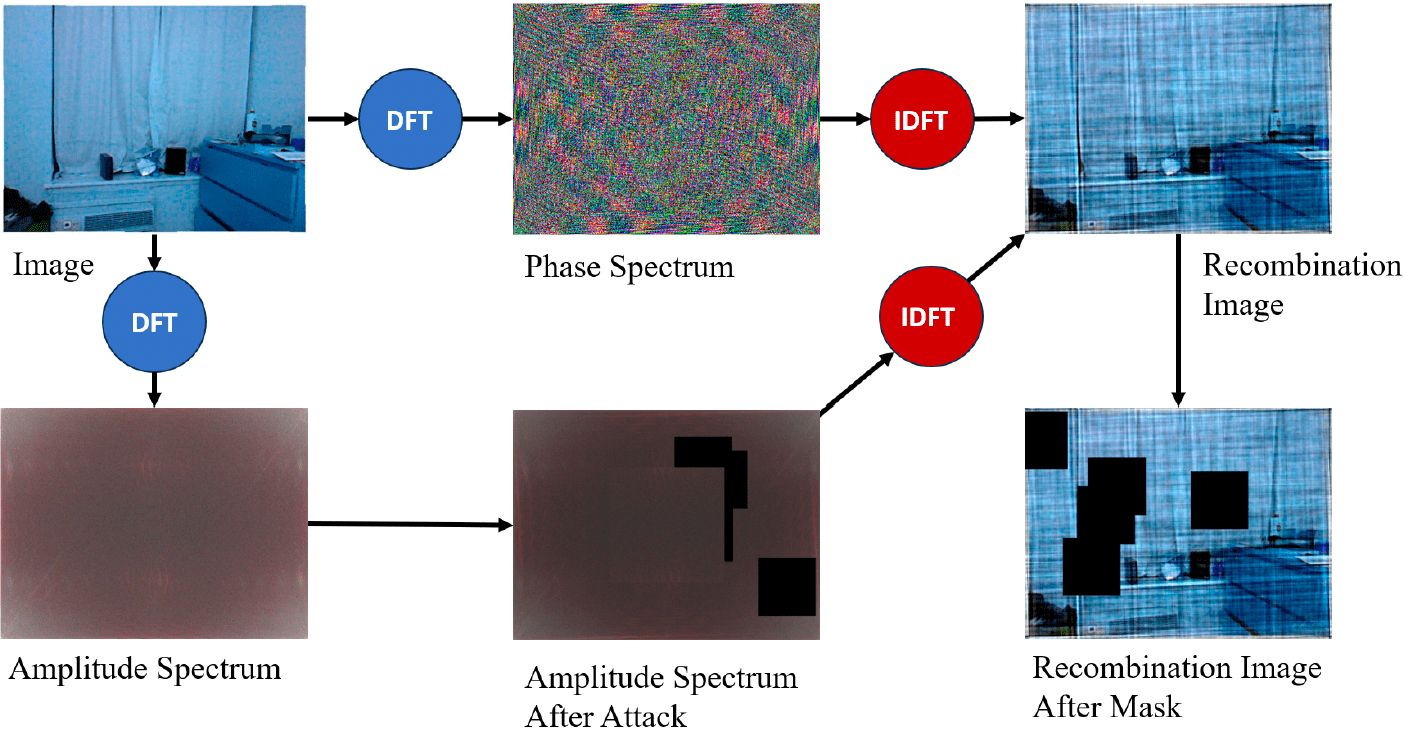}
    \caption{The overall pipeline of the Masking and Recombination in the Spatial and Frequency domains (MRSF) approach for robust monocular depth estimation.}
\label{fig:track2_innov_framework}
\end{figure}

In this section, we will elucidate the motivation behind and introduce the two components of our Masking and Recombination in the Spatial and Frequency domains (MRSF) approach: 1) masking image regions in the spatial domain and 2) reconstructing the image in the frequency domain. Figure~\ref{fig:track2_innov_framework} provides an overview of MRSF.

\noindent\textbf{Motivation}.
While considerable progress has been made in learning-based depth estimation models, their training and testing processes often rely on clean datasets, disregarding the OoD situations. In practical scenarios, common corruptions are more likely to occur, which can have safety-critical implications for applications such as autonomous driving and robot navigation.

Upon thorough examination of various corrupted images in the RoboDepth Challenge, we have observed that the corruption effects introduced in the competition are inclined to contain high-frequency interference, consequently resulting in substantial alterations to the local information.

To rectify the perturbations to local information and address the issue of high-frequency interference, we employ a robust data augmentation technique, MRSF, that encompasses both spatial and frequency domain operations. This approach enabled us to capture the global information of the image while simultaneously addressing the high-frequency disturbances introduced by the attacking images.

\noindent\textbf{SDA: Spatial Domain Augmentation}.
To facilitate the model’s enhanced understanding of global image information and improve its robustness, we employ a masking method to augment the images in the spatial domain. Initially, we randomly select $N$ points within the $640\times480$ images of NYU Depth V2 \cite{silberman2012nyu2}. These points serve as the top-left corners for generating patch masks, each of a fixed size (specifically, we use a square of size $a\times a$ in the practical implementation, where $a$ is the mask length). If a mask extended beyond the boundaries of the original image, the exceeding portion will be discarded to ensure that only parts within the image are retained.

\begin{figure}[t]
    \centering
    \includegraphics[width=1\linewidth]{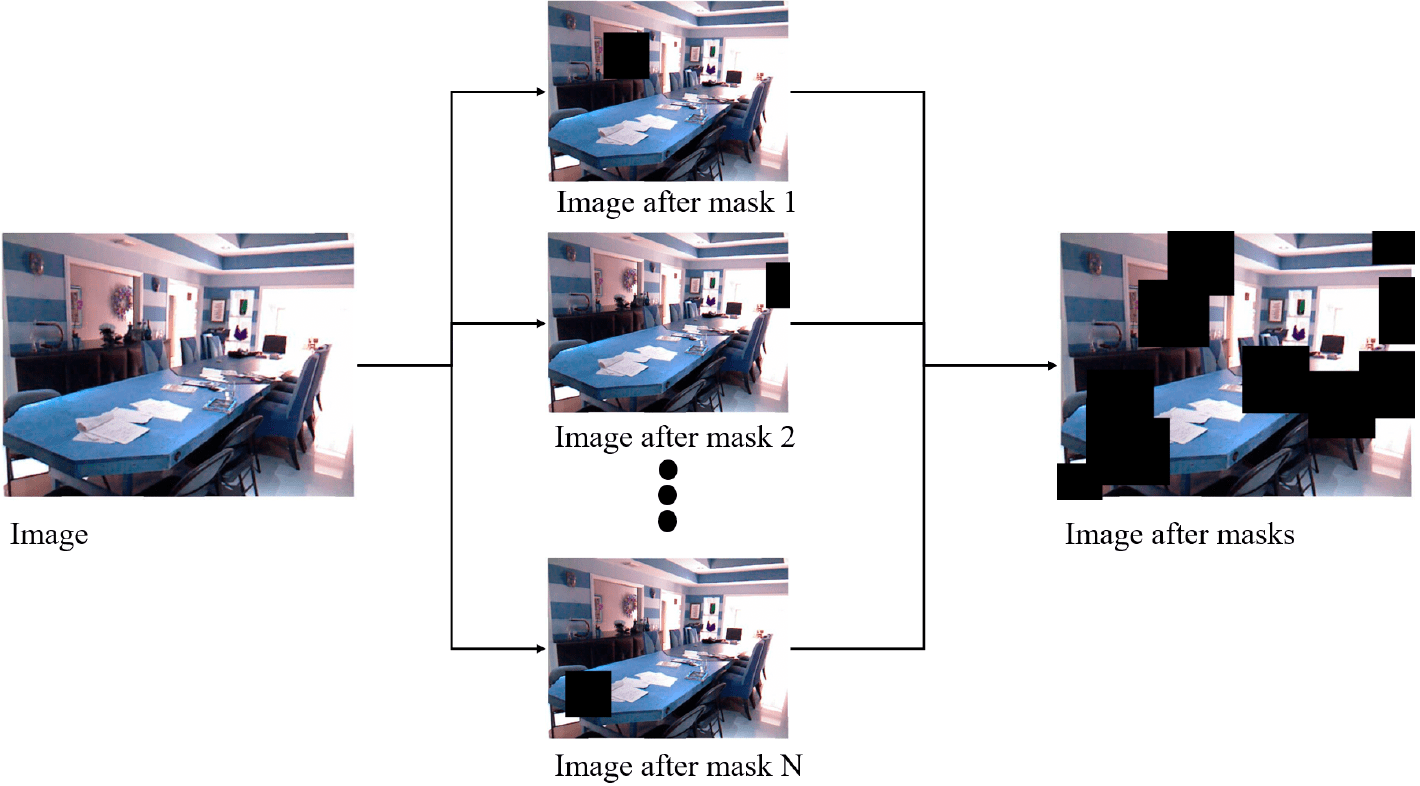}
    \caption{The overall pipeline of the Spatial Domain Augmentation (SDA) operation.}
\label{fig:track2_innov_sda}
\end{figure}

\begin{figure}[t]
    \centering
    \includegraphics[width=0.99\linewidth]{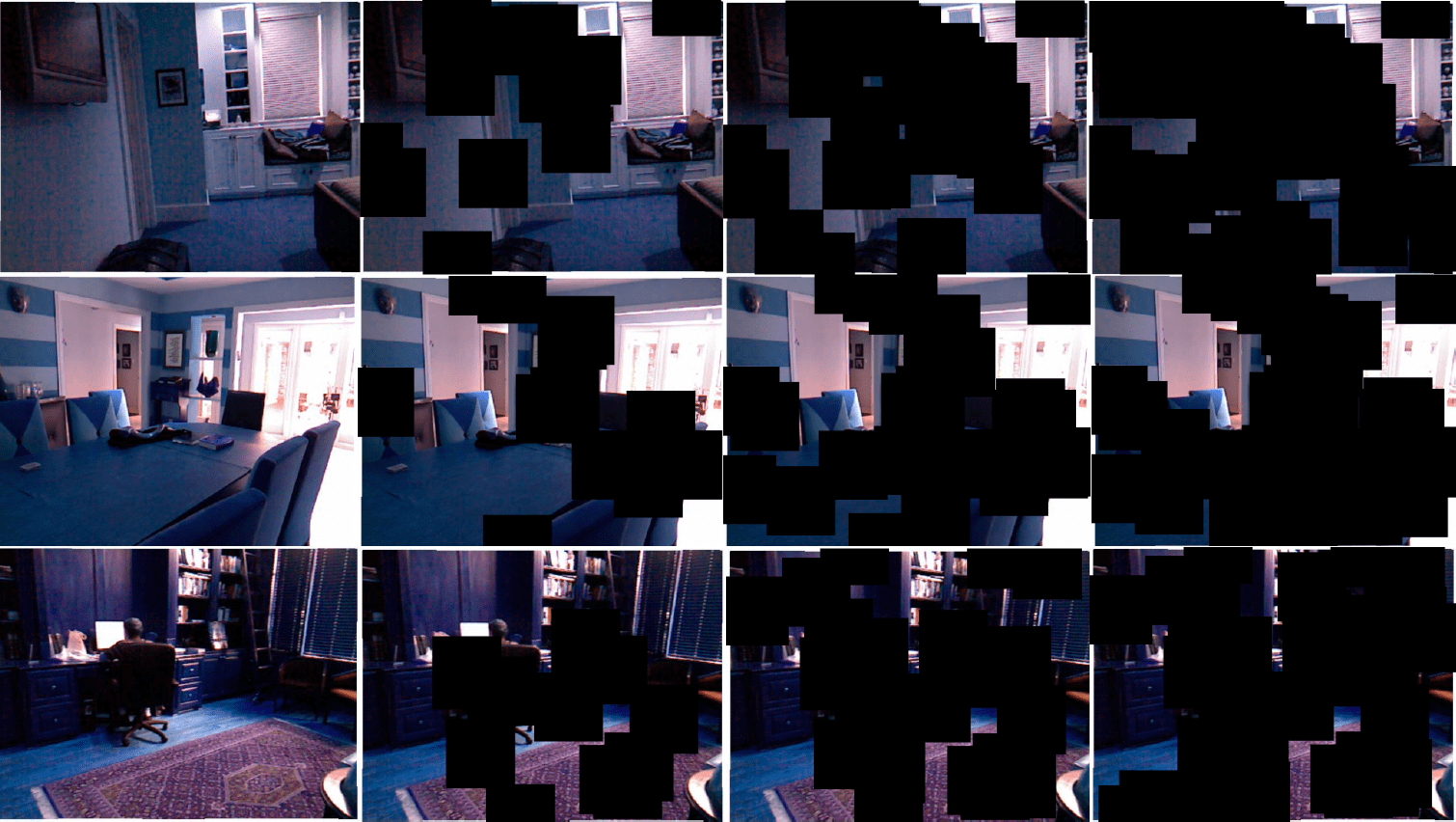}
    \caption{Illustrative example of images after applying the SDA method. From left to right: the $N$ values in SDA are set to $0$ (original image), $3$, $6$, and $12$, respectively, while $a$ remains fixed at $120$.}
\label{fig:track2_innov_sda_effect}
\end{figure}

The overall pipeline of SDA is shown in Figure~\ref{fig:track2_innov_sda}. Through this process, we generate $N$ mask images based on the $N$ points. Subsequently, we perform a logical \texttt{OR} operation on these $N$ mask images, merging them into a final image mask. By applying this mask to the original image, we obtain the augmented image in the spatial domain. The SDA method relies on two critical hyperparameters, namely $N$ and $a$, which have a substantial impact on its performance. Figure~\ref{fig:track2_innov_sda_effect} provides examples of setting $N$ to different values. To determine their optimal values, we conducted an extensive series of experiments. Through careful analysis, we discover that setting $N$ to $12$ and $a$ to $120$ resulted in the model achieving its peak performance. This finding highlights the importance of precisely tuning these hyperparameters in order to maximize the effectiveness of the SDA method.

\noindent\textbf{FDA: Frequency Domain Augmentation}.
We apply a rotation of angle $\theta$ to the original input image, resulting in a new image. Subsequently, we perform a two-dimensional Fourier transform to convert both images into the frequency domain, yielding two frequency domain representations. While preserving the phase spectrum of the frequency domain images, we extract the magnitude values from each frequency domain representation.

\begin{figure}[t]
    \centering
    \includegraphics[width=1\linewidth]{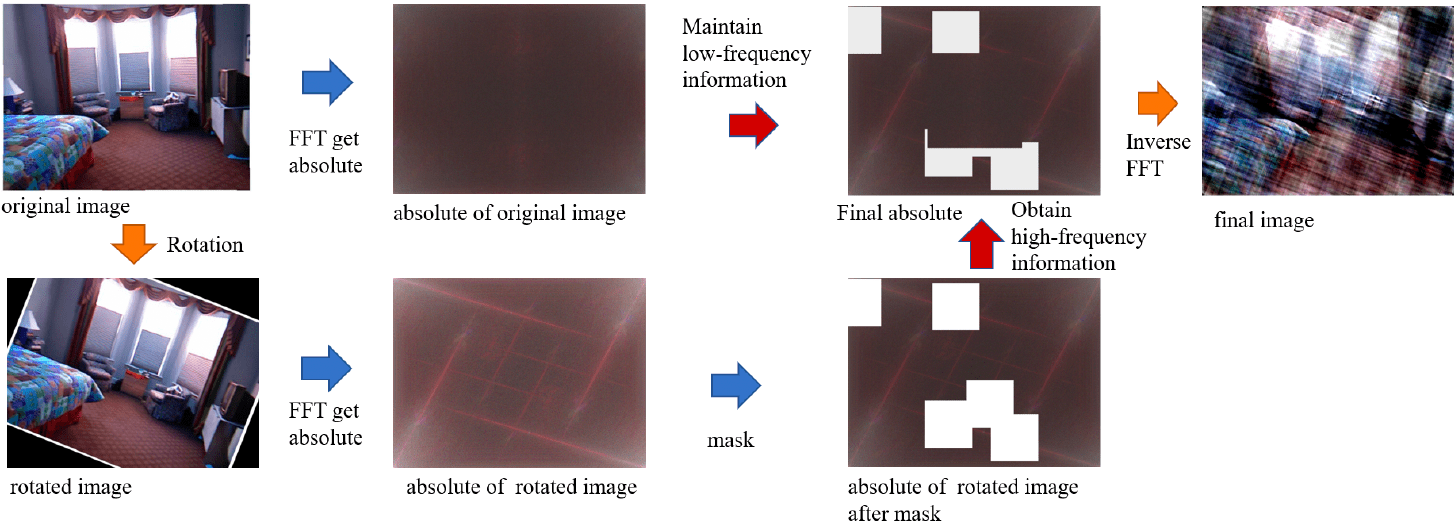}
    \caption{The overall pipeline of the Frequency Domain Augmentation (FDA) operation.}
\label{fig:track2_innov_fda}
\end{figure}

\begin{figure}[t]
    \centering
    \includegraphics[width=0.99\linewidth]{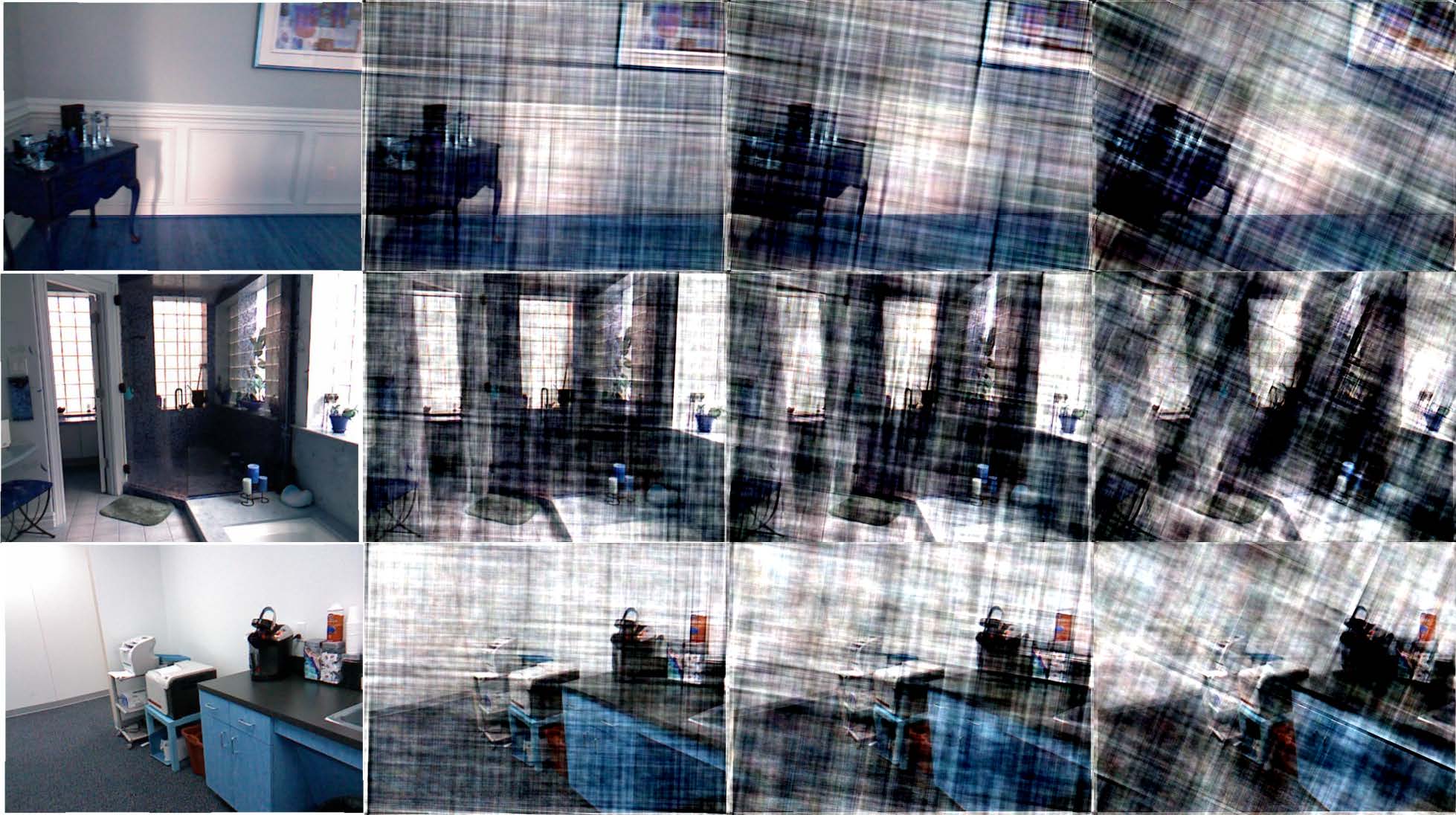}
    \caption{Illustrative examples of images after applying the FDA method. From left to right: the $\theta$ values in FDA are set to $0$ (original image), $3$, $6$, and $12$, respectively.}
\label{fig:track2_innov_fda_effect}
\end{figure}

\begin{figure}[t]
    \centering
    \includegraphics[width=0.99\linewidth]{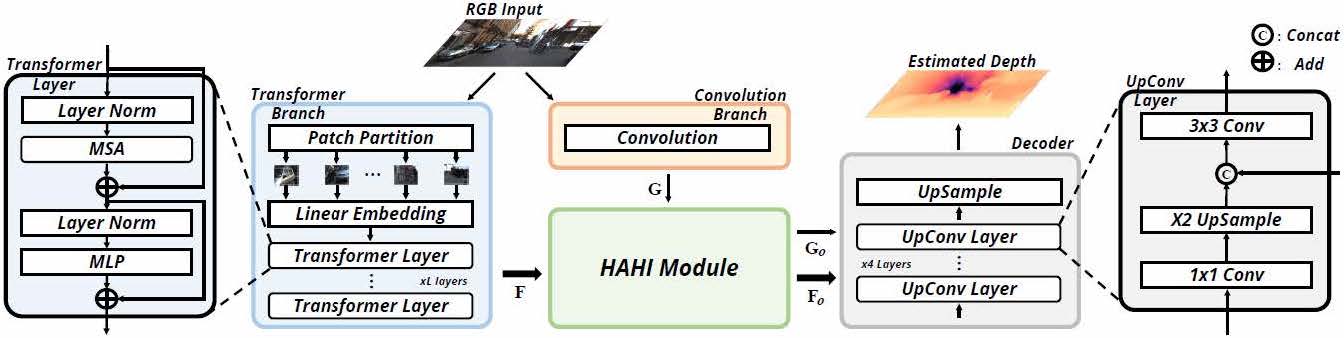}
    \caption{The framework overview of DepthFormer \cite{li2022depthformer}.}
\label{fig:track2_innov_depthformer}
\end{figure}

\begin{table*}[t]
\caption{Quantitative results on the RoboDepth competition leaderboard (Track \# 2). The \textbf{best} and \underline{second best} scores of each metric are highlighted in \textbf{bold} and \underline{underline}, respectively.}
\centering\scalebox{0.78}{
    \begin{tabular}{l|cccc|ccc}
    \toprule
    \textbf{Method} & \cellcolor{blue!10}\textbf{Abs Rel~$\downarrow$} & \cellcolor{blue!10}\textbf{Sq Rel~$\downarrow$} & \cellcolor{blue!10}\textbf{RMSE~$\downarrow$} & \cellcolor{blue!10}\textbf{log RMSE~$\downarrow$} & \cellcolor{red!10}$\delta<1.25$~$\uparrow$ & \cellcolor{red!10}$\delta<1.25^2$~$\uparrow$ & \cellcolor{red!10}$\delta<1.25^3$~$\uparrow$
    \\\midrule\midrule
    USTCxNetEaseFuxi & $\mathbf{0.088}$ & \underline{$0.046$} & \underline{$0.347$} & $\mathbf{0.115}$ & $\mathbf{0.940}$ & \underline{$0.985$} & \underline{$0.996$}
    \\
    OpenSpaceAI & \underline{$0.095$} & $\mathbf{0.045}$ & $\mathbf{0.341}$ & \underline{$0.117$} & \underline{$0.928$} & $\mathbf{0.990}$ & $\mathbf{0.998}$
    \\
    GANCV & $0.104$ & $0.060$ & $0.391$ & $0.131$ & $0.898$ & $0.982$ & $0.995$
    \\\midrule
    \textbf{AIIA-RDepth~(Ours)} & $0.123$ & $0.088$ & $0.480$ & $0.162$ & $0.861$ & $0.975$ & $0.993$
    \\\bottomrule
\end{tabular}
}
\label{tab:track2_innov_leaderboard}
\end{table*}

\begin{table*}[t]
\caption{Quantitative results of different components in the proposed MRSF framework. The \textbf{best} and \underline{second best} scores of each metric are highlighted in \textbf{bold} and \underline{underline}, respectively.}
\centering\scalebox{0.78}{
    \begin{tabular}{l|cccc|ccc}
    \toprule
    \textbf{Method} & \cellcolor{blue!10}\textbf{Abs Rel~$\downarrow$} & \cellcolor{blue!10}\textbf{Sq Rel~$\downarrow$} & \cellcolor{blue!10}\textbf{RMSE~$\downarrow$} & \cellcolor{blue!10}\textbf{log RMSE~$\downarrow$} & \cellcolor{red!10}$\delta<1.25$~$\uparrow$ & \cellcolor{red!10}$\delta<1.25^2$~$\uparrow$ & \cellcolor{red!10}$\delta<1.25^3$~$\uparrow$
    \\\midrule\midrule
    DepthFormer \cite{li2022depthformer} & $0.131$ & $0.095$ & $0.507$ & $0.170$ & $0.827$ & $0.963$ & $0.987$
    \\\midrule
    + SDA & \underline{$0.127$} & $\mathbf{0.088}$ & $\mathbf{0.480}$ & $\mathbf{0.162}$ & \underline{$0.850$} & $0.967$ & $0.988$
    \\
    + FDA & $0.128$ & \underline{$0.087$} & \underline{$0.462$} & \underline{$0.160$} & \underline{$0.850$} & \underline{$0.969$} & \underline{$0.989$}
    \\\midrule
    + \textbf{MRSF} & $\mathbf{0.123}$ & $\mathbf{0.088}$ & $\mathbf{0.480}$ & $\mathbf{0.162}$ & $\mathbf{0.861}$ & $\mathbf{0.975}$ & $\mathbf{0.993}$
    \\\bottomrule
\end{tabular}
}
\label{tab:track2_innov_ablation}
\end{table*}

The overall pipeline of FDA is shown in Figure~\ref{fig:track2_innov_fda}. The first image tends to retain its low-frequency components, with the low-frequency signal defined as the area with a size of $S$ around the center of the frequency domain image; while the remaining portion represented the high-frequency signal. The second image tends to exclusively preserve its high-frequency components. We then reconstruct the low-frequency and high-frequency components of the two frequency domain representations, resulting in a single reconstructed frequency domain image. Subsequently, we apply a mask to the high-frequency portion of this frequency domain image to enhance the model’s robustness to high-frequency information. We obtain the final image by performing an inverse Fourier transform.

In this method, two crucial hyperparameters, \textit{i.e.} $\theta$ and $S$, play significant roles in the success of the FDA. Figure~\ref{fig:track2_innov_fda_effect} provides examples of setting $\theta$ to different values. We conducted extensive experiments on these two parameters and found that the model tends to achieve the optimal performance when $\theta$ is set to $24$ degrees and $S$ is set to $50\times50$.

\noindent\textbf{MRSF: Masking \& Recombination in Spatial \& Frequency Domains}.
After incorporating the SDA and FDA methods, we observe significant performance improvements in the OoD testing set. Therefore, combining these two methods became a natural idea. Our approach involves concatenating the two methods and assigning a certain probability for their usage, denoted as $\rho1$ and $\rho2$, respectively, during a joint data augmentation. Regarding the issue of the order in which the methods are applied, we conducted experiments and found that when SDA is applied first, the masks have already disrupted the model’s frequency domain properties. Consequently, conducting an attack in the frequency domain after the SDA stage results in images with significant discrepancies in both frequency and spatial domains compared to the original image, rendering the data augmentation ineffective. To address this, we adopted the strategy of performing FDA first, followed by a spatial domain enhancement, to achieve our combined data augmentation.

In our MRSF approach, the two mentioned hyperparameters, $\rho1$, and $\rho2$, play a significant role in balancing the augmentation effects brought by SDA and FDA. We conducted extensive experiments on these two parameters and determined that the optimal performance tends to be achieved when $\rho1$ and $\rho2$ are both set to $0.5$.

\subsubsection{Experimental Analysis}
\noindent\textbf{Implementation Details}.
The MRSF framework is built upon DepthFormer \cite{li2022depthformer} -- a state-of-the-art model in the field of monocular depth estimation which incorporates a Swin Transformer backbone to capture the global context of the input image through an effective attention mechanism. It also utilizes a CNN to extract local information and employs a hierarchical aggregation and heterogeneous interaction module to fuse the features obtained from both components. Figure~\ref{fig:track2_innov_depthformer} provides an overview of DepthFormer \cite{li2022depthformer}. We resort to the Monocular-Depth-Estimation-Toolbox \cite{lidepthtoolbox2022} for the implementation of our baseline. The model is trained on the official training split of the NYU Depth V2 dataset \cite{silberman2012nyu2}, which contains $24000$ RGB-depth pairs with a spatial resolution of $640\times480$. 

\noindent\textbf{Comparative Study}.
We summarize the competition results in Table~\ref{tab:track2_innov_leaderboard}. Our approach achieved the fourth position in the second track of the RoboDepth competition and was honored with the innovative prize. Subsequently, we proceed to study the effects brought by SDA, FDA, and MRSF. The results are shown in Table~\ref{tab:track2_innov_ablation}. Specifically, for MRSF, we employed a stochastic approach where we randomly applied the FDA method to attack the model in the frequency domain, particularly targeting the high-frequency components, with a certain probability. Within the already perturbed images, we further applied the SDA model in the spatial domain using a random mask, again with a certain probability. Through this fusion approach, our method exhibited performance improvements beyond those achieved by either individual method alone.

\noindent\textbf{Ablation Study}.
After completing the baseline testing, we proceed to ablate the effects brought by SDA and FDA. When applying masks in the spatial domain, the number of masks ($N$) and the size of individual masks ($a$) are two critical hyperparameters to determine. We conducted numerous experiments regarding these two parameters and show the results in Table~\ref{tab:track2_innov_sda}.

Initially, we made a conjecture that the model’s performance is correlated with the total area of the masks; when the total area remains constant, the impact of $N$ and $a$ on the model’s performance would be limited. This conjecture was validated in the first three experimental groups. Subsequently, while keeping the size of individual masks ($a$) fixed, we varied the number of masks ($N$) and found that the model achieved optimal performance when $N$ was set to $12$ and $a$ was set to $120$. When the mask size is too large or too small, the model’s performance does not reach its optimal level. Our experiments have demonstrated that the model tends to achieve the best possible performance when the total area of the masks is approximately $75\%$ of the original input resolution.

\begin{table*}[t]
\caption{Ablation results of Spatial Domain Augmentation (SDA) with different hyperparameters on the testing set of the second track of the RoboDepth competition. The \textbf{best} and \underline{second best} scores of each metric are highlighted in \textbf{bold} and \underline{underline}, respectively.}
\centering\scalebox{0.78}{
    \begin{tabular}{l|cccc|ccc}
    \toprule
    \textbf{Method} & \cellcolor{blue!10}\textbf{Abs Rel~$\downarrow$} & \cellcolor{blue!10}\textbf{Sq Rel~$\downarrow$} & \cellcolor{blue!10}\textbf{RMSE~$\downarrow$} & \cellcolor{blue!10}\textbf{log RMSE~$\downarrow$} & \cellcolor{red!10}$\delta<1.25$~$\uparrow$ & \cellcolor{red!10}$\delta<1.25^2$~$\uparrow$ & \cellcolor{red!10}$\delta<1.25^3$~$\uparrow$
    \\\midrule\midrule
    DepthFormer \cite{li2022depthformer} & $0.131$ & $0.095$ & $0.507$ & $0.170$ & $0.827$ & $0.963$ & \underline{$0.987$}
    \\\midrule
    + SDA ($N=12, a=60$) & \underline{$0.128$} & $0.091$ & $0.491$ & $0.166$ & $0.839$ & $0.964$ & \underline{$0.987$}
    \\
    + SDA ($N=48, a=30$) & $\mathbf{0.127}$ & $0.091$ & $0.491$ & $0.165$ & $0.840$ & \underline{$0.965$} & \underline{$0.987$}
    \\
    + SDA ($N=3, a=120$) & $\mathbf{0.127}$ & \underline{$0.089$} & $0.489$ & $0.165$ & $0.843$ & \underline{$0.965$} & \underline{$0.987$}
    \\
    + SDA ($N=6, a=120$) & $\mathbf{0.127}$ & \underline{$0.089$} & \underline{$0.484$} & \underline{$0.163$} & \underline{$0.846$} & \underline{$0.965$} & $\mathbf{0.988}$
    \\
    + SDA ($N=12, a=120$) & $\mathbf{0.127}$ & $\mathbf{0.088}$ & $\mathbf{0.480}$ & $\mathbf{0.162}$ & $\mathbf{0.850}$ & $\mathbf{0.967}$ & $\mathbf{0.988}$
    \\\bottomrule
\end{tabular}
}
\label{tab:track2_innov_sda}
\end{table*}

\begin{table*}[t]
\caption{Ablation results of Frequency Domain Augmentation (FDA) with different hyperparameters on the testing set of the second track of the RoboDepth competition. The \textbf{best} and \underline{second best} scores of each metric are highlighted in \textbf{bold} and \underline{underline}, respectively.}
\centering\scalebox{0.78}{
    \begin{tabular}{l|cccc|ccc}
    \toprule
    \textbf{Method} & \cellcolor{blue!10}\textbf{Abs Rel~$\downarrow$} & \cellcolor{blue!10}\textbf{Sq Rel~$\downarrow$} & \cellcolor{blue!10}\textbf{RMSE~$\downarrow$} & \cellcolor{blue!10}\textbf{log RMSE~$\downarrow$} & \cellcolor{red!10}$\delta<1.25$~$\uparrow$ & \cellcolor{red!10}$\delta<1.25^2$~$\uparrow$ & \cellcolor{red!10}$\delta<1.25^3$~$\uparrow$
    \\\midrule\midrule
    DepthFormer \cite{li2022depthformer} & $0.131$ & $0.095$ & $0.507$ & $0.170$ & $0.827$ & $0.963$ & $0.987$
    \\\midrule
    + FDA ($\theta=3$) & \underline{$0.127$} & \underline{$0.089$} & $0.477$ & $0.163$ & $0.846$ & $0.966$ & $0.987$
    \\
    + FDA ($\theta=6$) & \underline{$0.127$} & \underline{$0.089$} & $0.477$ & $0.163$ & \underline{$0.847$} & $0.966$ & $0.987$
    \\
    + FDA ($\theta=12$) & \underline{$0.127$} & $\mathbf{0.087}$ & $0.471$ & \underline{$0.161$} & \underline{$0.847$} & \underline{$0.968$} & \underline{$0.988$}
    \\
    + FDA ($\theta=24$) & $\mathbf{0.128}$ & $\mathbf{0.087}$ & $\mathbf{0.462}$ & $\mathbf{0.160}$ & $\mathbf{0.850}$ & $\mathbf{0.969}$ & $\mathbf{0.989}$
    \\
    + FDA ($\theta=48$) & $0.131$ & $0.090$ & \underline{$0.464$} & \underline{$0.161$} & $\mathbf{0.850}$ & \underline{$0.968$} & $\mathbf{0.989}$
    \\\bottomrule
\end{tabular}
}
\label{tab:track2_innov_fda}
\end{table*}

Furthermore, we conduct extensive experiments in the frequency domain augmentation. Our method primarily focused on testing various rotation angles, as depicted in Table~\ref{tab:track2_innov_fda}. Ultimately, we found that the optimal value for $\theta$ is $24$ degrees. This is because excessively small $\theta$ values would result in minimal changes to the image, while excessively large values may lead to the loss of crucial information. Additionally, the partitioning of high-frequency and low-frequency information is an important parameter that we explored through experiments. Eventually, we discovered that the model would perform well when low-frequency information was preserved within a rectangular region of size $50\times50$ at the center of the frequency domain image.

\subsubsection{Solution Summary}
In this work, to address the challenge of robust monocular depth estimation, we employed a multi-stage approach. Firstly, we utilized a masking technique to selectively occlude regions in the input image to enhance the spatial domain representation learning. Subsequently, we performed frequency domain operations to separate the high-frequency and low-frequency information of the image, followed by a series of robustness-enhancing techniques applied specifically to the high-frequency information. By employing these two augmentations, we significantly improved the model’s robustness on the OoD testing dataset. We evaluated our proposed approach in the second track of the RoboDepth Challenge and achieved the innovative prize.

\section{Conclusion \& Future Direction}
\label{sec:conclusion}
This paper has presented the results of the RoboDepth Challenge. In this challenge, we were dedicated to encouraging out-of-distribution depth estimation to meet safety-critical requirements. Our top-performing participants presented diverse designs and techniques that are proven useful in improving the robustness of depth estimation models under common corruptions. We compared and summarized robustness enhancement strategies from different aspects and drew insightful observations behind them. It is worth mentioning again that promising results have been achieved on the challenging testing sets across two tracks of this challenge. We believe that the  RoboDepth Challenge has shed light on the development of depth estimation systems that are reliable, scalable, and generalizable.

Future editions of the RoboDepth Challenge seek further improvements from the following aspects:
\begin{itemize}
    \item Extension of the scale and diversity of robustness evaluation sets. The current RoboDepth benchmarks only considered two distinct data sources and five discrete severity levels. Simulating continuous severity changes on more depth estimation datasets is desirable.
    \item Integration of more depth estimation tasks. While this challenge mainly focused on monocular depth estimation, it is important to study also the robustness of other related tasks, such as stereo, panorama, and surrounding-view depth estimation.
    \item Exploration of other techniques that could improve the OoD robustness. The recent vision foundation models have opened up new possibilities for unified and generalizable visual perception. It would be interesting to combine these models for robust depth estimation.
    \item Pursuit of both robustness and efficiency. Since the depth estimation system might require in-vehicle deployment, the use of certain techniques, such as model ensemble and TTA, would become unreasonable. It is thus crucial to design suitable latency constraints.
\end{itemize}

\section{Acknowledgements}
\label{sec:acknowledgements}

This competition is sponsored by Baidu Research, USA (\url{http://research.baidu.com}). 

This research is part of the programme DesCartes and is supported by the National Research Foundation, Prime Minister’s Office, Singapore under its Campus for Research Excellence and Technological Enterprise (CREATE) programme. This work is affiliated with the WP4 of the DesCartes programme, with an identity number: A-8000237-00-00.

We sincerely thank the support from the ICRA 2023 organizing committee.

\section{Appendix}
\label{sec:appendix}

In this appendix, we supplement the following materials to support the findings and conclusions drawn in the main body of this paper:
\begin{itemize}
    \item Section~\ref{sec:certificates} attaches the certificates that are awarded to our participants.
    \item Section~\ref{sec:public-resources-used} acknowledges the public resources used during the course of this work.
\end{itemize}

\subsection{Competition Certificates}
\label{sec:certificates}

\begin{landscape}
\begin{figure}
    \centering
    \includegraphics[width=0.8\linewidth]{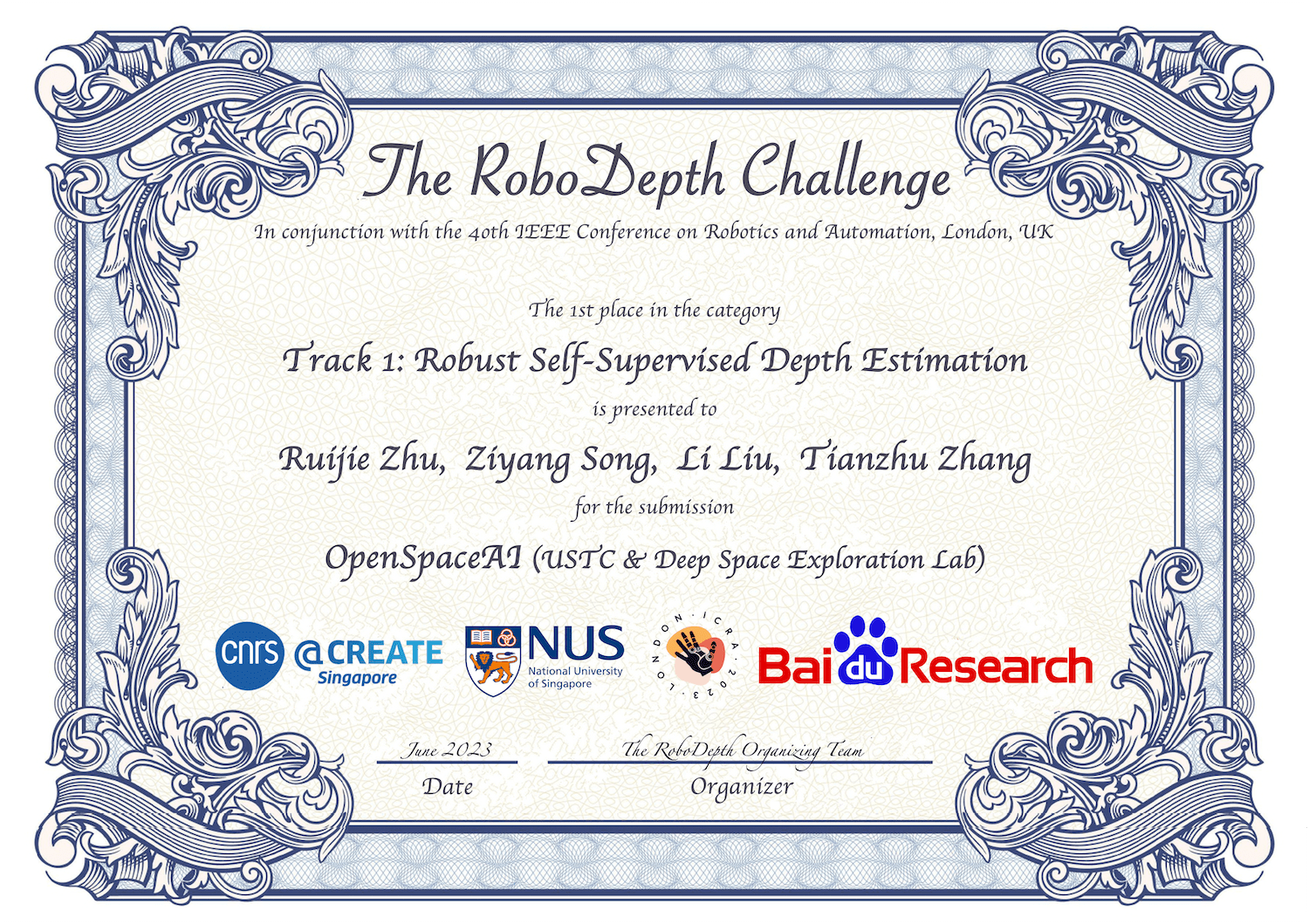}
    \caption{The certificate awarded to the \texttt{OpenSpaceAI} team in the first track of the RoboDepth Challenge.}
\label{fig:certificates_track1_1st}
\end{figure}
\end{landscape}

\begin{landscape}
\begin{figure}
    \centering
    \includegraphics[width=0.8\linewidth]{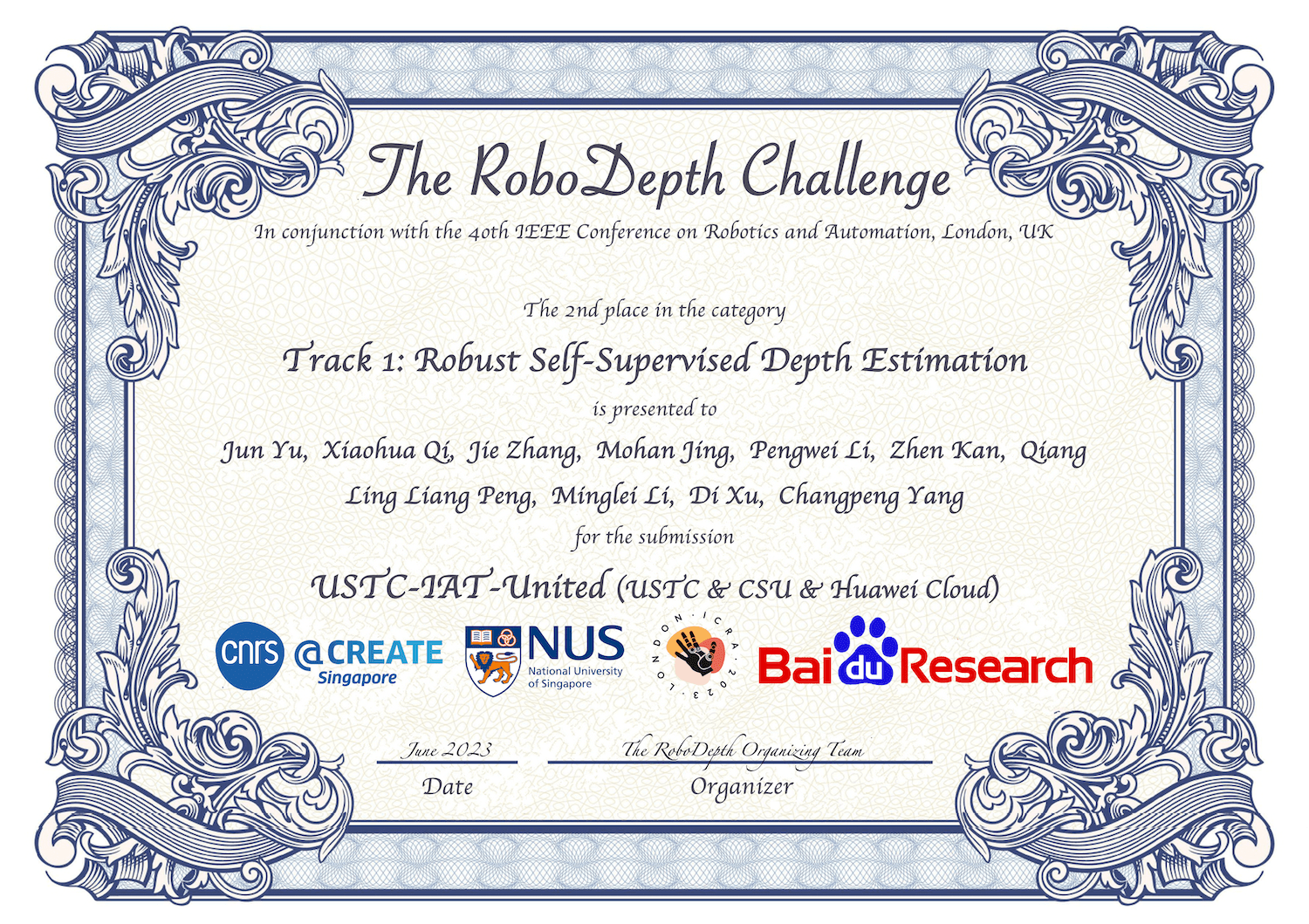}
    \caption{The certificate awarded to the \texttt{USTC-IAT-United} team in the first track of the RoboDepth Challenge.}
\label{fig:certificates_track1_2nd}
\end{figure}
\end{landscape}

\begin{landscape}
\begin{figure}
    \centering
    \includegraphics[width=0.8\linewidth]{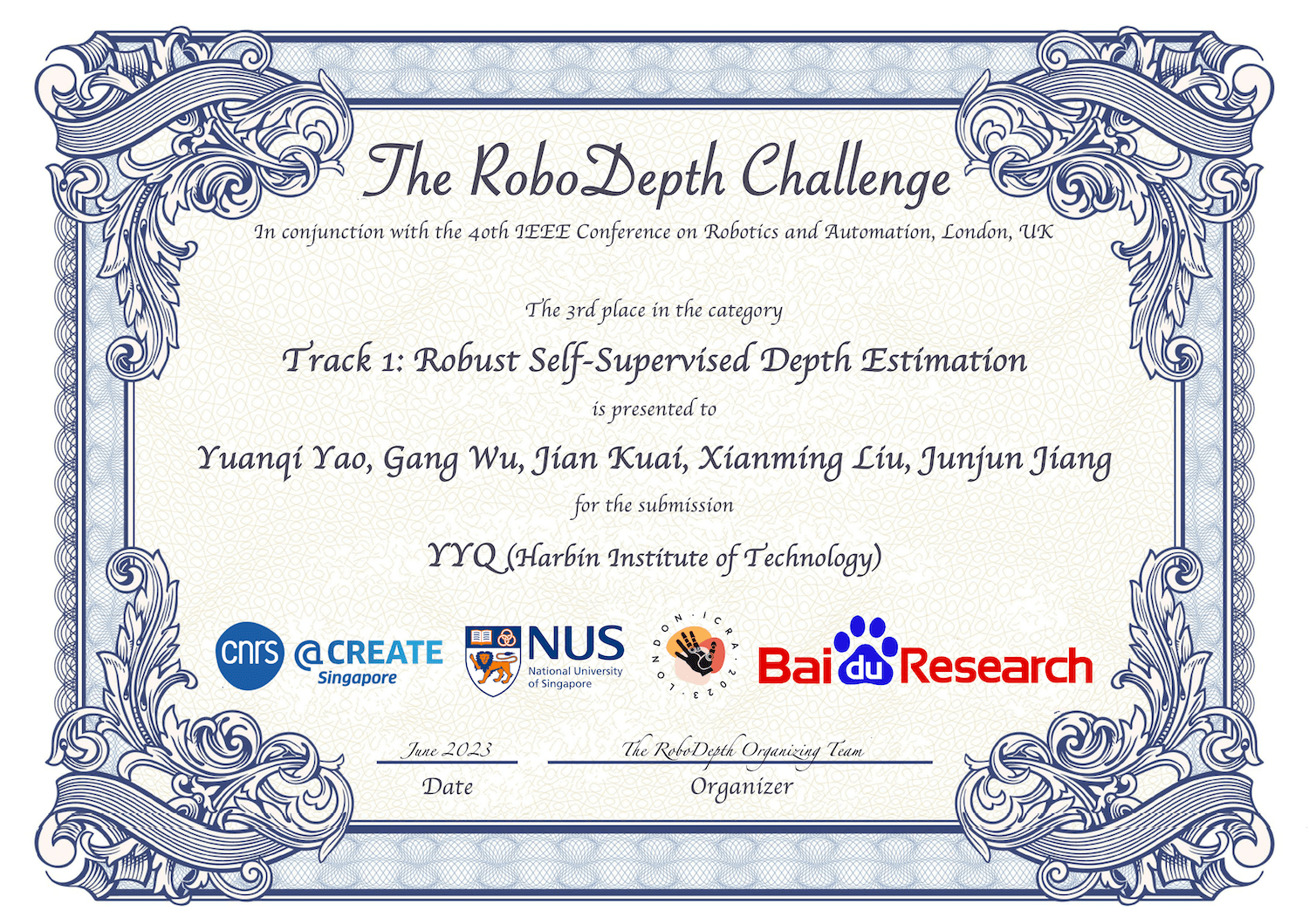}
    \caption{The certificate awarded to the \texttt{YYQ} team in the first track of the RoboDepth Challenge.}
\label{fig:certificates_track1_3rd}
\end{figure}
\end{landscape}

\begin{landscape}
\begin{figure}
    \centering
    \includegraphics[width=0.8\linewidth]{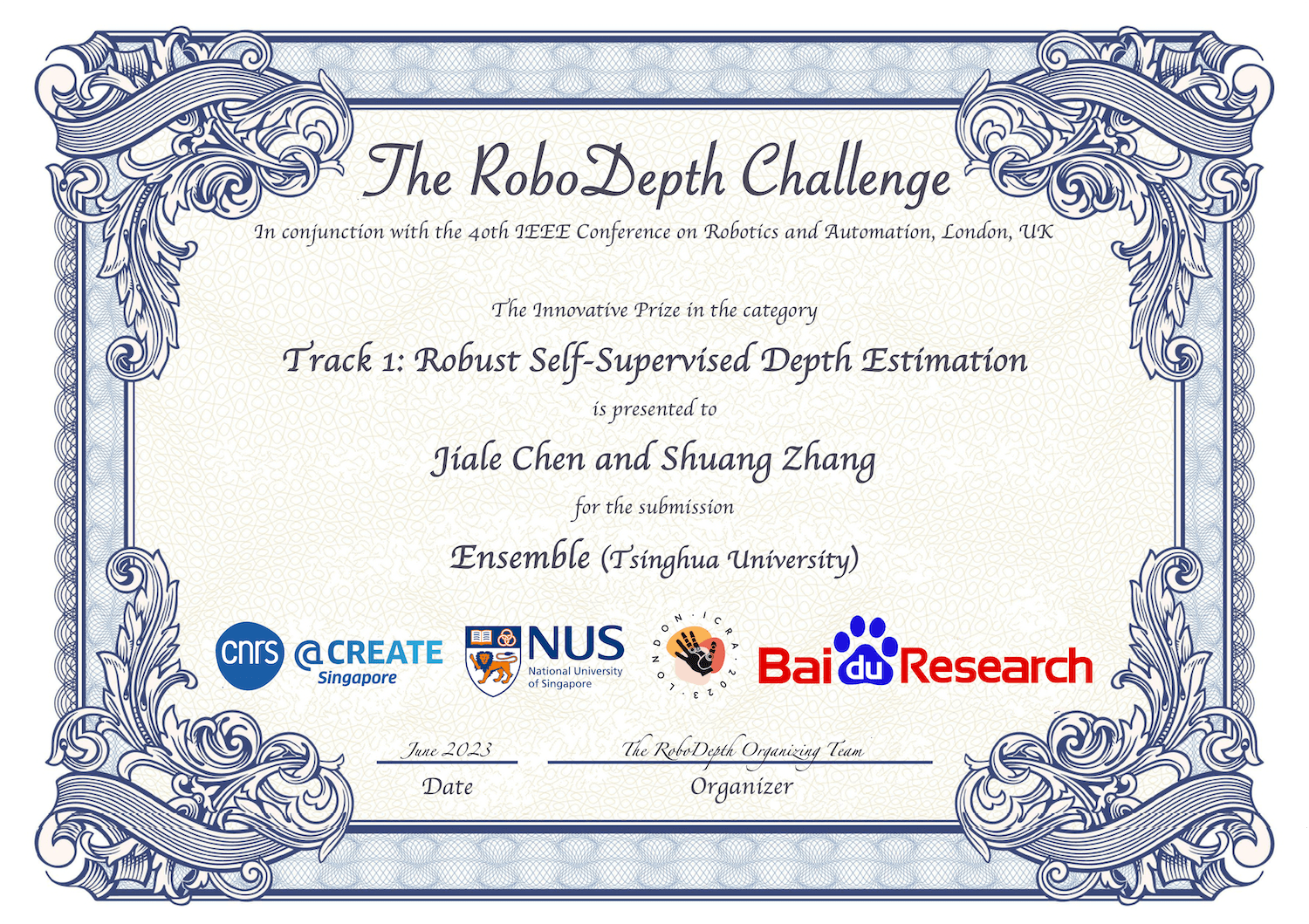}
    \caption{The certificate awarded to the \texttt{Ensemble} team in the first track of the RoboDepth Challenge.}
\label{fig:certificates_track1_innov1}
\end{figure}
\end{landscape}

\begin{landscape}
\begin{figure}
    \centering
    \includegraphics[width=0.8\linewidth]{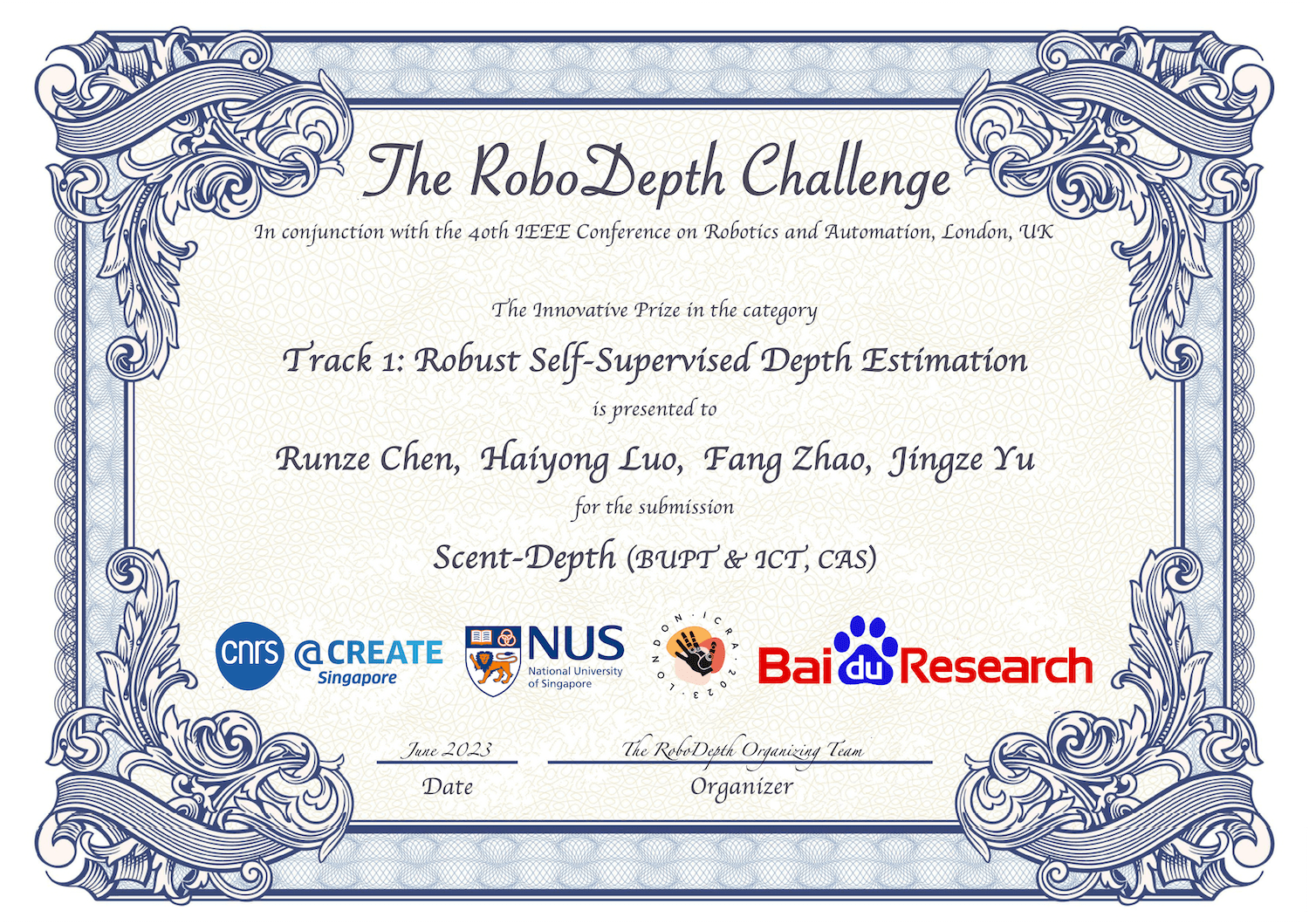}
    \caption{The certificate awarded to the \texttt{Scent-Depth} team in the first track of the RoboDepth Challenge.}
\label{fig:certificates_track1_innov2}
\end{figure}
\end{landscape}

\begin{landscape}
\begin{figure}
    \centering
    \includegraphics[width=0.8\linewidth]{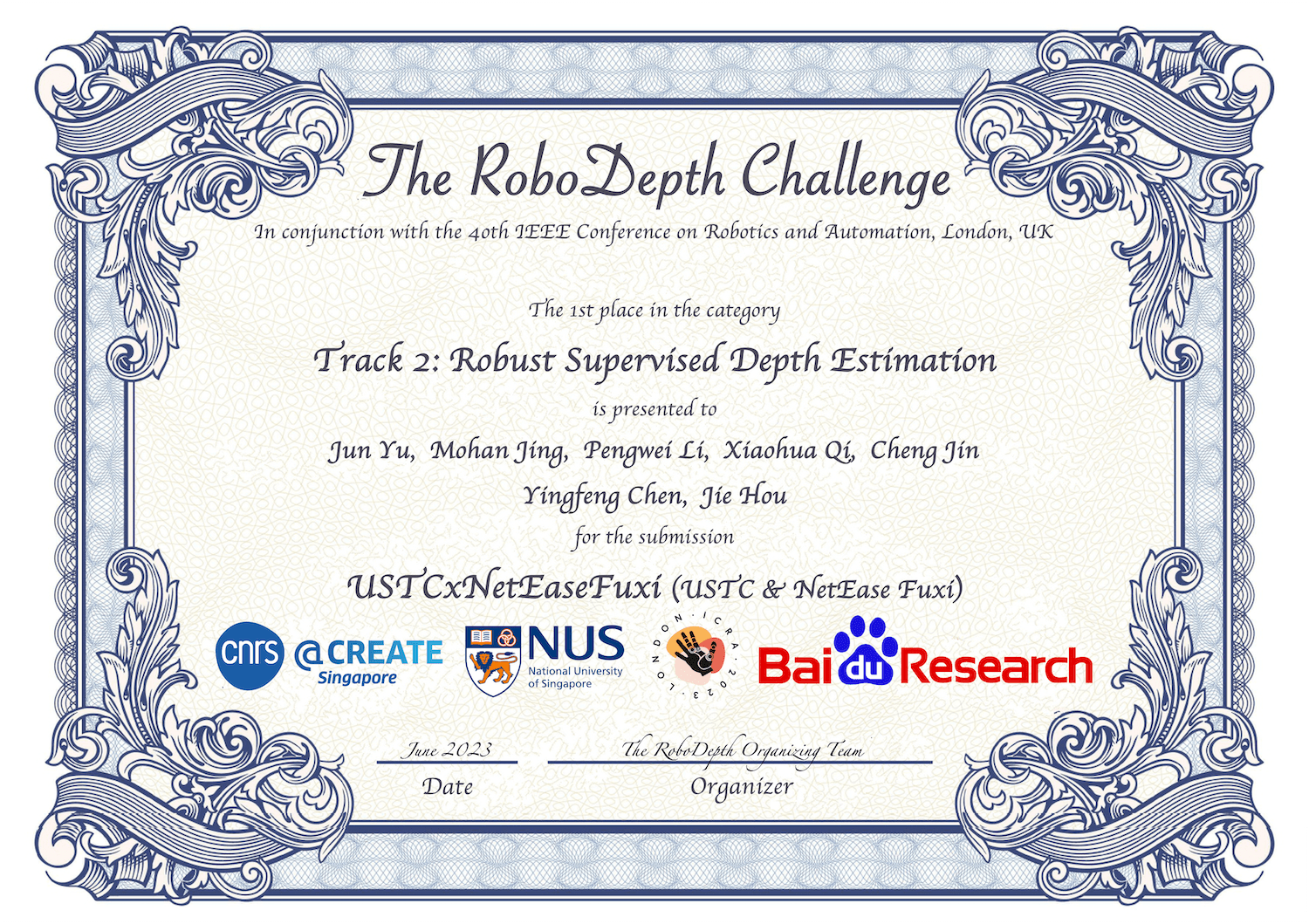}
    \caption{The certificate awarded to the \texttt{USTCxNetEaseFuxi} team in the second track of the RoboDepth Challenge.}
\label{fig:certificates_track2_1st}
\end{figure}
\end{landscape}

\begin{landscape}
\begin{figure}
    \centering
    \includegraphics[width=0.8\linewidth]{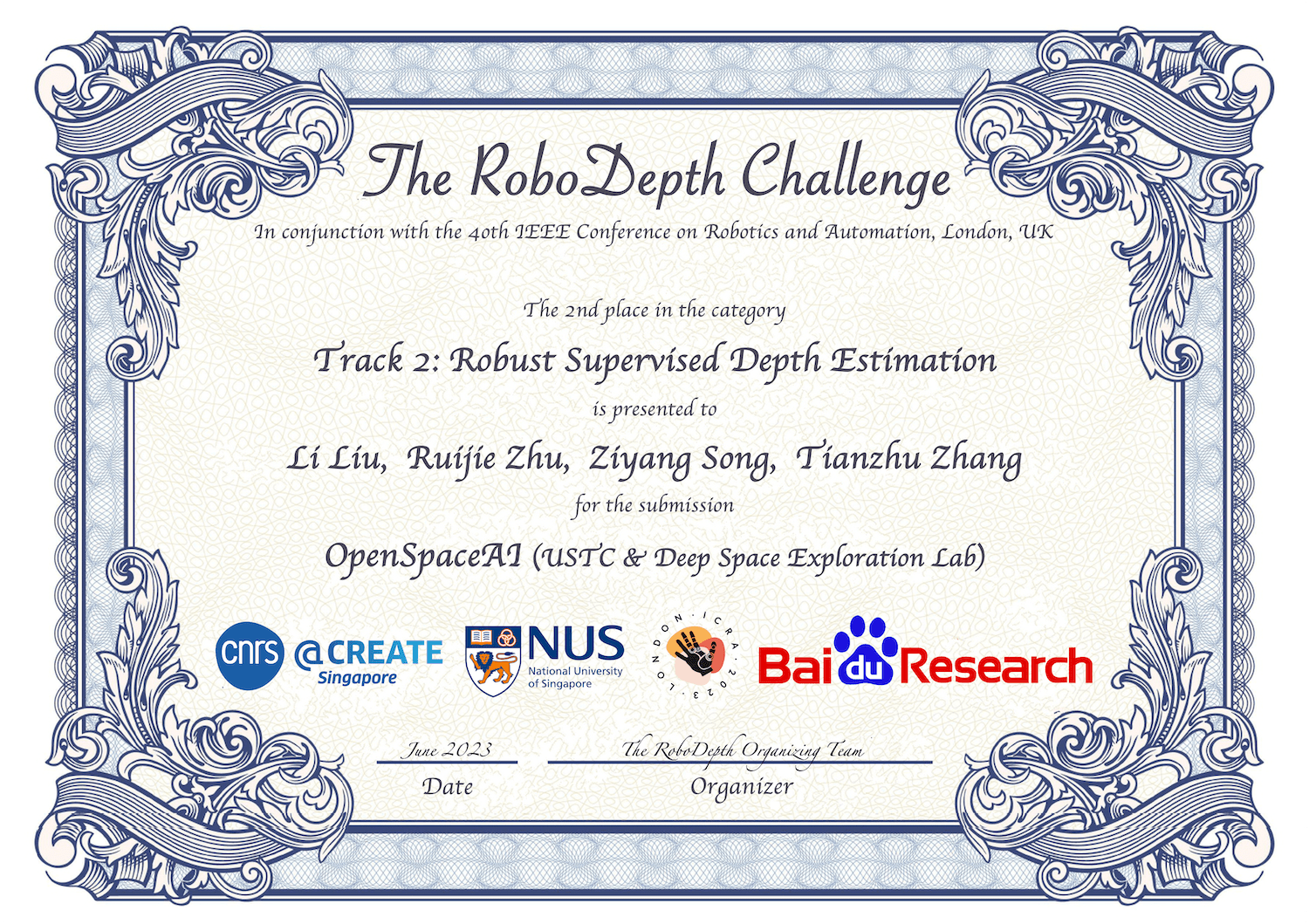}
    \caption{The certificate awarded to the \texttt{OpenSpaceAI} team in the second track of the RoboDepth Challenge.}
\label{fig:certificates_track2_2nd}
\end{figure}
\end{landscape}

\begin{landscape}
\begin{figure}
    \centering
    \includegraphics[width=0.8\linewidth]{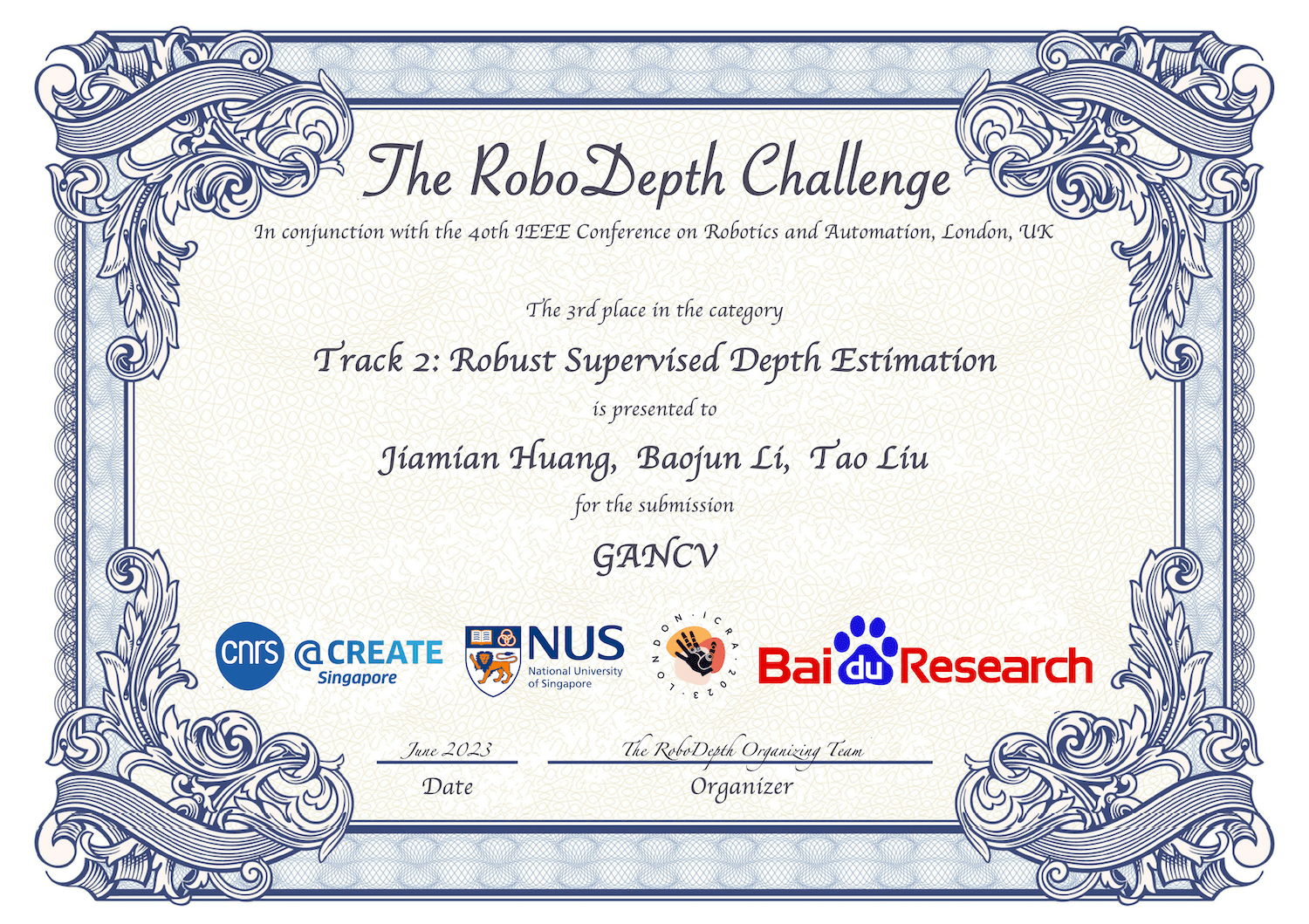}
    \caption{The certificate awarded to the \texttt{GANCV} team in the second track of the RoboDepth Challenge.}
\label{fig:certificates_track2_3rd}
\end{figure}
\end{landscape}

\begin{landscape}
\begin{figure}
    \centering
    \includegraphics[width=0.8\linewidth]{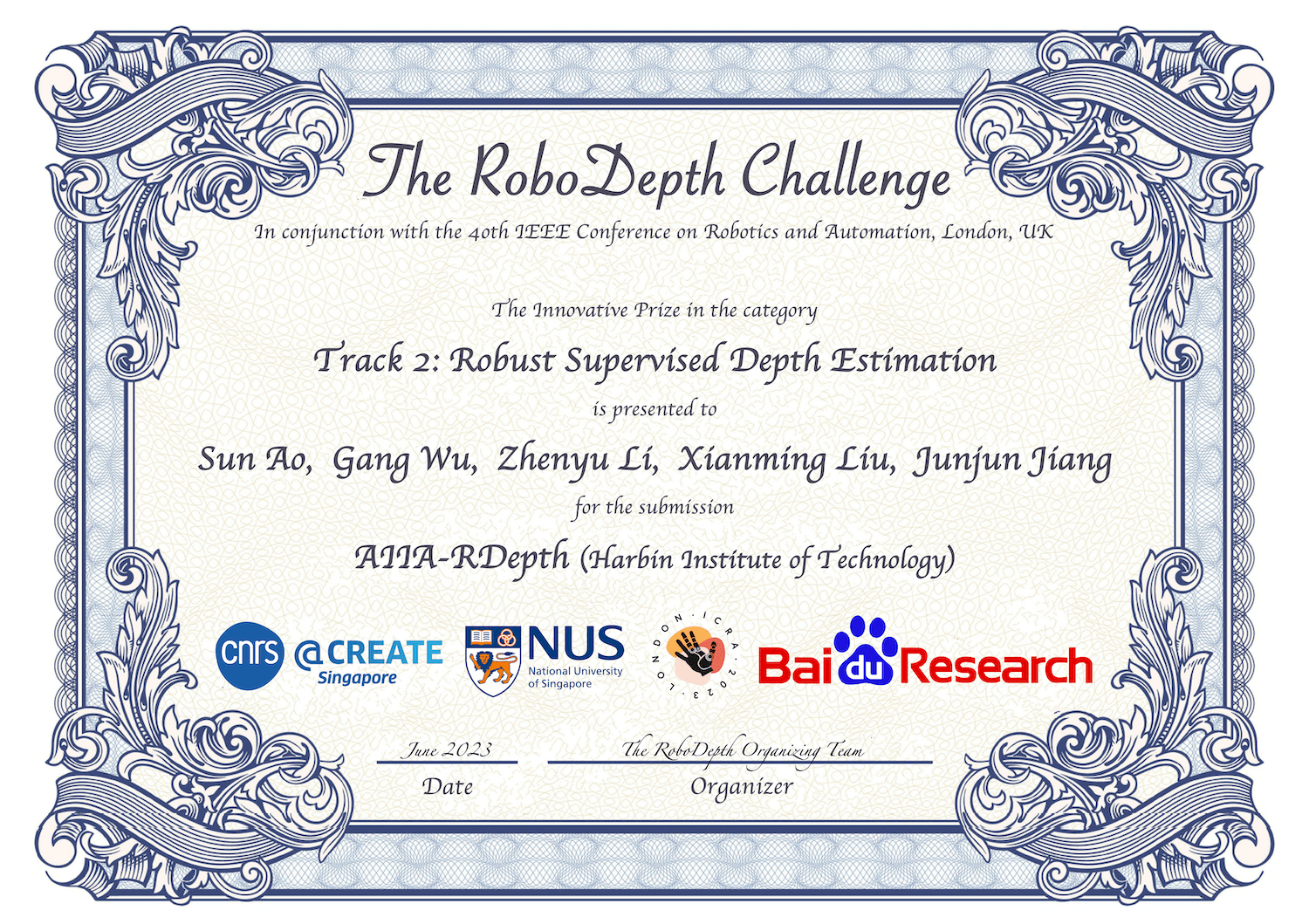}
    \caption{The certificate awarded to the \texttt{AIIA-RDepth} team in the second track of the RoboDepth Challenge.}
\label{fig:certificates_track2_innov}
\end{figure}
\end{landscape}

In this section, we attach the certificates that are awarded to the top-performing participants in the RoboDepth Challenge. Specifically, the certificates awarded to winners from the first track are shown in Figure~\ref{fig:certificates_track1_1st}, Figure~\ref{fig:certificates_track1_2nd}, Figure~\ref{fig:certificates_track1_3rd}, Figure~\ref{fig:certificates_track1_innov1}, and Figure~\ref{fig:certificates_track1_innov2}. The certificates awarded to winners from the second track are shown in Figure~\ref{fig:certificates_track2_1st}, Figure~\ref{fig:certificates_track2_2nd}, Figure~\ref{fig:certificates_track2_3rd}, and Figure~\ref{fig:certificates_track2_innov}.

\clearpage

\clearpage

\clearpage

\clearpage

\clearpage

\clearpage

\clearpage

\clearpage

\clearpage

\clearpage
\subsection{Public Resources Used}
\label{sec:public-resources-used}

In this section, we acknowledge the use of public resources, during the course of this work:
\begin{itemize}
    \item KITTI Vision Benchmark\footnote{\url{https://www.cvlibs.net/datasets/kitti}.} \dotfill CC BY-NC-SA 3.0
    \item NYU Depth Dataset V2\footnote{\url{https://cs.nyu.edu/~silberman/datasets/nyu_depth_v2.html}.} \dotfill Unknown
    \item MonoDepth2\footnote{\url{https://github.com/nianticlabs/monodepth2}.} \dotfill Custom MonoDepth2 License
    \item MonoViT\footnote{\url{https://github.com/zxcqlf/MonoViT}.} \dotfill MIT License
    \item Lite-Mono\footnote{\url{https://github.com/noahzn/Lite-Mono}.} \dotfill Unknown
    \item Monocular-Depth-Estimation-Toolbox\footnote{\url{https://github.com/zhyever/Monocular-Depth-Estimation-Toolbox}.} \dotfill Apache License 2.0
    \item DepthFormer\footnote{\url{https://github.com/zhyever/Monocular-Depth-Estimation-Toolbox/tree/main/configs/depthformer}.} \dotfill Apache License 2.0
    \item ImageCorruptions\footnote{\url{https://github.com/bethgelab/imagecorruptions}.} \dotfill Apache License 2.0
    \item 3DCC\footnote{\url{https://github.com/EPFL-VILAB/3DCommonCorruptions}.} \dotfill Attribution-NC 4.0 International
    \item ImageNet-C\footnote{\url{https://github.com/hendrycks/robustness}.} \dotfill Apache License 2.0
\end{itemize}

\clearpage

\bibliographystyle{plain}
\bibliography{ref}

\begin{thebibliography}{100}

\bibitem{behley2019semanticKITTI}
Jens Behley, Martin Garbade, Andres Milioto, Jan Quenzel, Sven Behnke, Cyrill
  Stachniss, and Juergen Gall.
\newblock Semantickitti: A dataset for semantic scene understanding of lidar
  sequences.
\newblock In {\em IEEE/CVF International Conference on Computer Vision (ICCV)},
  pages 9297--9307, 2019.

\bibitem{bhat2021adabins}
Shariq~Farooq Bhat, Ibraheem Alhashim, and Peter Wonka.
\newblock Adabins: Depth estimation using adaptive bins.
\newblock In {\em IEEE/CVF Conference on Computer Vision and Pattern
  Recognition (CVPR)}, pages 4009--4018, 2021.

\bibitem{bhat2023zero}
Shariq~Farooq Bhat, Reiner Birkl, Diana Wofk, Peter Wonka, and Matthias
  Müller.
\newblock Zoedepth: Zero-shot transfer by combining relative and metric depth.
\newblock {\em arXiv preprint arXiv:2302.12288}, 2023.

\bibitem{caesar2020nuScenes}
Holger Caesar, Varun Bankiti, Alex~H Lang, Sourabh Vora, Venice~Erin Liong,
  Qiang Xu, Anush Krishnan, Yu~Pan, Giancarlo Baldan, and Oscar Beijbom.
\newblock nuscenes: A multimodal dataset for autonomous driving.
\newblock In {\em IEEE/CVF Conference on Computer Vision and Pattern
  Recognition (CVPR)}, pages 11621--11631, 2020.

\bibitem{caron2021dino}
Mathilde Caron, Hugo Touvron, Ishan Misra, Hervé Jégou, Julien Mairal, Piotr
  Bojanowski, and Armand Joulin.
\newblock Emerging properties in self-supervised vision transformers.
\newblock In {\em IEEE/CVF International Conference on Computer Vision (ICCV)},
  pages 9650--9660, 2021.

\bibitem{RobustNav}
Prithvijit Chattopadhyay, Judy Hoffman, Roozbeh Mottaghi, and Aniruddha
  Kembhavi.
\newblock Robustnav: Towards benchmarking robustness in embodied navigation.
\newblock In {\em IEEE/CVF International Conference on Computer Vision (ICCV)},
  pages 15691--15700, 2021.

\bibitem{chen2021apr}
Guangyao Chen, Peixi Peng, Li~Ma, Jia Li, Lin Du, and Yonghong Tian.
\newblock Amplitude-phase recombination: Rethinking robustness of convolutional
  neural networks in frequency domain.
\newblock In {\em IEEE/CVF Conference on Computer Vision and Pattern
  Recognition (CVPR)}, pages 458--467, 2021.

\bibitem{chen2021amplitude}
Guangyao Chen, Peixi Peng, Li~Ma, Jia Li, Lin Du, and Yonghong Tian.
\newblock Amplitude-phase recombination: Rethinking robustness of convolutional
  neural networks in frequency domain.
\newblock In {\em IEEE/CVF International Conference on Computer Vision (ICCV)},
  pages 458--467, 2021.

\bibitem{chen2021ipt}
Hanting Chen, Yunhe Wang, Tianyu Guo, Chang Xu, Yiping Deng, Zhenhua Liu, Siwei
  Ma, Chunjing Xu, Chao Xu, and Wen Gao.
\newblock Pre-trained image processing transformer.
\newblock In {\em IEEE/CVF Conference on Computer Vision and Pattern
  Recognition (CVPR)}, pages 12299--12310, 2021.

\bibitem{chen2023towards}
Runnan Chen, Youquan Liu, Lingdong Kong, Nenglun Chen, Xinge Zhu, Yuexin Ma,
  Tongliang Liu, and Wenping Wang.
\newblock Towards label-free scene understanding by vision foundation models.
\newblock {\em arXiv preprint arXiv:2306.03899}, 2023.

\bibitem{chen2023clip2Scene}
Runnan Chen, Youquan Liu, Lingdong Kong, Xinge Zhu, Yuexin Ma, Yikang Li,
  Yuenan Hou, Yu~Qiao, and Wenping Wang.
\newblock Clip2scene: Towards label-efficient 3d scene understanding by clip.
\newblock In {\em IEEE/CVF Conference on Computer Vision and Pattern
  Recognition (CVPR)}, pages 7020--7030, 2023.

\bibitem{chen2023tridepth}
Xingyu Chen, Ruonan Zhang, Ji~Jiang, Yan Wang, Ge~Li, and Thomas~H Li.
\newblock Self-supervised monocular depth estimation: Solving the
  edge-fattening problem.
\newblock In {\em IEEE/CVF Winter Conference on Applications of Computer Vision
  (WACV)}, pages 5776--5786, 2023.

\bibitem{cheng2022physical}
Zhiyuan Cheng, James Liang, Hongjun Choi, Guanhong Tao, Zhiwen Cao, Dongfang
  Liu, and Xiangyu Zhang.
\newblock Physical attack on monocular depth estimation with optimal
  adversarial patches.
\newblock In {\em European Conference on Computer Vision (ECCV)}, pages
  514--532, 2022.

\bibitem{cordts2016cityscapes}
Marius Cordts, Mohamed Omran, Sebastian Ramos, Timo Rehfeld, Markus Enzweiler,
  Rodrigo Benenson, Uwe Franke, Stefan Roth, and Bernt Schiele.
\newblock The cityscapes dataset for semantic urban scene understanding.
\newblock In {\em IEEE/CVF Conference on Computer Vision and Pattern
  Recognition (CVPR)}, pages 3213--3223, 2016.

\bibitem{cubuk2019autoaugment}
Ekin~D. Cubuk, Barret Zoph, Dandelion Mane, Vijay Vasudevan, and Quoc~V. Le.
\newblock Autoaugment: Learning augmentation strategies from data.
\newblock In {\em IEEE/CVF Conference on Computer Vision and Pattern
  Recognition (CVPR)}, pages 113--123, 2019.

\bibitem{devries2017cutout}
Terrance DeVries and Graham~W. Taylor.
\newblock Improved regularization of convolutional neural networks with cutout.
\newblock {\em arXiv preprint arXiv:1708.04552}, 2017.

\bibitem{dong2022survey}
Xingshuai Dong, Matthew~A. Garratt, Sreenatha~G. Anavatti, and Hussein~A.
  Abbass.
\newblock Towards real-time monocular depth estimation for robotics: A survey.
\newblock {\em IEEE Transactions on Intelligent Transportation Systems (TITS)},
  23(10):16940--16961, 2022.

\bibitem{dosovitskiy2020vit}
Alexey Dosovitskiy, Lucas Beyer, Alexander Kolesnikov, Dirk Weissenborn,
  Xiaohua Zhai, Thomas Unterthiner, Mostafa Dehghani, Matthias Minderer, Georg
  Heigold, Sylvain Gelly, Jakob Uszkoreit, and Neil Houlsby.
\newblock An image is worth 16x16 words: Transformers for image recognition at
  scale.
\newblock In {\em International Conference on Learning Representations (ICLR)},
  2020.

\bibitem{duan2021advdrop}
Ranjie Duan, Yuefeng Chen, Dantong Niu, Yun Yang, A.~Kai Qin, and Yuan He.
\newblock Advdrop: Adversarial attack to dnns by dropping information.
\newblock In {\em IEEE/CVF International Conference on Computer Vision (ICCV)},
  pages 7506--7515, 2021.

\bibitem{eigen2015predicting}
David Eigen and Rob Fergus.
\newblock Predicting depth, surface normals and semantic labels with a common
  multi-scale convolutional architecture.
\newblock In {\em IEEE/CVF International Conference on Computer Vision (ICCV)},
  pages 2650--2658, 2015.

\bibitem{eigen2014depth}
David Eigen, Christian Puhrsch, and Rob Fergus.
\newblock Depth map prediction from a single image using a multi-scale deep
  network.
\newblock In {\em Advances in Neural Information Processing System (NeurIPS)},
  2014.

\bibitem{esser2021vqgan}
Patrick Esser, Robin Rombach, and Bjorn Ommer.
\newblock Taming transformers for high-resolution image synthesis.
\newblock In {\em IEEE/CVF Conference on Computer Vision and Pattern
  Recognition (CVPR)}, pages 12873--12883, 2021.

\bibitem{VOC}
Mark Everingham, Luc Gool, Christopher~K. Williams, John Winn, and Andrew
  Zisserman.
\newblock The pascal visual object classes (voc) challenge.
\newblock {\em International Journal of Computer Vision (IJCV)},
  88(2):303–338, 2010.

\bibitem{fong2022panoptic-nuScenes}
Whye~Kit Fong, Rohit Mohan, Juana~Valeria Hurtado, Lubing Zhou, Holger Caesar,
  Oscar Beijbom, and Abhinav Valada.
\newblock Panoptic nuscenes: A large-scale benchmark for lidar panoptic
  segmentation and tracking.
\newblock {\em IEEE Robotics and Automation Letters (RA-L)}, pages 3795--3802,
  2022.

\bibitem{DDAD}
Adrien Gaidon, Greg Shakhnarovich, Rares Ambrus, Vitor Guizilini, Igor
  Vasiljevic, Matthew Walter, Sudeep Pillai, and Nick Kolkin.
\newblock The dense depth for autonomous driving (ddad) challenge.
\newblock \url{https://sites.google.com/view/mono3d-workshop}, 2021.

\bibitem{garg2016unsupervised}
Ravi Garg, BG~Vijay Kumar, Gustavo Carneiro, and Ian Reid.
\newblock Unsupervised cnn for single view depth estimation: Geometry to the
  rescue.
\newblock In {\em European Conference on Computer Vision (ECCV)}, pages
  740--756, 2016.

\bibitem{geiger2012kitti}
Andreas Geiger, Philip Lenz, and Raquel Urtasun.
\newblock Are we ready for autonomous driving? the kitti vision benchmark
  suite.
\newblock In {\em IEEE/CVF Conference on Computer Vision and Pattern
  Recognition (CVPR)}, pages 3354--3361, 2012.

\bibitem{geirhos2018imagenet-trained}
Robert Geirhos, Patricia Rubisch, Claudio Michaelis, Matthias Bethge, Felix~A
  Wichmann, and Wieland Brendel.
\newblock Imagenet-trained {CNN}s are biased towards texture; increasing shape
  bias improves accuracy and robustness.
\newblock In {\em International Conference on Learning Representations (ICLR)},
  2019.

\bibitem{godard2017unsupervised}
Clément Godard, Oisin~Mac Aodha, and Gabriel~J. Brostow.
\newblock Unsupervised monocular depth estimation with left-right consistency.
\newblock In {\em IEEE/CVF Conference on Computer Vision and Pattern
  Recognition (CVPR)}, pages 270--279, 2017.

\bibitem{godard2019monodepth2}
Clément Godard, Oisin~Mac Aodha, Michael Firman, and Gabriel~J. Brostow.
\newblock Digging into self-supervised monocular depth prediction.
\newblock In {\em IEEE/CVF International Conference on Computer Vision (ICCV)},
  pages 3828--3838, 2019.

\bibitem{he2022mae}
Kaiming He, Xinlei Chen, Saining Xie, Yanghao Li, Piotr Dollár, and Ross
  Girshick.
\newblock Masked autoencoders are scalable vision learners.
\newblock In {\em IEEE/CVF Conference on Computer Vision and Pattern
  Recognition (CVPR)}, pages 16000--16009, 2022.

\bibitem{ImageNet-C}
Dan Hendrycks and Thomas Dietterich.
\newblock Benchmarking neural network robustness to common corruptions and
  perturbations.
\newblock In {\em International Conference on Learning Representations (ICLR)},
  2019.

\bibitem{hendrycks2020augmix}
Dan Hendrycks, Norman Mu, Ekin~D. Cubuk, Barret Zoph, Justin Gilmer, and Balaji
  Lakshminarayanan.
\newblock Augmix: A simple data processing method to improve robustness and
  uncertainty.
\newblock In {\em International Conference on Learning Representations (ICLR)},
  2020.

\bibitem{hu2022seasondepth}
Hanjiang Hu, Baoquan Yang, Zhijian Qiao, Shiqi Liu, Ding Zhao, and Hesheng
  Wang.
\newblock Seasondepth: Cross-season monocular depth prediction dataset and
  benchmark under multiple environments.
\newblock In {\em International Conference on Machine Learning Workshops
  (ICMLW)}, 2022.

\bibitem{huang2017arbitrary}
Xun Huang and Serge Belongie.
\newblock Arbitrary style transfer in real-time with adaptive instance
  normalization.
\newblock In {\em IEEE/CVF International Conference on Computer Vision (ICCV)},
  pages 1501--1510, 2017.

\bibitem{MobileAI}
Andrey Ignatov, Grigory Malivenko, David Plowman, Samarth Shukla, and Radu
  Timofte.
\newblock Fast and accurate single-image depth estimation on mobile devices,
  mobile ai 2021 challenge: Report.
\newblock In {\em IEEE/CVF Conference on Computer Vision and Pattern
  Recognition Workshops (CVPRW)}, pages 2545--2557, 2021.

\bibitem{jaderberg2015spatial}
Max Jaderberg, Karen Simonyan, and Andrew Zisserman.
\newblock Spatial transformer networks.
\newblock In {\em Advances in Neural Information Processing Systems (NeurIPS)},
  volume~28, 2015.

\bibitem{ji2019semi}
Rongrong Ji, Ke~Li, Yan Wang, Xiaoshuai Sun, Feng Guo, Xiaowei Guo, Yongjian
  Wu, Feiyue Huang, and Jiebo Luo.
\newblock Semi-supervised adversarial monocular depth estimation.
\newblock {\em IEEE Transactions on Pattern Analysis and Machine Intelligence
  (PAMI)}, 42(10):2410--2422, 2019.

\bibitem{johnston2020self}
Adrian Johnston and Gustavo Carneiro.
\newblock Self-supervised monocular trained depth estimation using
  self-attention and discrete disparity volume.
\newblock In {\em IEEE/CVF Conference on Computer Vision and Pattern
  Recognition (CVPR)}, pages 4756--4765, 2020.

\bibitem{jung2021fsre}
Hyunyoung Jung, Eunhyeok Park, and Sungjoo Yoo.
\newblock Fine-grained semantics-aware representation enhancement for
  self-supervised monocular depth estimation.
\newblock In {\em IEEE/CVF International Conference on Computer Vision (ICCV)},
  pages 12642--12652, 2021.

\bibitem{Cityscapes-C}
Christoph Kamann and Carsten Rother.
\newblock Benchmarking the robustness of semantic segmentation models.
\newblock In {\em IEEE/CVF Conference on Computer Vision and Pattern
  Recognition (CVPR)}, pages 8828--8838, 2020.

\bibitem{kannan2018pairing}
Harini Kannan, Alexey Kurakin, and Ian Goodfellow.
\newblock Adversarial logit pairing.
\newblock {\em arXiv preprint arXiv:1803.06373}, 2018.

\bibitem{kar20223d}
O{\u{g}}uzhan~Fatih Kar, Teresa Yeo, Andrei Atanov, and Amir Zamir.
\newblock 3d common corruptions and data augmentation.
\newblock In {\em IEEE/CVF Conference on Computer Vision and Pattern
  Recognition (CVPR)}, pages 18963--18974, 2022.

\bibitem{kirillov2019fpn}
Alexander Kirillov, Ross Girshick, Kaiming He, and Piotr Dollár.
\newblock Panoptic feature pyramid networks.
\newblock In {\em IEEE/CVF Conference on Computer Vision and Pattern
  Recognition (CVPR)}, pages 6399--6408, 2019.

\bibitem{kirillov2023segment}
Alexander Kirillov, Eric Mintun, Nikhila Ravi, Hanzi Mao, Chloe Rolland, Laura
  Gustafson, Tete Xiao, Spencer Whitehead, Alexander~C. Berg, Wan-Yen Lo, Piotr
  Dollár, and Ross Girshick.
\newblock Segment anything.
\newblock {\em arXiv preprint arXiv:2304.02643}, 2023.

\bibitem{klodt2018supervising}
Maria Klodt and Andrea Vedaldi.
\newblock Supervising the new with the old: learning sfm from sfm.
\newblock In {\em European Conference on Computer Vision (ECCV)}, pages
  698--713, 2018.

\bibitem{kong2023rethinking}
Lingdong Kong, Youquan Liu, Runnan Chen, Yuexin Ma, Xinge Zhu, Yikang Li,
  Yuenan Hou, Yu~Qiao, and Ziwei Liu.
\newblock Rethinking range view representation for lidar segmentation.
\newblock {\em arXiv preprint arXiv:2303.05367}, 2023.

\bibitem{kong2023robo3d}
Lingdong Kong, Youquan Liu, Xin Li, Runnan Chen, Wenwei Zhang, Jiawei Ren,
  Liang Pan, Kai Chen, and Ziwei Liu.
\newblock Robo3d: Towards robust and reliable 3d perception against
  corruptions.
\newblock {\em arXiv preprint arXiv:2303.17597}, 2023.

\bibitem{kong2023conDA}
Lingdong Kong, Niamul Quader, and Venice~Erin Liong.
\newblock Conda: Unsupervised domain adaptation for lidar segmentation via
  regularized domain concatenation.
\newblock In {\em IEEE International Conference on Robotics and Automation
  (ICRA)}, pages 9338--9345, 2023.

\bibitem{kong2022laserMix}
Lingdong Kong, Jiawei Ren, Liang Pan, and Ziwei Liu.
\newblock Lasermix for semi-supervised lidar semantic segmentation.
\newblock In {\em IEEE/CVF Conference on Computer Vision and Pattern
  Recognition (CVPR)}, pages 21705--21715, 2023.

\bibitem{kong2023robodepth_benchmark}
Lingdong Kong, Shaoyuan Xie, Hanjiang Hu, Benoit Cottereau, Lai~Xing Ng, and
  Wei~Tsang Ooi.
\newblock The robodepth benchmark for robust out-of-distribution depth
  estimation under corruptions.
\newblock \url{https://github.com/ldkong1205/RoboDepth}, 2023.

\bibitem{ArgoverseStereo}
Henrik Kretzschmar, Alex Liniger, Jose~M. Alvarez, Yan Wang, Vincent Casser,
  Fisher Yu, Marco Pavone, Bo~Li, Andreas Geiger, Peter Ondruska, Li~Erran Li,
  Dragomir Angelov, John Leonard, and Luc~Van Gool.
\newblock The argoverse stereo competition.
\newblock \url{https://cvpr2022.wad.vision}, 2022.

\bibitem{kuznietsov2017semi}
Yevhen Kuznietsov, Jorg Stuckler, , and Bastian Leibe.
\newblock Semi-supervised deep learning for monocular depth map prediction.
\newblock In {\em IEEE/CVF Conference on Computer Vision and Pattern
  Recognition (CVPR)}, pages 6647--6655, 2017.

\bibitem{laga2020survey}
Hamid Laga, Laurent~Valentin Jospin, Farid Boussaid, and Mohammed Bennamoun.
\newblock A survey on deep learning techniques for stereo-based depth
  estimation.
\newblock {\em IEEE Transactions on Pattern Analysis and Machine Intelligence
  (PAMI)}, 44(4):1738--1764, 2020.

\bibitem{lang2019pointpillars}
Alex~H Lang, Sourabh Vora, Holger Caesar, Lubing Zhou, Jiong Yang, and Oscar
  Beijbom.
\newblock Pointpillars: Fast encoders for object detection from point clouds.
\newblock In {\em IEEE/CVF Conference on Computer Vision and Pattern
  Recognition (CVPR)}, pages 12697--12705, 2019.

\bibitem{lee2019big}
Jin~Han Lee, Myung-Kyu Han, Dong~Wook Ko, and Il~Hong Suh.
\newblock From big to small: Multi-scale local planar guidance for monocular
  depth estimation.
\newblock {\em arXiv preprint arXiv:1907.10326}, 2019.

\bibitem{lee2022mpvit}
Youngwan Lee, Jonghee Kim, Jeffrey Willette, and Sung~Ju Hwang.
\newblock Mpvit: Multi-path vision transformer for dense prediction.
\newblock In {\em IEEE/CVF Conference on Computer Vision and Pattern
  Recognition (CVPR)}, pages 7287--7296, 2022.

\bibitem{li2022uncertainty}
Xiaotong Li, Yongxing Dai, Yixiao Ge, Jun Liu, Ying Shan, and Ling-Yu Duan.
\newblock Uncertainty modeling for out-of-distribution generalization.
\newblock In {\em International Conference on Learning Representations (ICLR)},
  2022.

\bibitem{li2021panodepth}
Yuyan Li, Zhixin Yan, Ye~Duan, and Liu Ren.
\newblock Panodepth: A two-stage approach for monocular omnidirectional depth
  estimation.
\newblock In {\em IEEE International Conference on 3D Vision (3DV)}, pages
  648--658, 2021.

\bibitem{li2018megadepth}
Zhengqi Li and Noah Snavely.
\newblock Megadepth: Learning single-view depth prediction from internet
  photos.
\newblock In {\em IEEE/CVF Conference on Computer Vision and Pattern
  Recognition (CVPR)}, pages 2041--2050, 2018.

\bibitem{lidepthtoolbox2022}
Zhenyu Li.
\newblock Monocular depth estimation toolbox.
\newblock \url{https://github.com/zhyever/Monocular-Depth-Estimation-Toolbox},
  2022.

\bibitem{li2022simipu}
Zhenyu Li, Zehui Chen, Ang Li, Liangji Fang, Qinhong Jiang, Xianming Liu,
  Junjun Jiang, Bolei Zhou, and Hang Zhao.
\newblock Simipu: Simple 2d image and 3d point cloud unsupervised pre-training
  for spatial-aware visual representations.
\newblock In {\em AAAI Conference on Artificial Intelligence (AAAI)}, 2022.

\bibitem{li2022depthformer}
Zhenyu Li, Zehui Chen, Xianming Liu, and Junjun Jiang.
\newblock Depthformer: Exploiting long-range correlation and local information
  for accurate monocular depth estimation.
\newblock {\em arXiv preprint arXiv:2203.14211}, 2022.

\bibitem{li2022binsformer}
Zhenyu Li, Xuyang Wang, Xianming Liu, and Junjun Jiang.
\newblock Binsformer: Revisiting adaptive bins for monocular depth estimation.
\newblock {\em arXiv preprint arXiv:2204.00987}, 2022.

\bibitem{liang2021swin-ir}
Jingyun Liang, Jiezhang Cao, Guolei Sun, Kai Zhang, Luc~Van Gool, and Radu
  Timofte.
\newblock Swinir: Image restoration using swin transformer.
\newblock In {\em IEEE/CVF International Conference on Computer Vision (ICCV)},
  pages 1833--1844, 2021.

\bibitem{COCO}
Tsung-Yi Lin, Michael Maire, Serge Belongie, James Hays, Pietro Perona, Deva
  Ramanan, Piotr Dollár, and C.~Lawrence Zitnick.
\newblock Microsoft coco: Common objects in context.
\newblock In {\em European Conference on Computer Vision (ECCV)}, pages
  740--755, 2014.

\bibitem{liu2015deep}
Fayao Liu, Chunhua Shen, and Guosheng Lin.
\newblock Deep convolutional neural fields for depth estimation from a single
  image.
\newblock In {\em IEEE/CVF Conference on Computer Vision and Pattern
  Recognition (CVPR)}, pages 5162--5170, 2015.

\bibitem{liu2023segment_any_point_cloud}
Youquan Liu, Lingdong Kong, Jun Cen, Runnan Chen, Wenwei Zhang, Liang Pan, Kai
  Chen, and Ziwei Liu.
\newblock The segment any point cloud codebase.
\newblock \url{https://github.com/youquanl/Segment-Any-Point-Cloud}, 2023.

\bibitem{liu2023segment}
Youquan Liu, Lingdong Kong, Jun Cen, Runnan Chen, Wenwei Zhang, Liang Pan, Kai
  Chen, and Ziwei Liu.
\newblock Segment any point cloud sequences by distilling vision foundation
  models.
\newblock {\em arXiv preprint arXiv:2306.09347}, 2023.

\bibitem{liu2022swin-v2}
Ze~Liu, Han Hu, Yutong Lin, Zhuliang Yao, Zhenda Xie, Yixuan Wei, Jia Ning, Yue
  Cao, Zheng Zhang, Li~Dong, Furu Wei, and Baining Guo.
\newblock Swin transformer v2: Scaling up capacity and resolution.
\newblock In {\em IEEE/CVF Conference on Computer Vision and Pattern
  Recognition (CVPR)}, pages 12009--12019, 2022.

\bibitem{liu2021swin}
Ze~Liu, Yutong Lin, Yue Cao, Han Hu, Yixuan Wei, Zheng Zhang, Stephen Lin, and
  Baining Guo.
\newblock Swin transformer: Hierarchical vision transformer using shifted
  windows.
\newblock In {\em IEEE/CVF International Conference on Computer Vision (ICCV)},
  pages 10012--10022, 2021.

\bibitem{loshchilov2018adamw}
Ilya Loshchilov and Frank Hutter.
\newblock Decoupled weight decay regularization.
\newblock In {\em International Conference on Learning Representations (ICLR)},
  2018.

\bibitem{luo2019every}
Chenxu Luo, Zhenheng Yang, Peng Wang, Yang Wang, Wei Xu, Ram Nevatia, and Alan
  Yuille.
\newblock Every pixel counts++: Joint learning of geometry and motion with 3d
  holistic understanding.
\newblock {\em IEEE Transactions on Pattern Analysis and Machine Intelligence
  (PAMI)}, 42(10):2624--2641, 2019.

\bibitem{madry2018adversarial}
Aleksander Madry, Aleksandar Makelov, Ludwig Schmidt, Dimitris Tsipras, and
  Adrian Vladu.
\newblock Towards deep learning models resistant to adversarial attacks.
\newblock In {\em International Conference on Learning Representations (ICLR)},
  2018.

\bibitem{michaelis2019dragon}
Claudio Michaelis, Benjamin Mitzkus, Robert Geirhos, Evgenia Rusak, Oliver
  Bringmann, Alexander~S. Ecker, Matthias Bethge, and Wieland Brendel.
\newblock Benchmarking robustness in object detection: Autonomous driving when
  winter is coming.
\newblock {\em arXiv preprint arXiv:1907.07484}, 2019.

\bibitem{ming2021survey}
Yue Ming, Xuyang Meng, Chunxiao Fan, and Hui Yu.
\newblock Deep learning for monocular depth estimation: A reviews.
\newblock {\em Neurocomputing}, 438:14--33, 2021.

\bibitem{ning2023ait}
Jia Ning, Chen Li, Zheng Zhang, Zigang Geng, Qi~Dai, Kun He, and Han Hu.
\newblock All in tokens: Unifying output space of visual tasks via soft token.
\newblock {\em arXiv preprint arXiv:2301.02229}, 2023.

\bibitem{oord2017vqvae}
Aaron Van~Den Oord and Oriol Vinyals.
\newblock Neural discrete representation learning.
\newblock In {\em Advances in Neural Information Processing System (NeurIPS)},
  2017.

\bibitem{oquab2023dinov2}
Maxime Oquab, Timothée Darcet, Théo Moutakanni, Huy Vo, Marc Szafraniec,
  Vasil Khalidov, Pierre Fernandez, Daniel Haziza, Francisco Massa, Alaaeldin
  El-Nouby, Mahmoud Assran, Nicolas Ballas, Wojciech Galuba, Russell Howes,
  Po-Yao Huang, Shang-Wen Li, Ishan Misra, Michael Rabbat, Vasu Sharma, Gabriel
  Synnaeve, Hu~Xu, Hervé Jegou, Julien Mairal, Patrick Labatut, Armand Joulin,
  and Piotr Bojanowski.
\newblock Dinov2: Learning robust visual features without supervision.
\newblock {\em arXiv preprint arXiv:2304.07193}, 2023.

\bibitem{Codalab}
Adrien Pavao, Isabelle Guyon, Anne-Catherine Letournel, Xavier Baró, Hugo
  Escalante, Sergio Escalera, Tyler Thomas, and Zhen Xu.
\newblock Codalab competitions: An open source platform to organize scientific
  challenges.
\newblock {\em PhD Dissertation, Université Paris-Saclay, FRA}, 2022.

\bibitem{pillai2019superdepth}
Sudeep Pillai, Rareş Ambruş, and Adrien Gaidon.
\newblock Superdepth: Self-supervised, super-resolved monocular depth
  estimation.
\newblock In {\em IEEE International Conference on Robotics and Automation
  (ICRA)}, pages 9250--9256, 2019.

\bibitem{poggi2020uncertainty}
Matteo Poggi, Filippo Aleotti, Fabio Tosi, and Stefano Mattoccia.
\newblock On the uncertainty of self-supervised monocular depth estimation.
\newblock In {\em IEEE/CVF Conference on Computer Vision and Pattern
  Recognition (CVPR)}, pages 3227--3237, 2020.

\bibitem{radford2021clip}
Alec Radford, Jong~Wook Kim, Chris Hallacy, Aditya Ramesh, Gabriel Goh,
  Sandhini Agarwal, Girish Sastry, Amanda Askell, Pamela Mishkin, Jack Clark,
  Gretchen Krueger, and Ilya Sutskever.
\newblock Learning transferable visual models from natural language
  supervision.
\newblock In {\em International Conference on Machine Learning (ICML)}, pages
  8748--8763, 2021.

\bibitem{NTIRE}
Pierluigi~Zama Ramirez, Fabio Tosi, Luigi~Di Stefano, Radu Timofte, Alex
  Costanzino, Matteo~Poggi andSamuele Salti, Stefano Mattoccia, Jun Shi, Dafeng
  Zhang, Yong A, Yixiang Jin, Dingzhe Li, Chao Li, Zhiwen Liu, Qi~Zhang, Yixing
  Wang, and Shi Yin.
\newblock Ntire 2023 challenge on hr depth from images of specular and
  transparent surfaces.
\newblock In {\em IEEE/CVF Conference on Computer Vision and Pattern
  Recognition Workshops (CVPRW)}, pages 1384--1395, 2023.

\bibitem{ranftl2021dpt}
Ren{\'e} Ranftl, Alexey Bochkovskiy, and Vladlen Koltun.
\newblock Vision transformers for dense prediction.
\newblock In {\em IEEE/CVF International Conference on Computer Vision (ICCV)},
  pages 12179--12188, 2021.

\bibitem{ranftl2022towards}
René Ranftl, Katrin Lasinger, David Hafner, Konrad Schindler, and Vladlen
  Koltun.
\newblock Towards robust monocular depth estimation: Mixing datasets for
  zero-shot cross-dataset transfer.
\newblock {\em IEEE Transactions on Pattern Analysis and Machine Intelligence
  (PAMI)}, 44(3):1623--1637, 2022.

\bibitem{ranjan2019competitive}
Anurag Ranjan, Varun Jampani, Lukas Balles, Kihwan Kim, Deqing Sun, Jonas
  Wulff, and Michael~J. Black.
\newblock Competitive collaboration: Joint unsupervised learning of depth,
  camera motion, optical flow and motion segmentation.
\newblock In {\em IEEE/CVF Conference on Computer Vision and Pattern
  Recognition (CVPR)}, pages 12240--12249, 2019.

\bibitem{PointCloud-C}
Jiawei Ren, Lingdong Kong, Liang Pan, and Ziwei Liu.
\newblock The pointcloud-c benchmark for robust point cloud perception under
  corruptions.
\newblock \url{https://github.com/ldkong1205/PointCloud-C}, 2022.

\bibitem{ren2022modelnet-c}
Jiawei Ren, Liang Pan, and Ziwei Liu.
\newblock Benchmarking and analyzing point cloud classification under
  corruptions.
\newblock {\em International Conference on Machine Learning (ICML)}, 2022.

\bibitem{rombach2022stable}
Robin Rombach, Andreas Blattmann, Dominik Lorenz, Patrick Esser, and Björn
  Ommer.
\newblock High-resolution image synthesis with latent diffusion models.
\newblock In {\em IEEE/CVF Conference on Computer Vision and Pattern
  Recognition (CVPR)}, pages 10684--10695, 2022.

\bibitem{ronneberger2015unet}
Olaf Ronneberger, Philipp Fischer, and Thomas Brox.
\newblock U-net: Convolutional networks for biomedical image segmentation.
\newblock In {\em International Conference on Medical Image Computing and
  Computer-Assisted Intervention (MICCAI)}, pages 234--241, 2015.

\bibitem{schellevis2019maskocc}
Maarten Schellevis.
\newblock Improving self-supervised single view depth estimation by masking
  occlusion.
\newblock {\em arXiv preprint arXiv:1908.11112}, 2019.

\bibitem{schon2021mgnet}
Markus Schön, Michael Buchholz, and Klaus Dietmayer.
\newblock Mgnet: Monocular geometric scene understanding for autonomous
  driving.
\newblock In {\em IEEE/CVF International Conference on Computer Vision (ICCV)},
  pages 15804--15815, 2021.

\bibitem{silberman2012nyu2}
Nathan Silberman, Derek Hoiem, Pushmeet Kohli, and Rob Fergus.
\newblock Indoor segmentation and support inference from rgbd images.
\newblock In {\em European Conference on Computer Vision (ECCV)}, pages
  746--760, 2012.

\bibitem{spencer2020general}
Jaime Spencer, Richard Bowden, and Simon Hadfield.
\newblock Defeat-net: General monocular depth via simultaneous unsupervised
  representation learning.
\newblock In {\em IEEE/CVF Conference on Computer Vision and Pattern
  Recognition (CVPR)}, pages 14402--14413, 2020.

\bibitem{MDEC}
Jaime Spencer, C.~Stella Qian, Chris Russell, Simon Hadfield, Erich Graf, Wendy
  Adams, Andrew~J. Schofield, James~H. Elder, Richard Bowden, Heng Cong,
  Stefano Mattoccia, Matteo Poggi, Zeeshan~Khan Suri, Yang Tang, Fabio Tosi,
  Hao Wang, Youmin Zhang, Yusheng Zhang, and Chaoqiang Zhao.
\newblock The monocular depth estimation challenge.
\newblock In {\em IEEE/CVF Winter Conference on Applications of Computer Vision
  Workshops (WACVW)}, pages 623--632, 2023.

\bibitem{MDEC2}
Jaime Spencer, C.~Stella Qian, Michaela Trescakova, Chris Russell, Simon
  Hadfield, Erich Graf, Wendy Adams, Andrew~J. Schofield, James Elder, Richard
  Bowden, Ali Anwar, Hao Chen, Xiaozhi Chen, Kai Cheng, Yuchao Dai, Huynh~Thai
  Hoa, Sadat Hossain, Jianmian Huang, Mohan Jing, Bo~Li, Chao Li, Baojun Li,
  Zhiwen Liu, Stefano Mattoccia, Siegfried Mercelis, Myungwoo Nam, Matteo
  Poggi, Xiaohua Qi, Jiahui Ren, Yang Tang, Fabio Tosi, Linh Trinh, S~M~Nadim
  Uddin, Khan~Muhammad Umair, Kaixuan Wang, Yufei Wang, Yixing Wang, Mochu
  Xiang, Guangkai Xu, Wei Yin, Jun Yu, Qi~Zhang, and Chaoqiang Zhao.
\newblock The second monocular depth estimation challenge.
\newblock In {\em IEEE/CVF Conference on Computer Vision and Pattern
  Recognition Workshops (CVPRW)}, pages 3063--3075, 2023.

\bibitem{sun2022scdepthv3}
Libo Sun, Jia-Wang Bian, Huangying Zhan, Wei Yin, Ian Reid, and Chunhua Shen.
\newblock Sc-depthv3: Robust self-supervised monocular depth estimation for
  dynamic scenes.
\newblock {\em arXiv preprint arXiv:2211.03660}, 2022.

\bibitem{tosi2019learning}
Fabio Tosi, Filippo Aleotti, Matteo Poggi, and Stefano Mattoccia.
\newblock Learning monocular depth estimation infusing traditional stereo
  knowledge.
\newblock In {\em IEEE/CVF Conference on Computer Vision and Pattern
  Recognition (CVPR)}, pages 9799--9809, 2019.

\bibitem{uhrig2017kitti}
Jonas Uhrig, Nick Schneider, Lukas Schneider, Uwe Franke, Thomas Brox, and
  Andreas Geiger.
\newblock Sparsity invariant cnns.
\newblock In {\em IEEE International Conference on 3D Vision (3DV)}, pages
  11--20, 2017.

\bibitem{casser2019depth}
Casser Vincent, Soeren Pirk, Reza Mahjourian, and Anelia Angelova.
\newblock Depth prediction without the sensors: Leveraging structure for
  unsupervised learning from monocular videos.
\newblock In {\em AAAI Conference on Artificial Intelligence (AAAI)}, pages
  8001--8008, 2019.

\bibitem{wang2018learning}
Chaoyang Wang, José~Miguel Buenaposada, Rui Zhu, and Simon Lucey.
\newblock Learning depth from monocular videos using direct methods.
\newblock In {\em IEEE/CVF Conference on Computer Vision and Pattern
  Recognition (CVPR)}, pages 2022--2030, 2018.

\bibitem{AdvMix}
Jiahang Wang, Sheng Jin, Wentao Liu, Weizhong Liu, Chen Qian, and Ping Luo.
\newblock When human pose estimation meets robustness: Adversarial algorithms
  and benchmarks.
\newblock In {\em IEEE/CVF Conference on Computer Vision and Pattern
  Recognition (CVPR)}, pages 11855--11864, 2021.

\bibitem{wang2020sdc}
Lijun Wang, Jianming Zhang, Oliver Wang, Zhe Lin, and Huchuan Lu.
\newblock Sdc-depth: Semantic divide-and-conquer network for monocular depth
  estimation.
\newblock In {\em IEEE/CVF Conference on Computer Vision and Pattern
  Recognition (CVPR)}, pages 541--550, 2020.

\bibitem{wang2023survey_robustness}
Shunxin Wang, Raymond Veldhuis, and Nicola Strisciuglio.
\newblock The robustness of computer vision models against common corruptions:
  A survey.
\newblock {\em arXiv preprint arXiv:2305.06024}, 2023.

\bibitem{watson2021manydepth}
Jamie Watson, Oisin~Mac Aodha, Victor Prisacariu, Gabriel Brostow, and Michael
  Firman.
\newblock The temporal opportunist: Self-supervised multi-frame monocular
  depth.
\newblock In {\em IEEE/CVF Conference on Computer Vision and Pattern
  Recognition (CVPR)}, pages 1164--1174, 2021.

\bibitem{woo2018cbam}
Sanghyun Woo, Jongchan Park, Joon-Young Lee, and In~So Kweon.
\newblock Cbam: Convolutional block attention module.
\newblock In {\em European Conference on Computer Vision (ECCV)}, pages 3--19,
  2018.

\bibitem{xiang2018posecnn}
Yu~Xiang, Tanner Schmidt, Venkatraman Narayanan, and Dieter Fox.
\newblock Posecnn: A convolutional neural network for 6d object pose estimation
  in cluttered scenes.
\newblock In {\em Robotics: Science and Systems (RSS)}, 2018.

\bibitem{xie2023robobev_codebase}
Shaoyuan Xie, Lingdong Kong, Wenwei Zhang, Jiawei Ren, Liang Pan, Kai Chen, and
  Ziwei Liu.
\newblock The robobev benchmark for robust bird's eye view detection under
  common corruption and domain shift.
\newblock \url{https://github.com/Daniel-xsy/RoboBEV}, 2023.

\bibitem{xie2023robobev}
Shaoyuan Xie, Lingdong Kong, Wenwei Zhang, Jiawei Ren, Liang Pan, Kai Chen, and
  Ziwei Liu.
\newblock Robobev: Towards robust bird's eye view perception under corruptions.
\newblock {\em arXiv preprint arXiv:2304.06719}, 2023.

\bibitem{xie2023adv}
Shaoyuan Xie, Zichao Li, Zeyu Wang, and Cihang Xie.
\newblock On the adversarial robustness of camera-based 3d object detection.
\newblock {\em arXiv preprint arXiv:2301.10766}, 2023.

\bibitem{xie2023revealing}
Zhenda Xie, Zigang Geng, Jingcheng Hu, Zheng Zhang, Han Hu, and Yue Cao.
\newblock Revealing the dark secrets of masked image modeling.
\newblock In {\em IEEE/CVF Conference on Computer Vision and Pattern
  Recognition (CVPR)}, pages 14475--14485, 2023.

\bibitem{xie2022simmim}
Zhenda Xie, Zheng Zhang, Yue Cao, Yutong Lin, Jianmin Bao, Zhuliang Yao,
  Qi~Dai, and Han Hu.
\newblock Simmim: A simple framework for masked image modeling.
\newblock In {\em IEEE/CVF Conference on Computer Vision and Pattern
  Recognition (CVPR)}, pages 9653--9663, 2022.

\bibitem{xue2020dnet}
Feng Xue, Guirong Zhuo, Ziyuan Huang, Wufei Fu, Zhuoyue Wu, and Marcelo~H. Ang.
\newblock Toward hierarchical self-supervised monocular absolute depth
  estimation for autonomous driving applications.
\newblock In {\em IEEE/RSJ International Conference on Intelligent Robots and
  Systems (IROS)}, pages 2330--2337, 2020.

\bibitem{yan2021cadepth}
Jiaxing Yan, Hong Zhao, Penghui Bu, and YuSheng Jin.
\newblock Channel-wise attention-based network for self-supervised monocular
  depth estimation.
\newblock In {\em IEEE International Conference on 3D Vision (3DV)}, pages
  464--473, 2021.

\bibitem{second}
Yan Yan, Yuxing Mao, and Bo~Li.
\newblock Second: Sparsely embedded convolutional detection.
\newblock {\em Sensors}, 18(10):3337, 2018.

\bibitem{tang2021transdepth}
Guanglei Yang, Hao Tang, Mingli Ding, Nicu Sebe, and Elisa Ricci.
\newblock Transformer-based attention networks for continuous pixel-wise
  prediction.
\newblock In {\em IEEE/CVF Conference on Computer Vision and Pattern
  Recognition (CVPR)}, pages 16269--16279, 2021.

\bibitem{yang2020d3vo}
Nan Yang, Lukas von Stumberg, Rui Wang, and Daniel Cremers.
\newblock D3vo: Deep depth, deep pose and deep uncertainty for monocular visual
  odometry.
\newblock In {\em IEEE/CVF Conference on Computer Vision and Pattern
  Recognition (CVPR)}, pages 1281--1292, 2020.

\bibitem{yang2020mobile3d}
Xingbin Yang, Liyang Zhou, Hanqing Jiang, Zhongliang Tang, Yuanbo Wang, Hujun
  Bao, and Guofeng Zhang.
\newblock Mobile3drecon: real-time monocular 3d reconstruction on a mobile
  phone.
\newblock {\em IEEE Transactions on Visualization and Computer Graphics
  (TVCG)}, 26(12):3446--3456, 2020.

\bibitem{Kinetics-C}
Chenyu Yi, Siyuan Yang, Haoliang Li, Yap peng Tan, and Alex Kot.
\newblock Benchmarking the robustness of spatial-temporal models against
  corruptions.
\newblock In {\em Advances in Neural Information Processing System (NeurIPS)},
  2021.

\bibitem{centerpoint}
Tianwei Yin, Xingyi Zhou, and Philipp Krahenbuhl.
\newblock Center-based 3d object detection and tracking.
\newblock In {\em IEEE/CVF Conference on Computer Vision and Pattern
  Recognition (CVPR)}, pages 11784--11793, 2021.

\bibitem{yuan2022newcrfs}
Weihao Yuan, Xiaodong Gu, Zuozhuo Dai, Siyu Zhu, and Ping Tan.
\newblock New crfs: Neural window fully-connected crfs for monocular depth
  estimation.
\newblock {\em arXiv preprint arXiv:2203.01502}, 2022.

\bibitem{yucel2021real}
Mehmet~Kerim Yucel, Valia Dimaridou, Anastasios Drosou, and Albert Saa-Garriga.
\newblock Real-time monocular depth estimation with sparse supervision on
  mobile.
\newblock In {\em IEEE/CVF Conference on Computer Vision and Pattern
  Recognition (CVPR)}, pages 2428--2437, 2021.

\bibitem{yun2019cutmix}
Sangdoo Yun, Dongyoon Han, Seong~Joon Oh, Sanghyuk Chun, Junsuk Choe, and
  Youngjoon Yoo.
\newblock Cutmix: Regularization strategy to train strong classifiers with
  localizable features.
\newblock In {\em IEEE/CVF Conference on Computer Vision and Pattern
  Recognition (CVPR)}, pages 6023--6032, 2019.

\bibitem{zamir2022restormer}
Syed~Waqas Zamir, Aditya Arora, Salman Khan, Munawar Hayat, Fahad~Shahbaz Khan,
  and Ming-Hsuan Yang.
\newblock Restormer: Efficient transformer for high-resolution image
  restoration.
\newblock In {\em IEEE/CVF Conference on Computer Vision and Pattern
  Recognition (CVPR)}, pages 5728--5739, 2022.

\bibitem{RVC}
Oliver Zendel, Angela Dai, Xavier~Puig Fernandez, Andreas Geiger, Vladen
  Koltun, Peter Kontschieder, Adam Kortylewski, Tsung-Yi Lin, Torsten Sattler,
  Daniel Scharstein, Hendrik Schilling, Jonas Uhrig, and Jonas Wulff.
\newblock The robust vision challenge.
\newblock \url{http://www.robustvision.net}, 2022.

\bibitem{zhang2018mixup}
Hongyi Zhang, Moustapha Cisse, Yann~N. Dauphin, and David Lopez-Paz.
\newblock mixup: Beyond empirical risk minimization.
\newblock In {\em International Conference on Learning Representations (ICLR)},
  2018.

\bibitem{zhang2023litemono}
Ning Zhang, Francesco Nex, George Vosselman, and Norman Kerle.
\newblock Lite-mono: A lightweight cnn and transformer architecture for
  self-supervised monocular depth estimation.
\newblock In {\em IEEE/CVF Conference on Computer Vision and Pattern
  Recognition (CVPR)}, pages 18537--18546, 2023.

\bibitem{zhang2022dynadepth}
Sen Zhang, Jing Zhang, and Dacheng Tao.
\newblock Towards scale-aware, robust, and generalizable unsupervised monocular
  depth estimation by integrating imu motion dynamics.
\newblock In {\em European Conference on Computer Vision (ECCV)}, pages
  143--160, 2022.

\bibitem{zhao2020survey}
Chaoqiang Zhao, Qiyu Sun, Chongzhen Zhang, Yang Tang, and Feng Qian.
\newblock Monocular depth estimation based on deep learning: An overview.
\newblock {\em Science China Technological Sciences}, 63(9):1612--1627, 2020.

\bibitem{zhao2021monovit}
Chaoqiang Zhao, Youmin Zhang, Matteo Poggi, Fabio Tosi, Xianda Guo, Zheng Zhu,
  Guan Huang, Yang Tang, and Stefano Mattoccia.
\newblock Monovit: Self-supervised monocular depth estimation with a vision
  transformer.
\newblock In {\em IEEE International Conference on 3D Vision (3DV)}, pages
  668--678, 2022.

\bibitem{zhao2016loss}
Hang Zhao, Orazio Gallo, Iuri Frosio, and Jan Kautz.
\newblock Loss functions for image restoration with neural networks.
\newblock {\em IEEE Transactions on Computational Imaging (TCI)}, 3(1):47--57,
  2016.

\bibitem{zhao2023unleashing}
Wenliang Zhao, Yongming Rao, Zuyan Liu, Benlin Liu, Jie Zhou, and Jiwen Lu.
\newblock Unleashing text-to-image diffusion models for visual perception.
\newblock {\em arXiv preprint arXiv:2303.02153}, 2023.

\bibitem{zhou2021diffnet}
Hang Zhou, David Greenwood, and Sarah Taylor.
\newblock Self-supervised monocular depth estimation with internal feature
  fusion.
\newblock In {\em British Machine Vision Conference (BMVC)}, 2021.

\bibitem{zhou2017sfm}
Tinghui Zhou, Matthew Brown, Noah Snavely, and David~G. Lowe.
\newblock Unsupervised learning of depth and ego-motion from video.
\newblock In {\em IEEE/CVF Conference on Computer Vision and Pattern
  Recognition (CVPR)}, pages 1851--1858, 2017.

\end{thebibliography}
\end{document}